\pdfoutput=1

\documentclass[11pt]{article}

\usepackage{acl}

\usepackage{times}
\usepackage{latexsym}

\usepackage[T1]{fontenc}

\usepackage[utf8]{inputenc}

\usepackage{microtype}

\usepackage{graphicx}
\usepackage{amsmath}
\usepackage{amssymb}
\usepackage{booktabs}
\usepackage[font=normalsize]{caption}
\usepackage{textcomp}

\usepackage{subcaption}
\usepackage{xcolor}
\usepackage{algorithm}
\usepackage{algpseudocode}
\usepackage{booktabs}

\usepackage[many]{tcolorbox}  
\newtcolorbox{boxL}{
    fontupper = \color{black},
    rounded corners,
    arc = 6pt,
    colframe = black!50, 
    boxrule = 0pt, 
    bottomrule = 4.5pt ,
    breakable,
}
\usepackage{enumitem}
\usepackage{pythonhighlight}
\usepackage{mathtools}
\usepackage{amsthm}
\usepackage{breqn}

\usepackage[capitalize]{cleveref}
\crefname{section}{Sec.}{Secs.}
\Crefname{section}{Section}{Sections}
\Crefname{table}{Table}{Tables}
\crefname{table}{Tab.}{Tabs.}

\title{PRompt Optimization in Multi-Step Tasks (PROMST): \\Integrating Human Feedback and Heuristic-based Sampling}

\author{Yongchao Chen \\
  MIT / Harvard University \\
  \small\texttt{yongchaochen@fas.harvard.edu} \\\And
  Jacob Arkin \\
  MIT \\
  \small\texttt{jarkin@mit.edu} 
  \\\And
  Yilun Hao \\
  MIT \\
  \small\texttt{yilunhao@mit.edu}
  \AND
  Yang Zhang \\
  MIT-IBM Watson AI Lab \\
  \small\texttt{Yang.Zhang2@ibm.com}
  \\\And
  Nicholas Roy \\
  MIT \\
  \small\texttt{nickroy@csail.mit.edu} 
  \\\And
  Chuchu Fan \\
  MIT \\
  \small\texttt{chuchu@mit.edu} \\  }

\begin{document}
\maketitle
\begin{abstract}
Prompt optimization aims to find the best prompt to a large language model (LLM) for a given task. LLMs have been successfully used to help find and improve prompt candidates for single-step tasks. However, realistic tasks for agents are multi-step and introduce new challenges: (1) Prompt content is likely to be more extensive and complex, making it more difficult for LLMs to analyze errors, (2) the impact of an individual step is difficult to evaluate, and (3) different people may have varied preferences about task execution. While humans struggle to optimize prompts, they are good at providing feedback about LLM outputs; we therefore introduce a new LLM-driven discrete prompt optimization framework PRompt Optimization in Multi-Step Tasks (PROMST) that incorporates human-designed feedback rules to automatically offer direct suggestions for improvement. We also use an extra learned heuristic model that predicts prompt performance to efficiently sample from prompt candidates. This approach significantly outperforms both human-engineered prompts and several other prompt optimization methods across 11 representative multi-step tasks (an average 10.6\%-29.3\% improvement to current best methods on five LLMs respectively). We believe our work can serve as a benchmark for automatic prompt optimization for LLM-driven multi-step tasks. Datasets and Codes are available at \url{https://github.com/yongchao98/PROMST}. Project Page is available at \url{https://yongchao98.github.io/MIT-REALM-PROMST/}.
\end{abstract}

\begin{figure*}[h]
  \centering
   \includegraphics[width=0.7\linewidth]{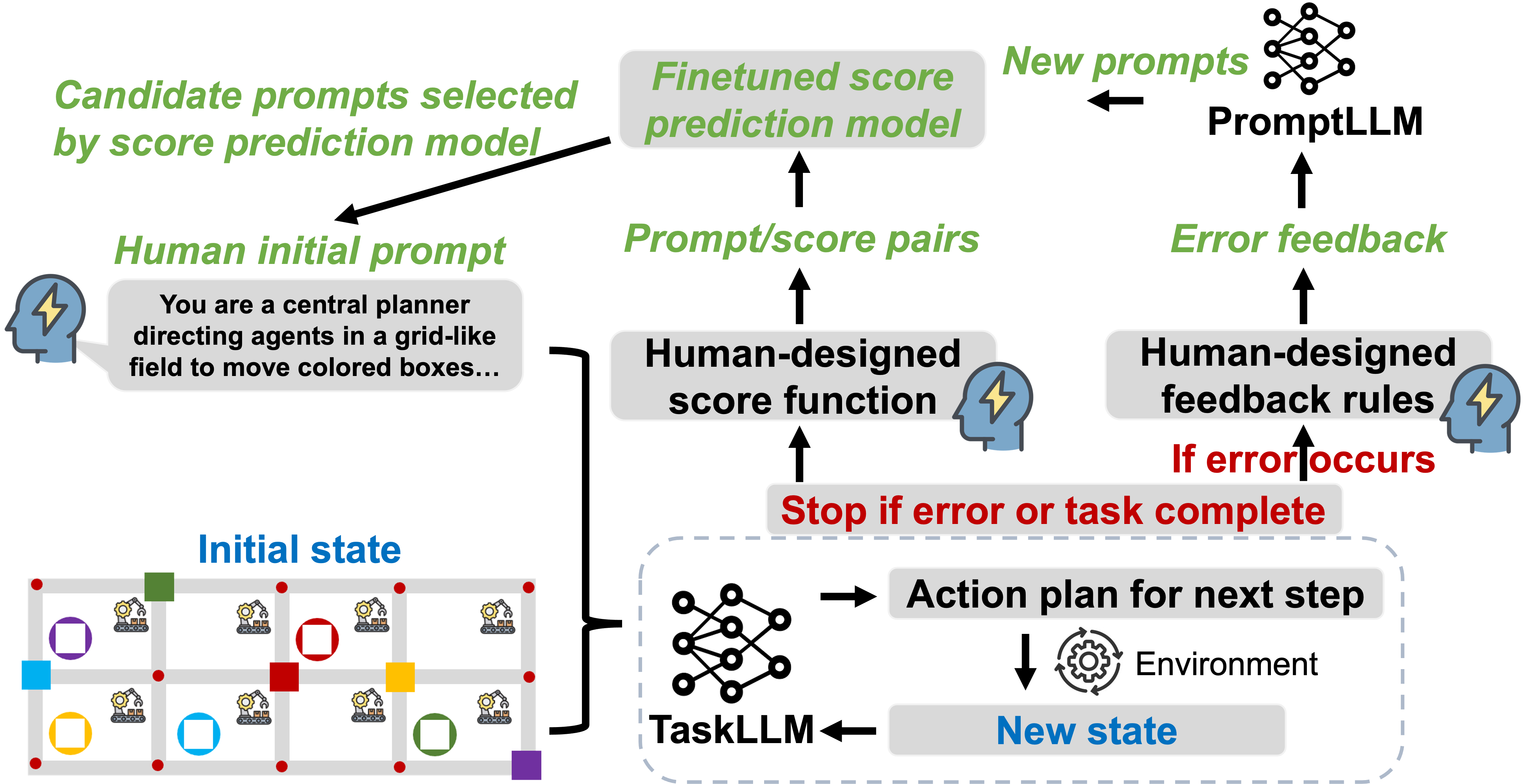}

   \caption{The PROMST framework. Given an initial human-designed prompt and the state of the environment for the current task, the TaskLLM iteratively generates an action and executes it until either an error occurs or the task is complete. Human-designed feedback rules automatically generate feedback about errors that is then provided as context to the PromptLLM when generating new prompt candidates. The task performance is scored according to a human-designed score function; this score can be used with the prompt to train a score prediction model online. Given new prompt candidates, this score prediction model is used to select a subset of candidates to evaluate for the next generation.}
   \label{fig:Framework}
\end{figure*}

\section{Introduction}
\label{submission}
The performance of large language models (LLMs) on a given task is sensitive to the prompt, so prompt engineering aims to create prompts that fully leverage the capabilities of LLMs. Due to the lack of access to model parameters for black-box LLMs, techniques for automatic prompt optimization have primarily focused on searching over the vast discrete space of tokenized language inputs~\citep{black-box-prompt-optimize}. Recent studies have shown that LLMs, combined with evolutionary algorithms, can help with this search by reasoning over errors made using existing prompts to suggest edits or generate new candidate prompts \cite{pryzant2023automatic,promptagent,LLM-as-optimizers}. These approaches have been evaluated on relatively simple one-step tasks, such as mathematical calculations~\citep{math1,math2}, instruction induction~\citep{Instruction-induction}, and factual analysis~\citep{reasoning}. The associated prompts are also relatively short, usually one to three sentences.

In this work, we aim to optimize prompts for LLM-driven agents solving multi-step tasks and propose a method called PRompt Optimization in Multi-Step Tasks (PROMST). In these tasks, an LLM is used to decide a system's actions (e.g., virtual software \citealt{autogen,webarena} or real robots~\citealt{chen2023autotamp,review-foundation-model-robotics}) as it interacts with an environment over multiple steps~\cite{LMRL-gym}. Engineering good prompts is hard due to the typical prompt length (300+ tokens) and individual task constraints and rules. The prompts needed for multi-step tasks are more complex to judge the long-horizon correctness of a single action. This difficulty hinders LLMs from automatically reasoning over errors and producing better prompts, which in turn reduces the effectiveness of current methods for automated prompt optimization. Prompt optimization in multi-step tasks is still an open challenge.

Considering that humans excel in analyzing errors and incorporating relevant domain knowledge into feedback, we formalize PROMST as a framework involving human input, as shown in Figure~\ref{fig:Framework}. Here, during the multi-step agent-environment interactions, the agent (indicated by `TaskLLM' in Figure~\ref{fig:Framework}) sometimes makes errors and fails the task. While some work has used LLMs to evaluate errors, we instead use human-designed feedback rules constructed a priori that address different types of errors. Depending on the error, feedback is automatically generated and passed as additional context to an LLM that is responsible for producing a new set of candidate prompts (indicated by `PromptLLM' in Figure~\ref{fig:Framework}). A score is assigned to each prompt indicating the agent's task performance given that prompt. Since the evaluation of many candidate prompts for multi-step tasks in environments can be expensive, we fine-tune a score prediction model online using prompt-score pairs which can be used as a heuristic to select a subset of the candidate prompts to evaluate.

Our experiments in 11 tasks show that the integration of human feedback and the score model greatly improves the prompt optimization process (10.6\%-29.3\% relative improvements over all baseline methods across different LLMs). PROMST achieves the best performance on most tasks. PROMST has also been shown to perform better in multi-trial settings when combined with dynamic approaches. We further show that the human-designed evaluation rules can be used to help align task performance with human preferences. Extensive experiments are conducted to validate the framework and investigate the underlying reasons why some prompts are more effective than others.

In summary, our contributions are : (1) To our best knowledge, PROMST is the first to explore automatic prompt optimization in multi-step agent tasks. We release all codes and prompts for 11 multi-step environments, which may serve as a benchmark for future research. (2) We show that the integration of human feedback and a fine-tuned score model outperforms existing methods across various tasks and LLMs. (3) Our research indicates that PROMST is orthogonal and integrates well with established dynamic approaches. (4) We find that human-designed rules for task evaluation help align optimized prompts with human preferences.

\section{Related Work}
\textbf{Prompt Optimization}\quad 
To improve performance of black-box API models, it is useful to engineer the discrete prompts for downstream tasks. Various `best practices' have emerged for human-designed task prompts, such as including examples \citep{brown2020language} or promoting reasoning chains \citep{kojima2022large, wei2022chain}. However, manually designing prompts requires extensive human trial-and-error and is sub-optimal; thus, many recent works focus on automating this process. Some methods approximate the gradients \citep{diao2022black} or emulate them via natural language \citep{pryzant2023automatic}. Others use edit operators to modify an initial prompt, driven either by reinforcement learning \citep{zhang2023tempera} or score-guided search \citep{prasad2023grips}. To help balance exploration and exploitation of prompts, several approaches have used LLM-driven evolution \citep{guo2023connecting, promptbreeder, eureka, ye2023prompt}. In several works, LLMs are directly used to generate prompt candidates \citep{zhou2023large} often with feedback about parent prompts \citep{promptagent, eureka}. Our work focuses on domains that include complex multi-step tasks, in which the evaluation and reflection processes are more challenging so that score prediction models and human feedback rules are introduced in order to mitigate this problem.

\textbf{LLM Based Agents for Multi-Step Tasks}\quad There are many recent works that use LLMs for multi-step planning. LLMs are used to interact with softwares and websites \cite{agentboard,autogen,webarena}, plan robot actions \cite{chen2023scalable-multi-agent,saycan,llms-zero-shot-planners,eureka,Grid-world,nl2tl}, and connect to external tools \cite{llm+p, chen2023autotamp,toolllm}. Instead of careful design of lengthy prompts to capture all the constraints, our approach uses prompt optimization to transition from a simple initial human-provided prompt to a high-performing prompt.

\textbf{LLM Self-reflection from Feedback}\quad In planning domains, it is useful to provide feedback about syntactic errors \citep{generalized-planning-in-pddl-domains, errors-are-useful-prompts}, potential infinite loops \citep{generalized-planning-in-pddl-domains}, failed action execution \citep{inner-monologue}, and generated trajectories \citep{chen2023autotamp}. Other recent work has shown that LLM-generated feedback via self-evaluation can improve performance on a variety of tasks \citep{yang2022re3, welleck2022generating, madaan2023self}, including prompt engineering \citep{promptagent} and reinforcement learning \citep{shinn2023reflexion,eureka}. Compared to above works, our work combines LLM self-reflection with human-provided feedback templates to help improve performance on the more challenging multi-step tasks.

Another type of methods is to utilize self-reflection for online dynamic feedback during task execution, such as Reflexion \cite{reflexion}. While Reflexion needs multiple trials for online optimization of action memory for each single test, our method PROMST only needs one trial for task execution and the prompt is optimized offline across multiple tests. In Section~\ref{sec:results and analysis}, we find that PROMST outperforms the offline variation of Reflexion and can perform better when combining with online Reflexion in the multi-trial setting.

\section{Methodology}
\subsection{Problem Formulation}
Given a base LLM \( \mathcal{B} \) and a target task \( T \), the goal of prompt optimization is to craft an optimized natural language prompt \( P \) that maximizes the performance of \( \mathcal{B} \) on \( T \). Here the prompt $P$ consists of multiple components, such as a task description, scoring rules, and safety constraints. In multi-step tasks, the state information of the environment at each step will be transformed into a text string \( S \) and provided to the LLM \( \mathcal{B} \) to make decisions. The history of state ($S$), action ($a$), and environment feedback ($e$) will also be reported to LLM. For the $i^{th}$ testing trial on a particular task, the probability of an action sequence [$a_{i,1}, a_{i,2},..., a_{i,j}$] is:
\begin{equation}
\begin{aligned}
p_{\mathcal{B}}([a_{i,1}, a_{i,2},\ldots, a_{i,j}]) &= \prod_{k=1}^{j} p_{\mathcal{B}}(a_{i,k} | S_{i,k}, P, \\
&\quad S_{i,k-1}, a_{i,k-1}, e_{i,k-1}\\
&\quad ,\ldots, S_{i,1}, a_{i,1}, e_{i,1})
\end{aligned}
\label{eq:equation1}
\end{equation}

The sequence [$a_{i,1}, a_{i,2}, ..., a_{i,j}$] is executed in the task environment and assigned a score based on human-designed rules or functions $R$. The goal of prompt optimization is to find the optimal natural language prompt \( P^* \) that maximizes a score function \( R \):
\begin{equation}
P^* = \underset{P \in A}{\text{arg max}} \sum_{i \in U} R(p_{\mathcal{B}}([a_{i,1}, a_{i,2}, ..., a_{i,j}])),
\label{eq:equation2}
\end{equation}
where \( A \) denotes the vast and complex space of all possible natural language prompts and \( U \) denotes the set of all the testing trials in a specific task.

\subsection{PROMST Framework}
Figure~\ref{fig:Framework} illustrates the general framework of PROMST. The goal is to more efficiently and strategically search over the vast space of possible prompts while integrating human-designed feedback of candidate prompt performance. LLMs are used in two key steps of PROMST: (1) the execution of the task via the current candidate prompt (`TaskLLM') and (2) the generation of new candidate prompts given any available feedback about the current prompt's performance on the task (`PromptLLM'). We refer to a single execution in a testing case of a task as a trial. In each trial, the TaskLLM executes the task over multiple rounds of interaction with the environment; for each round, the TaskLLM is provided both the current candidate prompt and the current trial's execution history and generates the next action for the agent to take. Task execution terminates when an error is detected or the task is complete. The candidate prompt $P$ is assigned a score for that trial via the human-designed score function. Each candidate prompt is evaluated over multiple trials in which the initial environment state (e.g. number of objects, number of agents) is varied, resulting in a final average score calculated over all the trials. Once the candidate prompts have all been evaluated and assigned automatic feedback, the top performers are selected as parents for a new generation of candidate prompts. The PromptLLM uses each parent prompt and its feedback to generate new candidate prompts. This process is also described in Algorithm~\ref{alg:PROMST}, \ref{alg:2}, \ref{alg:3} in Appendix~\ref{sec:appendix-algorithms}.

\textbf{Score Prediction Model}\quad In general, producing more candidate prompts per generation allows for more exploration over the space of possible prompts; however, there is a trade-off between the number of candidates per generation and the cost of evaluation, and multi-step tasks can be much more expensive to evaluate (we query the TaskLLM for each next action). To help mitigate the evaluation cost for a generation, we learn a score prediction model online that functions as a heuristic with which to choose a subset of the generated candidate prompts for actual evaluation.

Algorithm~\ref{alg:2} in Appendix~\ref{sec:appendix-algorithms} shows the process of implementing the score prediction model as a heuristic for filtering candidate prompts. We fine-tune a task-specific bidirectional Longformer-base (148M)~\cite{longformer} model. The prompt-score pairs on which we fine-tune are collected online during early iterations of PROMST; therefore, the score prediction model is not applied until $sd^{th}$ generation, where $sd$ is a hyperparameter. We continue to update the learned model at each generation with the new prompt-score pairs. To mitigate variance, we fine-tune multiple models on five rounds with the collected data following a random 4:1 train/test split. The generated prompt candidate $p'$ will be selected for task evaluation if:
\begin{dmath} \label{eq:equation3}
\text{E}[M_k(p')] + \operatorname{Var}[M_k(p')] + \text{E}[\text{error}_k] \geq \text{hyper\_M} \times \max(D.\text{score}())
\end{dmath}
where E$[M_k(p')]$ and Var$[M_k(p')$ are the mean and variance of predicted scores for $p'$ from five models. The E$[error_k]$ is the average testing error of five score models. The $\max(D.score())$ is the highest score of existing prompts. To balance efficient exploration and conservative filtering, we only filter prompt candidates when the score prediction model is sufficiently confident. When the variance of the prediction model is high, we want to be conservative in its application in order to reduce the chance that we filter out a good candidate. Similarly, we choose to be conservative in our filtering when the error is high. Equation~\ref{eq:equation3} is therefore one formulation that incorporates these general ideas. The hyperparameter hyper\_M allows this conservativeness to be tuned by users.

\begin{figure}[ht]
  \centering
   \includegraphics[width=0.9\linewidth]{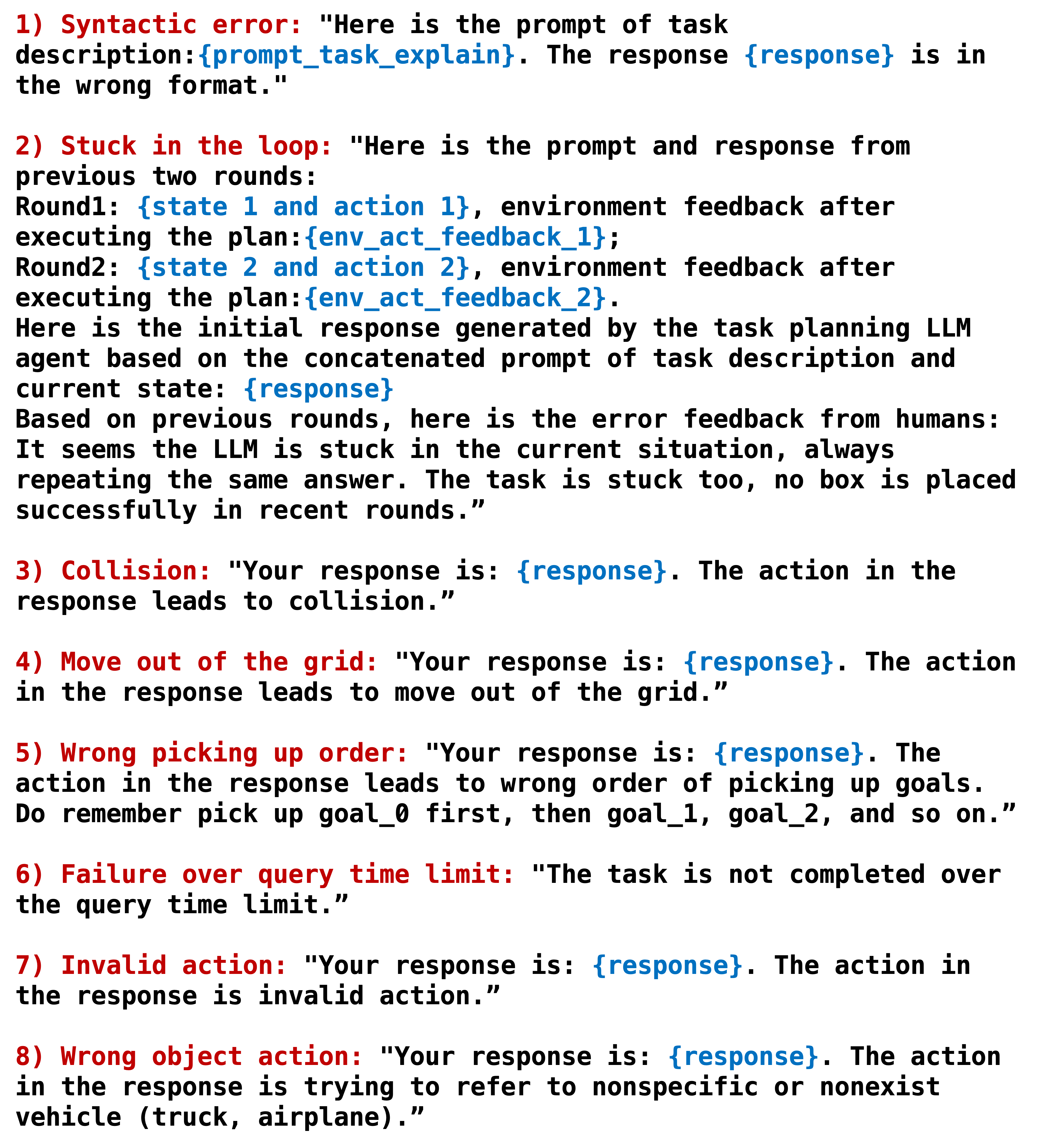}

   \caption{Eight examples of human-designed feedback templates. The blue-colored text represents the content specific to each instance of an error.}
   \label{fig:Human feedback}
\end{figure}

\textbf{Human-Designed Feedback Rules}\quad During task execution, the TaskLLM may encounter an error, resulting in the task being terminated. It is useful for the PromptLLM to have context about this error when generating new prompt candidates. Since automatic error analysis via LLMs is difficult for multi-step tasks (e.g. an agent stuck in an action loop), we instead use human-designed rules to automatically synthesize feedback, as shown in Figure~\ref{fig:Human feedback}. Some types of errors can be common across all tasks (e.g. syntactic errors), while others are task-specific. For types of human feedback in each task, see Appendix~\ref{appendix sec: human feedback types}. 

In Section~\ref{sec:results and analysis}, we do ablation experiments to show that variability over the wording of feedback has little impact on PROMST performance. Hence, PROMST does not require humans to iterate over possible versions of feedback templates via trial-and-error efforts. Additionally, for each task, humans just need to design 2 to 4 new error types since some of the error types are common across tasks. Even for the task of multi-hop question answering, error types like `syntactic errors' and `stuck in the loop' still exist and can be shared with other tasks.

\textbf{Candidate Prompt Generator}\quad We produce new candidate prompts from a parent prompt in two steps: (1) summarizing feedback via an LLM (`SumLLM') and (2) generating new prompt candidates via an LLM provided with the summarized feedback as context (`GenLLM'). In order to encourage exploration over more diverse candidate prompts, we randomly choose 10 instances of feedback. This random selection also likely promotes more frequent errors. Given the selected feedback, SumLLM produces a summary that is included as context to GenLLM for generating new candidates. See Algorithm~\ref{alg:3} in Appendix~\ref{sec:appendix-algorithms} for another description of this process and Appendix~\ref{appendix sec: Meta-prompts} for the meta-prompts used for SumLLM and GenLLM.

\section{Experiments}
\subsection{Environments}
\begin{figure}[ht]
  \centering
   \includegraphics[width=1.0\linewidth]{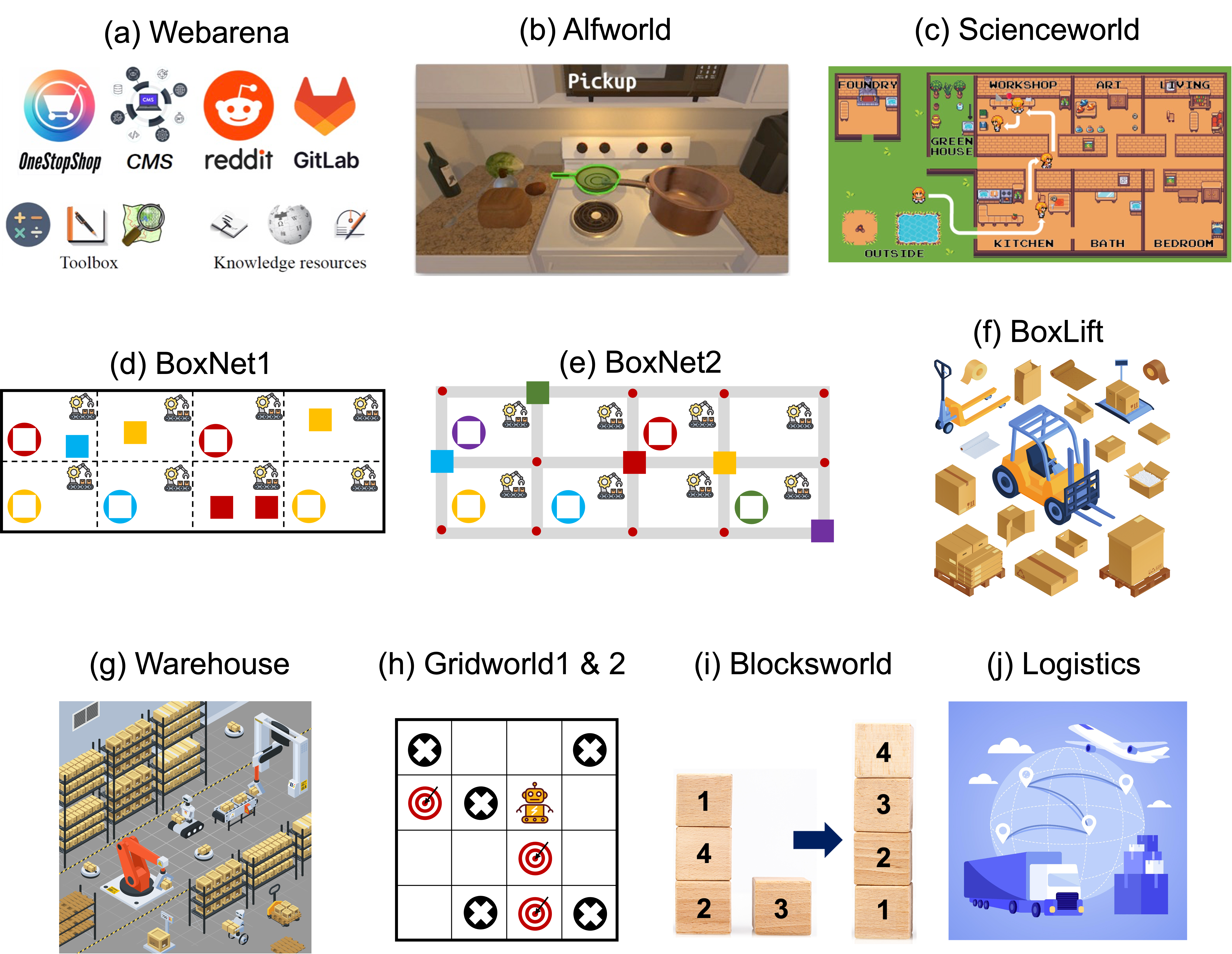}

   \caption{An illustration of the 11 environments used for multi-step task evaluation. See Appendix~\ref{appendix sec: Testing multi-step envs} for more details.}
   \label{fig:Envs}
\end{figure}

As shown in Figure~\ref{fig:Envs}, we test on 11 multi-step tasks requiring strong logical, geometrical, scientific, and commonsense reasoning capabilities \cite{webarena,alfworld,scienceworld,chen2023scalable-multi-agent,Grid-world,planbench}. Each environment requires the LLM agent to determine the next action in the large discrete action space. Please refer to Appendix~\ref{appendix sec: Testing multi-step envs} for a complete description of all tasks.

\subsection{Baselines}
We compare PROMST with six recent representative methods: Automatic Prompt Engineer (APE) \cite{zhou2023large}, Automatic Prompt Optimization (APO) \cite{pryzant2023automatic}, PromptAgent \cite{promptagent}, LLM-As-Optimizer \cite{LLM-as-optimizers}, PromptBreeder \cite{promptbreeder}, and Evolutionary Prompt Optimizer \cite{evolutionary-algorithm-1}. All the methods under comparison involve iterative optimization of prompts. Some methods require error feedback through LLM self-reflection, while others do not. For methods that need error feedback, we randomly select 10 instances of feedback, similar to the PROMST method but without the rules of human feedback. We also compare with the dynamic approach Reflexion \cite{reflexion} by modifying it into an offline framework.

\subsection{Experimental Setups}
For a fair comparison, all methods start the optimization from initial human-designed prompts; where possible, we use the provided publicly available prompts for each method. In all cases, we set the LLM sampling temperature to 0. For each method, we report the score of the best performing prompt on each task; in this case, the score is computed as:

\begin{equation}
S = \text{num}(\text{sub-goal}_{success})/\text{num}(\text{sub-goal}_{all}),
\label{eq:score-func}
\end{equation}

where the score $S$ is the ratio of the number of successfully completed sub-goals/sub-steps to the total number of sub-goals/sub-steps, i.e., the task progress score. Note that the task completion score can also serve as the metric, while it may be sparse in some situations. Both types of scores are positively correlated as shown in Appendix Table~\ref{ablation-table-completion-progress-score}. In Section~\ref{sec:results and analysis}, we also preliminarily test the impact of changing the scoring function $S$.

\textbf{Model Types} One interesting feature of these methods is that the LLM used to execute the task (`TaskLLM') and the LLM used to generate new candidate prompts (`PromptLLM') do not need to be the same model. We mainly test two combinations of models. The first uses GPT-3.5 (gpt-3.5-turbo-16k-0613) as the TaskLLM and GPT-4 (gpt-4-turbo-preview) \cite{gpt} as the PromptLLM, and the second uses GPT-4 for both the TaskLLM and PromptLLM. To verify the effectiveness of PROMST in varied models, we also evaluate it using Claude 3 Opus \cite{claude}, Mixtral-8x7B, and Mixtral-Large \cite{mixtral} as both TaskLLM and PromptLLM. Mixtral-8x7B is an open model, while all the others are closed. We also explore whether the optimized prompts specialized for one type of LLM can generalize better performance to other types of LLMs.

\textbf{Context Window Limit} The constraint of the model's context window is an issue for LLM-based agents, especially for longer multi-step tasks. Relying on the intuition that recency is important, in all the tested methods we use a sliding window of the history of state-action-feedback tuples, truncating the history by pruning older parts of the history that extend beyond the window length, which is a common technique used in LLM agent researches.

\begin{table*}[t]
\caption{Scores for initial (human) and optimized prompts on various multi-step tasks for different methods. P.Agent, LLMOP, P.Breeder, and P.Evolution refer to PromptAgent, LLM-As-Optimizer, PromptBreeder, Evolutionary Prompt Optimizer, respectively. GPT-3.5-0613 for TaskLLM and GPT-4 for PromptLLM.}
\label{result-table-gpt3-GPT-4}
\vskip 0.15in
\begin{center}
\begin{small}
\begin{sc}
\begin{tabular}{lcccccccccr}
\toprule
\multicolumn{1}{c}{} & \multicolumn{8}{c}{\textbf{GPT-3.5-0613-as-TaskLLM, GPT-4-as-PromptLLM}}\\
\multicolumn{1}{c}{Task} & \multicolumn{1}{c}{Human} & \multicolumn{1}{c}{APE} & \multicolumn{1}{c}{APO} & \multicolumn{1}{c}{P.Agent} & \multicolumn{1}{c}{LLMOP} & \multicolumn{1}{c}{P.Breeder} & \multicolumn{1}{c}{P.Evolution} & \multicolumn{1}{c}{PROMST}\\
\midrule
\multicolumn{1}{c}{Webarena} & \multicolumn{1}{c}{0.22} & \multicolumn{1}{c}{0.35} & \multicolumn{1}{c}{0.31} & \multicolumn{1}{c}{0.37} & \multicolumn{1}{c}{0.29} & \multicolumn{1}{c}{0.25} & \multicolumn{1}{c}{0.27} & \multicolumn{1}{c}{\textcolor{magenta}{0.39}} \\
\multicolumn{1}{c}{Alfworld} & \multicolumn{1}{c}{0.075} & \multicolumn{1}{c}{0.24} & \multicolumn{1}{c}{0.23} & \multicolumn{1}{c}{0.24} & \multicolumn{1}{c}{0.14} & \multicolumn{1}{c}{0.12} & \multicolumn{1}{c}{0.16} & \multicolumn{1}{c}{\textcolor{magenta}{0.30}} \\
\multicolumn{1}{c}{Scienceworld} & \multicolumn{1}{c}{0.18} & \multicolumn{1}{c}{0.19} & \multicolumn{1}{c}{0.19} & \multicolumn{1}{c}{\textcolor{magenta}{0.23}} & \multicolumn{1}{c}{0.19} & \multicolumn{1}{c}{0.20} & \multicolumn{1}{c}{0.22} & \multicolumn{1}{c}{0.21} \\
\multicolumn{1}{c}{BoxNet1} & \multicolumn{1}{c}{0.076} & \multicolumn{1}{c}{0.093} & \multicolumn{1}{c}{0.16}  & \multicolumn{1}{c}{0.13} & \multicolumn{1}{c}{0.098} & \multicolumn{1}{c}{0.11} & \multicolumn{1}{c}{0.12} & \multicolumn{1}{c}{\textcolor{magenta}{0.25}} \\
\multicolumn{1}{c}{BoxNet2} & \multicolumn{1}{c}{0.044} & \multicolumn{1}{c}{0.075} & \multicolumn{1}{c}{0.16}  & \multicolumn{1}{c}{0.17} & \multicolumn{1}{c}{0.086} & \multicolumn{1}{c}{0.090} & \multicolumn{1}{c}{0.075} & \multicolumn{1}{c}{\textcolor{magenta}{0.22}}\\
\multicolumn{1}{c}{BoxLift} & \multicolumn{1}{c}{0.31} & \multicolumn{1}{c}{0.69} & \multicolumn{1}{c}{0.70}  & \multicolumn{1}{c}{0.74} & \multicolumn{1}{c}{0.55} & \multicolumn{1}{c}{0.58} & \multicolumn{1}{c}{0.62} & \multicolumn{1}{c}{\textcolor{magenta}{0.90}}\\
\multicolumn{1}{c}{WareHouse} & \multicolumn{1}{c}{0.0} & \multicolumn{1}{c}{0.012} & \multicolumn{1}{c}{0.012} & \multicolumn{1}{c}{\textcolor{magenta}{0.036}} & \multicolumn{1}{c}{0.008} & \multicolumn{1}{c}{0.008} & \multicolumn{1}{c}{0.004} & \multicolumn{1}{c}{0.028}\\
\multicolumn{1}{c}{Gridworld1} & \multicolumn{1}{c}{0.23} & \multicolumn{1}{c}{0.30} & \multicolumn{1}{c}{0.35} & \multicolumn{1}{c}{0.32} & \multicolumn{1}{c}{0.28} & \multicolumn{1}{c}{0.26} & \multicolumn{1}{c}{0.24} & \multicolumn{1}{c}{\textcolor{magenta}{0.38}}\\
\multicolumn{1}{c}{Gridworld2} & \multicolumn{1}{c}{0.036} & \multicolumn{1}{c}{0.093} & \multicolumn{1}{c}{\textcolor{magenta}{0.17}} & \multicolumn{1}{c}{0.15} & \multicolumn{1}{c}{0.065} & \multicolumn{1}{c}{0.078} & \multicolumn{1}{c}{0.13} & \multicolumn{1}{c}{0.12}\\
\multicolumn{1}{c}{Blocksworld} & \multicolumn{1}{c}{0.19} & \multicolumn{1}{c}{0.25} & \multicolumn{1}{c}{0.42} & \multicolumn{1}{c}{0.48} & \multicolumn{1}{c}{0.29} & \multicolumn{1}{c}{0.22} & \multicolumn{1}{c}{0.27} & \multicolumn{1}{c}{\textcolor{magenta}{0.60}}\\
\multicolumn{1}{c}{Logistics} & \multicolumn{1}{c}{0.083} & \multicolumn{1}{c}{0.083} & \multicolumn{1}{c}{0.12} & \multicolumn{1}{c}{0.12} & \multicolumn{1}{c}{0.083} & \multicolumn{1}{c}{0.083} & \multicolumn{1}{c}{0.12} & \multicolumn{1}{c}{\textcolor{magenta}{0.18}}\\
\multicolumn{1}{c}{\textbf{Average}} & \multicolumn{1}{c}{0.13} & \multicolumn{1}{c}{0.22} & \multicolumn{1}{c}{0.26} & \multicolumn{1}{c}{0.27} & \multicolumn{1}{c}{0.19} & \multicolumn{1}{c}{0.18} & \multicolumn{1}{c}{0.20} & \multicolumn{1}{c}{\textcolor{magenta}{0.32}}\\

\bottomrule
\end{tabular}
\end{sc}
\end{small}
\end{center}
\vskip -0.1in
\end{table*}

\begin{table*}[t]
\caption{Scores for initial (human) and optimized prompts on various multi-step tasks for different methods. GPT-4 for both TaskLLM and PromptLLM.}
\label{result-table-GPT-4-GPT-4}
\vskip 0.15in
\begin{center}
\begin{small}
\begin{sc}
\begin{tabular}{lcccccccccr}
\toprule
\multicolumn{1}{c}{} & \multicolumn{8}{c}{\textbf{GPT-4-as-TaskLLM, GPT-4-as-PromptLLM}}\\
\multicolumn{1}{c}{Task} & \multicolumn{1}{c}{Human} & \multicolumn{1}{c}{APE} & \multicolumn{1}{c}{APO} & \multicolumn{1}{c}{P.Agent} & \multicolumn{1}{c}{LLMOP} & \multicolumn{1}{c}{P.Breeder} & \multicolumn{1}{c}{P.Evolution} & \multicolumn{1}{c}{PROMST}\\
\midrule
\multicolumn{1}{c}{Webarena} & \multicolumn{1}{c}{0.57} & \multicolumn{1}{c}{0.59} & \multicolumn{1}{c}{\textcolor{magenta}{0.64}} & \multicolumn{1}{c}{0.60} & \multicolumn{1}{c}{0.58} & \multicolumn{1}{c}{0.58} & \multicolumn{1}{c}{0.59} & \multicolumn{1}{c}{0.62} \\
\multicolumn{1}{c}{Alfworld} & \multicolumn{1}{c}{0.45} & \multicolumn{1}{c}{0.49} & \multicolumn{1}{c}{0.50} & \multicolumn{1}{c}{0.53} & \multicolumn{1}{c}{0.50} & \multicolumn{1}{c}{0.47} & \multicolumn{1}{c}{0.49} & \multicolumn{1}{c}{\textcolor{magenta}{0.57}} \\
\multicolumn{1}{c}{Scienceworld} & \multicolumn{1}{c}{0.70} & \multicolumn{1}{c}{0.72} & \multicolumn{1}{c}{0.74} & \multicolumn{1}{c}{0.76} & \multicolumn{1}{c}{0.71} & \multicolumn{1}{c}{0.73} & \multicolumn{1}{c}{0.76} & \multicolumn{1}{c}{\textcolor{magenta}{0.81}} \\
\multicolumn{1}{c}{BoxNet1} & \multicolumn{1}{c}{0.65} & \multicolumn{1}{c}{0.72} & \multicolumn{1}{c}{0.72}  & \multicolumn{1}{c}{0.77} & \multicolumn{1}{c}{0.74} & \multicolumn{1}{c}{0.67} & \multicolumn{1}{c}{0.70} & \multicolumn{1}{c}{\textcolor{magenta}{0.79}} \\
\multicolumn{1}{c}{BoxNet2} & \multicolumn{1}{c}{0.34} & \multicolumn{1}{c}{0.38} & \multicolumn{1}{c}{0.36}  & \multicolumn{1}{c}{0.35} & \multicolumn{1}{c}{0.40} & \multicolumn{1}{c}{0.37} & \multicolumn{1}{c}{0.40} & \multicolumn{1}{c}{\textcolor{magenta}{0.42}}\\
\multicolumn{1}{c}{WareHouse} & \multicolumn{1}{c}{0.16} & \multicolumn{1}{c}{0.18} & \multicolumn{1}{c}{0.27} & \multicolumn{1}{c}{0.34} & \multicolumn{1}{c}{0.30} & \multicolumn{1}{c}{0.25} & \multicolumn{1}{c}{0.22} & \multicolumn{1}{c}{\textcolor{magenta}{0.51}}\\
\multicolumn{1}{c}{Gridworld1} & \multicolumn{1}{c}{0.73} & \multicolumn{1}{c}{0.78} & \multicolumn{1}{c}{0.82} & \multicolumn{1}{c}{\textcolor{magenta}{0.89}} & \multicolumn{1}{c}{0.83} & \multicolumn{1}{c}{0.76} & \multicolumn{1}{c}{0.80} & \multicolumn{1}{c}{0.86}\\
\multicolumn{1}{c}{Gridworld2} & \multicolumn{1}{c}{0.26} & \multicolumn{1}{c}{0.50} & \multicolumn{1}{c}{0.44} & \multicolumn{1}{c}{0.41} & \multicolumn{1}{c}{0.41} & \multicolumn{1}{c}{0.31} & \multicolumn{1}{c}{0.29} & \multicolumn{1}{c}{\textcolor{magenta}{0.60}}\\
\multicolumn{1}{c}{Blocksworld} & \multicolumn{1}{c}{0.71} & \multicolumn{1}{c}{0.74} & \multicolumn{1}{c}{0.83} & \multicolumn{1}{c}{0.87} & \multicolumn{1}{c}{0.76} & \multicolumn{1}{c}{0.75} & \multicolumn{1}{c}{0.77} & \multicolumn{1}{c}{\textcolor{magenta}{0.95}}\\
\multicolumn{1}{c}{Logistics} & \multicolumn{1}{c}{0.50} & \multicolumn{1}{c}{0.53} & \multicolumn{1}{c}{0.58} & \multicolumn{1}{c}{0.61} & \multicolumn{1}{c}{0.54} & \multicolumn{1}{c}{0.53} & \multicolumn{1}{c}{0.56} & \multicolumn{1}{c}{\textcolor{magenta}{0.74}}\\
\multicolumn{1}{c}{\textbf{Average}} & \multicolumn{1}{c}{0.51} & \multicolumn{1}{c}{0.56} & \multicolumn{1}{c}{0.59} & \multicolumn{1}{c}{0.61} & \multicolumn{1}{c}{0.58} & \multicolumn{1}{c}{0.54} & \multicolumn{1}{c}{0.56} & \multicolumn{1}{c}{\textcolor{magenta}{0.69}}\\

\bottomrule
\end{tabular}
\end{sc}
\end{small}
\end{center}
\vskip -0.1in
\end{table*}

\begin{table*}[t]
\caption{Evaluation of different LLMs on a subset of the mutli-step tasks. The same LLM was used for the TaskLLM and PromptLLM in each case.}
\label{result-table-claude-mixtral}
\vskip 0.15in
\begin{center}
\begin{small}
\begin{sc}
\begin{tabular}{lcccccccccr}
\toprule
\multicolumn{1}{c}{} & \multicolumn{4}{c|}{\textbf{Claude-3-opus-20240229}} & \multicolumn{4}{c}{\textbf{Open-Mixtral-8x7b}}\\
\multicolumn{1}{c}{Task} & \multicolumn{1}{c}{Human} & \multicolumn{1}{c}{APO} & \multicolumn{1}{c}{P.Agent} & \multicolumn{1}{c|}{PROMST} & \multicolumn{1}{c}{Human} & \multicolumn{1}{c}{APO} & \multicolumn{1}{c}{P.Agent} & \multicolumn{1}{c}{PROMST}\\
\midrule
\multicolumn{1}{c}{Alfworld} & \multicolumn{1}{c}{0.32} & \multicolumn{1}{c}{0.40} & \multicolumn{1}{c}{0.42} & \multicolumn{1}{c|}{\textcolor{magenta}{0.49}} & \multicolumn{1}{c}{0.055} & \multicolumn{1}{c}{0.074} & \multicolumn{1}{c}{0.071} & \multicolumn{1}{c}{\textcolor{magenta}{0.10}}\\
\multicolumn{1}{c}{BoxNet2} & \multicolumn{1}{c}{0.42} & \multicolumn{1}{c}{0.47} & \multicolumn{1}{c}{0.44} & \multicolumn{1}{c|}{\textcolor{magenta}{0.53}} & \multicolumn{1}{c}{0.078} & \multicolumn{1}{c}{0.21} & \multicolumn{1}{c}{0.20} & \multicolumn{1}{c}{\textcolor{magenta}{0.24}}\\
\multicolumn{1}{c}{WareHouse} & \multicolumn{1}{c}{0.21} & \multicolumn{1}{c}{0.26} & \multicolumn{1}{c}{0.27} & \multicolumn{1}{c|}{\textcolor{magenta}{0.36}} & \multicolumn{1}{c}{0.020} & \multicolumn{1}{c}{0.093} & \multicolumn{1}{c}{0.13} & \multicolumn{1}{c}{\textcolor{magenta}{0.16}}\\
\multicolumn{1}{c}{Gridworld2} & \multicolumn{1}{c}{0.13} & \multicolumn{1}{c}{0.29} & \multicolumn{1}{c}{0.34} & \multicolumn{1}{c|}{\textcolor{magenta}{0.44}} & \multicolumn{1}{c}{0.013} & \multicolumn{1}{c}{0.038} & \multicolumn{1}{c}{0.075} & \multicolumn{1}{c}{\textcolor{magenta}{0.10}} \\
\multicolumn{1}{c}{\textbf{Average}} & \multicolumn{1}{c}{0.27} & \multicolumn{1}{c}{0.36} &  \multicolumn{1}{c}{0.37} & \multicolumn{1}{c|}{\textcolor{magenta}{0.46}} & \multicolumn{1}{c}{0.041} & \multicolumn{1}{c}{0.10} & \multicolumn{1}{c}{0.12} & \multicolumn{1}{c}{\textcolor{magenta}{0.15}}\\

\bottomrule
\end{tabular}
\end{sc}
\end{small}
\end{center}
\vskip -0.1in
\end{table*}

\begin{table}[t]
\label{result-table-modified4}
\vskip 0.15in
\begin{center}
\begin{small}
\begin{sc}
\begin{tabular}{lccccr}
\toprule
\multicolumn{4}{c}{\textbf{Mixtral-large-2402}}\\
\multicolumn{1}{c}{Human} & \multicolumn{1}{c}{APO} & \multicolumn{1}{c}{P.Agent} & \multicolumn{1}{c}{Promst}\\
\midrule
\multicolumn{1}{c}{0.28} & \multicolumn{1}{c}{0.33} & \multicolumn{1}{c}{0.33} & \multicolumn{1}{c}{\textcolor{magenta}{0.45}} \\
\multicolumn{1}{c}{0.26} & \multicolumn{1}{c}{0.31} & \multicolumn{1}{c}{\textcolor{magenta}{0.36}} & \multicolumn{1}{c}{0.30}\\
\multicolumn{1}{c}{0.16} & \multicolumn{1}{c}{0.23} & \multicolumn{1}{c}{0.28} & \multicolumn{1}{c}{\textcolor{magenta}{0.31}}\\
\multicolumn{1}{c}{0.12} & \multicolumn{1}{c}{\textcolor{magenta}{0.32}} &  \multicolumn{1}{c}{0.25} & \multicolumn{1}{c}{0.28}\\
\multicolumn{1}{c}{0.21} & \multicolumn{1}{c}{0.30} &  \multicolumn{1}{c}{0.30} & \multicolumn{1}{c}{\textcolor{magenta}{0.34}}\\

\bottomrule
\end{tabular}
\end{sc}
\end{small}
\end{center}
\vskip -0.1in
\end{table}

\textbf{Hyperparameters} For a fair comparison, we standardize all hyperparameters across methods, allowing each to explore the same number of prompt candidates at an equivalent level. The expansion number $n$ controls the number of kid prompts generated based on each parent prompt, which is set to 20 in the first level and 8 for all additional levels. In each level, top $k=5$ current prompts are selected as the parent prompts for further optimization. Search terminates once the recent three levels do not have any score improvements. In PROMST, we set hyper\_M = 0.8 in Equation~\ref{alg:3} to filter out prompts with low scores. We set $sd = 4$ so that the score model is not applied until $4^{th}$ generation.

\subsection{Results and Analysis}
\label{sec:results and analysis}
\textbf{Overall Better Performance}\quad Table~\ref{result-table-gpt3-GPT-4} and Table~\ref{result-table-GPT-4-GPT-4} show the main experimental results. Note that BoxLift task is not included in Table~\ref{result-table-GPT-4-GPT-4} since we find GPT-4 can already achieve a full score with the initial human prompt. Table~\ref{result-table-claude-mixtral} shows the experimental results on other three types of LLMs. Due to limited computational resources, we only select four representative tasks and two strongest baseline methods (APO, PromptAgent) when evaluating other LLMs. Table~\ref{result-table-generalization1} and Table~\ref{result-table-generalization2} in Appendix~\ref{appendix sec: Generalization to different models for optimized prompts} test the performance of the optimized prompts trained from one LLM with other types of TaskLLMs.

The main takeaways are: 1) PROMST performs the best in most tasks. On average, PROMST outperforms strongest baseline PromptAgent with GPT-3.5-0613 (0.27 vs 0.32), GPT-4 (0.61 vs 0.69), Claude-3-opus (0.36 vs 0.46), Open-Mixtral-8x7B (0.12 vs 0.15), and Mixtral-large (0.30 vs 0.34). 2) When testing the best prompts trained from GPT-3.5-0613 and GPT-4 with a different TaskLLM, we find that they still outperform human prompts. 3) However, each LLM does best with the prompts optimized on it. For example, the best prompts acquired when using GPT-3.5-0613 as the TaskLLM do not further improve performance when applied to GPT-4, and vice versa. 4) PROMST performs well when the TaskLLM and PromptLLM are the same LLM, showing that it does not rely on a stronger PromptLLM to pass extra knowledge into prompts, which can be regarded as cheating.

\begin{figure}[ht]
  \centering
   \includegraphics[width=1.0\linewidth]{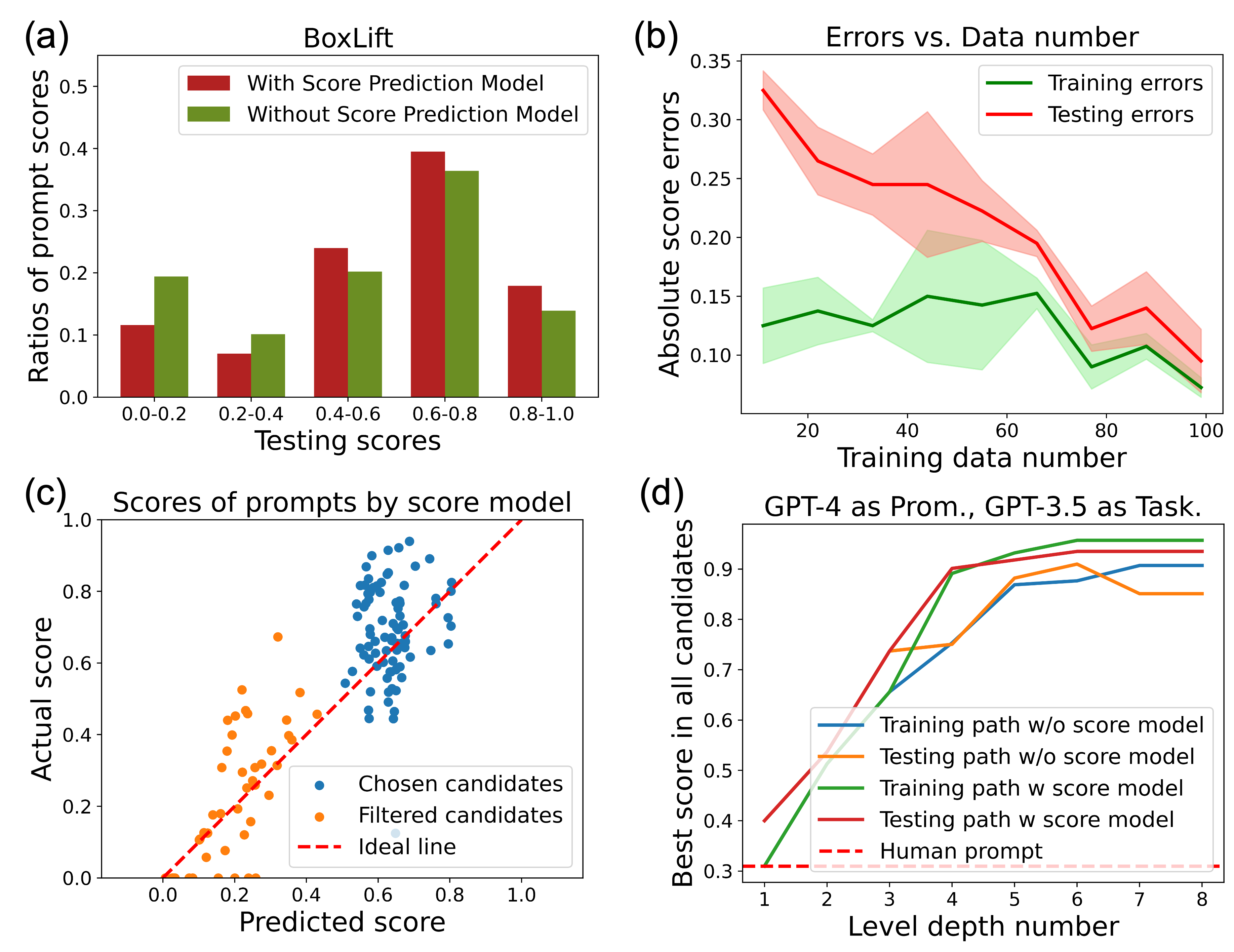}

   \caption{Several results inspecting the learned score prediction model. (a) The distribution/ratio of prompt scores with/without the score prediction model. (b) The prediction error of the model on the training data and heldout test data as the amount of training data increases. (c) A plot of the predicted score vs the actual score for various prompts; blue are the prompts that were chosen as parents for new candidates. (d) The trend of the best performing prompt during optimization for increasing iterations both with and without using the learned score prediction model.}
   \label{fig:Score model}
\end{figure}

\begin{figure}[ht]
  \centering
   \includegraphics[width=1.0\linewidth]{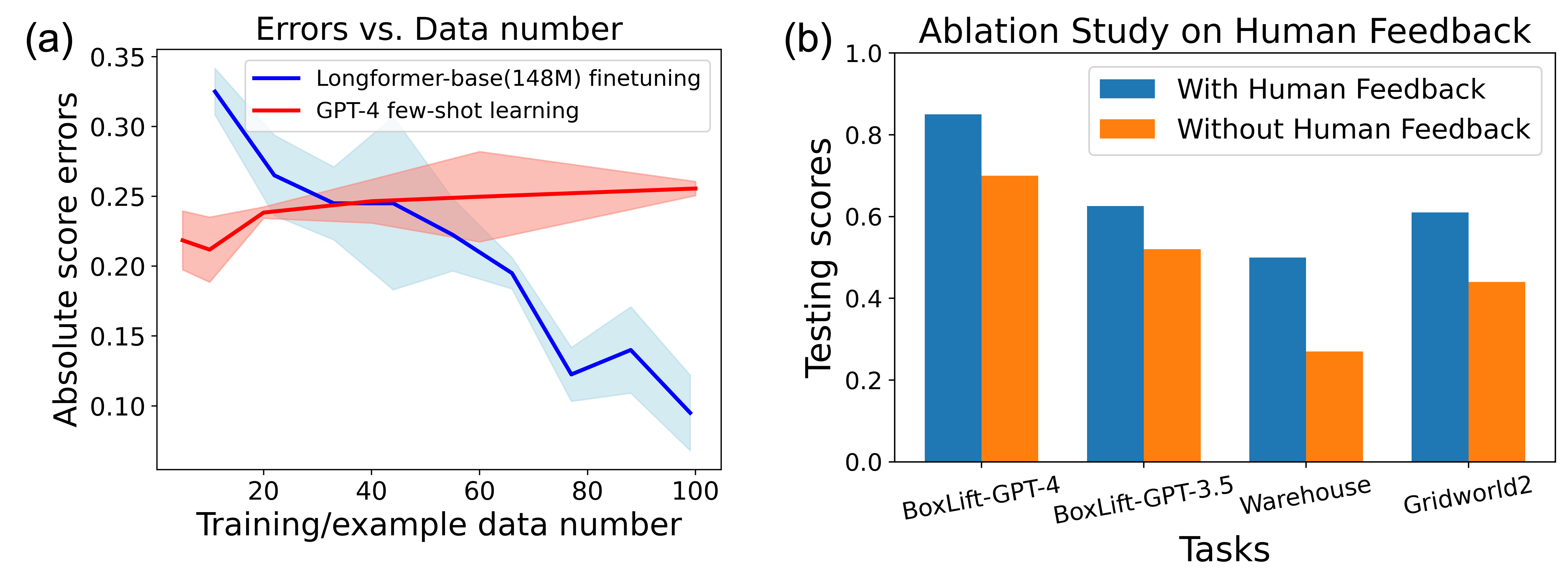}

   \caption{(a) Comparison of score prediction errors for few-shot GPT-4 vs finetuning Longformer for increasing amount of few-shot examples or training data, respectively. (b) An ablation study of the impact of the human-designed feedback rules on task performance for four multi-step tasks.}
   \label{fig:GPT-3.5GPT-4-few-shot}
\end{figure}

\textbf{Effects of Score Model}\quad To analyze the effects of the score model, we use BoxLift as a representative example, as shown in Figure~\ref{fig:Score model}. Figure~\ref{fig:Score model}a shows the distribution of prompt scores explored in all the levels (1-8) with and without the score prediction model implemented, respectively. The implementation of the score prediction model truly makes the exploration more efficient since less low-scored prompts are explored. Figure~\ref{fig:Score model}b shows the training and testing errors of the score model versus different amounts of collected training data. The overfitting effect decreases with increasing data number. Figure~\ref{fig:Score model}c tests the fine-tuned score models on levels 5-8.

We also evaluate the prompts that were filtered out by the score model and plot the predicted and actual scores. We find that nearly all chosen prompt candidates achieve scores higher than 0.4, and the filtered prompts have reliably low scores. We compare the evolution curves for PROMST with/without the score model, as shown in  Figure~\ref{fig:Score model}d. The results show that both the training and testing paths converge faster and achieve better scores using the score model. The ablation experiments in other environments also have the same trend (shown in Appendix~\ref{appendix sec: Extra abalation experiments of score models}). Overall, we find the score prediction model improves the efficiency and effectiveness of prompt search. For each prompt optimization process of PROMST, usually the best prompt appears at the iteration number 5 to 7, as visualized in Figure~\ref{fig:Score model}d and Appendix~\ref{appendix sec: Extra abalation experiments of score models}.

\textbf{Ablation on Methods of Score Model}\quad Instead of fine-tuning a pre-trained Longformer-base model, another way to acquire score prediction models is few-shot learning via GPT-4. Figure~\ref{fig:GPT-3.5GPT-4-few-shot}a compares these two methods under varied training/example data number. GPT-4 is given randomly selected prompt-score pairs as examples during the study. We find that the performance of GPT-4 few-shot learning cannot improve with the increasing number of examples. The fine-tuning method surpasses GPT-4 few-shot learning once the data number increases over 40.

\begin{table}[t]
\caption{Ablation studies on SumLLM.}
\label{table:ablation-table-SumLLM}
\vskip 0.15in
\begin{center}
\begin{small}
\begin{sc}
\begin{tabular}{lcccr}
\toprule
\multicolumn{3}{c}{\textbf{GPT-3.5-as-TaskLLM, GPT-4-as-PromptLLM}}\\
\multicolumn{1}{c}{Task} & \multicolumn{1}{c}{W SumLLM} & \multicolumn{1}{c}{WO SumLLM}\\
\midrule
\multicolumn{1}{c}{Alfworld} & \multicolumn{1}{c}{\textcolor{magenta}{0.30}} & \multicolumn{1}{c}{0.23} \\
\multicolumn{1}{c}{BoxNet2} & \multicolumn{1}{c}{\textcolor{magenta}{0.22}} & \multicolumn{1}{c}{0.18} \\
\multicolumn{1}{c}{BoxLift} & \multicolumn{1}{c}{\textcolor{magenta}{0.90}} & \multicolumn{1}{c}{0.85}\\
\bottomrule
\end{tabular}
\end{sc}
\end{small}
\end{center}
\vskip -0.1in
\end{table}

\textbf{Ablation on SumLLM Component}\quad Since TaskLLM and GenLLM are necessary in the whole framework, we compare PROMST with/without SumLLM component in Table~\ref{table:ablation-table-SumLLM} to verify its effectiveness. The integration of SumLLM improves the performance on all the three representative tasks.

\begin{table*}[t]
\caption{Ablation studies on the templates used for human feedback.}
\label{table:ablation-table-human-feedback}
\vskip 0.15in
\begin{center}
\begin{small}
\begin{sc}
\begin{tabular}{lccccccr}
\toprule
\multicolumn{6}{c}{\textbf{GPT-3.5-as-TaskLLM, GPT-4-as-PromptLLM}}\\
\multicolumn{1}{c}{Task} & \multicolumn{1}{c}{Original} & \multicolumn{1}{c}{Paraphrased} & \multicolumn{1}{c}{Random} & \multicolumn{1}{c}{WO response component} & \multicolumn{1}{c}{WO stuck in loop}\\
\midrule
\multicolumn{1}{c}{BoxNet2} & \multicolumn{1}{c}{0.22} & \multicolumn{1}{c}{0.20} & \multicolumn{1}{c}{\textcolor{magenta}{0.23}} & \multicolumn{1}{c}{0.17} & \multicolumn{1}{c}{0.18}\\
\multicolumn{1}{c}{BoxLift} & \multicolumn{1}{c}{0.90} & \multicolumn{1}{c}{0.93} & \multicolumn{1}{c}{\textcolor{magenta}{0.97}} & \multicolumn{1}{c}{0.73} & \multicolumn{1}{c}{0.87}\\
\bottomrule
\end{tabular}
\end{sc}
\end{small}
\end{center}
\vskip -0.1in
\end{table*}

\textbf{Ablation on Human Feedback}\quad We compare the method with/without human feedback, both without the learned score model. As seen in Figure~\ref{fig:GPT-3.5GPT-4-few-shot}b, human feedback contributes to much higher scores across four tasks. In our work, the original human feedback templates did not require iterations via trial-and-error over possible versions. To demonstrate that designing human feedback rules is straightforward and requires minimal efforts, we test other four feedback templates in Table~\ref{table:ablation-table-human-feedback}. The results show that variability over the wording of the templates has little impact on the performance of PROMST. Thus, including the response in the feedback template is a useful task- and error-agnostic guiding principle. Appendix~\ref{appendix sec: Efforts for designing human feedback rules} articulates more specifically on the designing of four compared human feedback templates and the reasons why designing feedback rules is effortless.

\textbf{Preference Alignment via Score Function}\quad The choice of score functions impacts prompt optimization, in which humans may have different preferences for the same task. In Appendix~\ref{appendix sec: The influence of score functions}, we explore the impacts of varied score functions and find that PROMST can well align with human preferences by modifying score function formats.

\textbf{Explanability for Better Prompts}\quad We also try to dig out some mechanisms why the optimized prompts are better. In Figure~\ref{fig:Appendix-visua-token-len}, we plot prompt score vs. token length and perplexity, which implies some clues that longer prompts may be better. Meanwhile, when viewing through the discovered best prompts in Appendix~\ref{appendix sec: all prompts}, we find some clues about better component emergence, i.e., the best prompts tend to list all the careful points one by one clearly. We conduct an ablation study by summarizing detailed careful points into varying token lengths using GPT-4 and evaluating their performance. The results indicate that task scores consistently decline as token lengths decrease, underscoring the importance of clearly listing detailed points. More specific discussion is shown in Appendix~\ref{appendix sec: Explanability for better prompts}.

\textbf{Comparison and Combination with Reflexion}\quad In Appendix~\ref{appendix: Comparison and Combination with Dynamic Approach} we find that PROMST outperforms the dynamic approach, Reflexion, in prompt optimization and achieves enhanced performance when Reflexion is integrated in a multi-trial setting.

\section{Conclusion}
In this work we introduce an automatic prompt optimization framework for complex, multi-step agent tasks: PROMST. To handle the issues of task complexity, judging long-horizon correctness of individual actions, high prompt exploration cost, and human preference alignment, we propose the integration of human feedback, a learned score prediction model, and the modification of task score functions. Our approach generally outperforms six representative baselines on 11 different task environments over all the five LLMs. PROMST is orthogonal and combinatorial to existing dynamic approaches. The discovered best prompts have some inspiring characteristics for better performance.

\section{Limitations}
The limitations and potential societal risks of this work are as follows:\\

\textbf{Huge resource consumption of API calls}\quad Automatic prompt optimization requires significant computing resources and LLM API queries due to its search-based nature, which is a common issue in this research track. Though the introduction of score model makes the searching more efficient, the around 100 prompt candidate exploration is still a large burden.\\

\textbf{Score model increases computing demands of local devices}\quad The fine-tuned score prediction model trades-off the number of API queries for on-device computation by selecting good candidate prompts. Still, the training of extra score models increases the computing demands on local devices.\\

\textbf{Extra burden of designing human feedback rules}\quad Introducing human feedback into prompt optimization is a natural way because the current LLM cannot well summarize and reflect errors in multi-step tasks, as shown in ablation studies. Integrating human feedback is inspired by the phenomenon that humans are good at providing feedback about LLM errors but struggle to optimize prompts. We admit that requiring human pre-defined feedback will somewhat increase the burden of users and that introducing automatic feedback is one important work to be explored in the future.\\

\textbf{Fine-tuning score model requires enough data points}\quad The fine-tuning process of score models typically requires around 100 prompt-score pairs, which is suitable for black box prompt searching since over 100 data points are truly needed for satisfying performance. However, the score model may not be suitable if in the future a more efficient searching method appears so that data points are not such much.\\

\section*{Acknowledgements}

This work was supported by ONR under Award N00014-22-1-2478 and MIT-IBM Watson AI Lab. However, this article solely reflects the opinions and conclusions of its authors.

\bibliography{custom}
\appendix
\onecolumn

\newpage
\tableofcontents 

\newpage
\section{Types of human feedback for each task}
\label{appendix sec: human feedback types}
This table displays the types of errors and corresponding human feedback for each testing task. The specific contents of each feedback is shown in Figure~\ref{fig:Human feedback}.

\begin{table}[ht]
\centering
\begin{tabular}{p{0.15\linewidth} | p{0.8\linewidth}}
\hline
Webarena & Syntactic error; Stuck in the loop; Failure over query time limit; Invalid action\\
\hline
Alfworld & Syntactic error; Stuck in the loop; Failure over query time limit; Invalid action\\
\hline
Scienceworld & Syntactic error; Stuck in the loop; Failure over query time limit; Invalid action\\
\hline
BoxNet1 & Syntactic error; Stuck in the loop; Failure over query time limit\\
\hline
BoxNet2 & Syntactic error; Stuck in the loop; Failure over query time limit; Collision\\
\hline
BoxLift & Syntactic error; Stuck in the loop; Failure over query time limit\\
\hline
Warehouse & Syntactic error; Stuck in the loop; Failure over query time limit; Collision\\
\hline
Gridworld1 & Syntactic error; Stuck in the loop; Failure over query time limit; Collision; Move out of the grid;\\
\hline
Gridworld2 & Syntactic error; Stuck in the loop; Failure over query time limit; Collision; Move out of the grid; Wrong picking up order;\\
\hline
Blocksworld & Syntactic error; Stuck in the loop; Failure over query time limit; Invalid action\\
\hline
Logistics & Syntactic error; Stuck in the loop; Failure over query time limit; Invalid action; Wrong object action\\
\hline
\end{tabular}
\caption{Types of human feedback for each task}
\label{tab:Types of human feedback for each task}
\end{table}

\section{Algorithms \label{sec:appendix-algorithms}}
\begin{algorithm}
\caption{PRompt Optimization in Multi-Step Tasks (PROMST)\label{alg:PROMST}}
\begin{algorithmic}[1]
\Require $B_{i}$: list of best prompts in level i; $D$: dictionary of prompts recording corresponding scores, human feedbacks, and ancestor prompts; $feed\_rules$: human pre-defined feedback rules; $p_0$: initial prompt; $k$: beam width; $d$: search level depth; $n$: expansion number; $sd$: depth when score model training starts
\State $B_0 \gets \{p_0\}$; $D \gets \{\}$
\State $D[p_0] \gets feed\_rules(TaskLLM(p_0))$ \textcolor{blue}{\Comment{[prompt, score, feed, AnP]}}
\For{$i \gets 1$ \textbf{to} $d-1$}
    \ForAll{$p \in B_i$}
        \State $P_{new} \gets NewPrompt(p,D,i,sd,n)$
        \ForAll{$p_{new} \in P_{new}$}
        \State $D[p_{new}] \gets feed\_rules(TaskLLM(p_{new}))$ \textcolor{blue}{\Comment{Dictionary to record all prompt trials and information}}
        \EndFor
    \EndFor
    \State $B_{i+1} \gets \text{Top}_k(D)$ \textcolor{blue}{\Comment{Select top $k$ prompts by scores}}
    \State \textbf{Output} $B_{i+1}$
\EndFor
\State $\hat{p} \gets \text{argmax}_{p\in B_d} score(p)$ \textcolor{blue}{\Comment{The best prompt}}
\State \textbf{Output} $\hat{p}$
\State \textbf{return} $\hat{p}$
\end{algorithmic}
\end{algorithm}

\begin{algorithm}
\caption{NewPrompt() - line 5 of Algorithm 1\label{alg:2}}
\begin{algorithmic}[1]
\Require $p$: input prompt; $D$: dictionary of all prompts; $i$: current depth level; $sd$: depth level number when score model training starts; $n$: expansion number
\State $[feed, AnP] \gets D[p]$ \textcolor{blue}{\Comment{Human feedback and ancestor prompts (prompt trajectory leading to the current one)}}
\If{$i \geq sd$}
    \State $M_k \gets finetune(D), k = 1,...,5$ \textcolor{blue}{\Comment{Score model, input: prompt, output: score}}
    \State $P_{new} \gets \{\}$
    \State $iter \gets 0$
    \While{len($P_{new}$) $< n$ and $iter < 3n$}
        \State $iter \mathrel{+}= 1$
        \State $p' = PromptLLM(p, feed, AnP, 1)$
        \If{$E[M_k(p')] + Var[M_k(p')] + E[error_k]\geq hyper\_M*\max(D.score())$}
        \State $P_{new}$.add($p'$)
        \EndIf
    \EndWhile
    \State \textbf{return} $P_{new}$
\Else
    \State $\{p'_1, ..., p'_n\} = PromptLLM(p, feed, AnP, n)$
    \State \textbf{return} $\{p_1', ..., p_m'\}$
\EndIf
\end{algorithmic}
\end{algorithm}

\begin{algorithm}
\caption{PromptLLM() - line 8 and 15 of Algorithm 2\label{alg:3}}
\begin{algorithmic}[1]
\Require $p$: input prompt; $feed$: list of human feedbacks; $AnP$: list of prompt trajectory leading to the current one; $n$: expansion number
\State $P_{new} \gets \{\}$
\For{$i \gets 1$ \textbf{to} $n$}
    \State $feed2 \gets random\_select(feed, \min(10, len(feed))$
    \State $\{type1, type2,...\} \gets classify\_concat(feed2)$
    \State $feed2LLM = '$
    \ForAll{$feed\_type \in \{type1, type2,...\}$}
    \State $feed2LLM \mathrel{+}= SumLLM(p, feed\_type)$ \textcolor{blue}{\Comment{Summarize each type of feedback}}
    \EndFor
    \State $P_{new}$.add($GenLLM(p, feed2LLM, AnP)$)
\EndFor
\State \textbf{return} $P_{new}$
\end{algorithmic}
\end{algorithm}

\newpage
\section{Meta-prompts of $SumLLM$ and $GenLLM$}
\label{appendix sec: Meta-prompts}
\begin{boxL}
\textbf{Meta-prompt of $SumLLM$}\\
Imagine you are a prompt optimizer based on the human feedback and task execution feedback. I’m writing prompts for a language model designed for a task.\\

My current prompt of task specification is: \{current\_prompt\}, but this prompt gets the following examples wrong: \{feedback\_type\}\\

Based on all these errors and feedback, summarize the reasons and list all the aspects that can improve the prompt. Keep your summary concise and clear.
\end{boxL}

\begin{boxL}
\textbf{Meta-prompt of $GenLLM$}\\
Imagine you are a prompt optimizer based on the feedback from the human and task execution feedback. Here is the prompt of task description: \{prompt\_task\_explain\}\\

However, the response generated from the initial task description prompt owns some errors. Here are the error feedback from humans: \{error\_feedback\}\\

There is a list of former prompts including the current prompt, and each prompt is modified from its former prompts:\{trajectory\_prompts\}\\

Based on the feedback, think about why the task planning LLM agent makes the error and try to optimize the prompt of task description to avoid this error.\\

The new prompts should follow these guidelines: 1. The new prompts should solve the current prompt’s problems. 2. The new prompts should consider the list of prompts and evolve based on the current prompt.\\

Output the optimized prompt of task description without other texts:
\end{boxL}

\newpage
\section{Description of environments for multi-step tasks}
\label{appendix sec: Testing multi-step envs}
Here we describe the 11 environments for multi-step tasks on which the various methods were tested. They require strong logical, geometrical, scientific, and commonsense reasoning capabilities. \\

\textbf{Webarena}\quad Webarena (Figure~\ref{fig:Envs}a) is a real web environment containing four applications: online shopping, discussion forums, collaborative development, and business content management. It supports 11 different web browsing actions. The observation space consists of structured web content. WebArena offers multi-round and continuous web browsing interaction simulation.\\

\textbf{Alfworld}\quad Alfworld (Figure~\ref{fig:Envs}b) are Household tasks that require models to explore rooms and use commonsense reasoning to perform tasks, such as “put a pencil on the desk”. The execution scores are calculated by pre-defined subgoals based on necessary observations to finish a task and the success flag provided by environments.\\

\textbf{Scienceworld}\quad Scienceworld (Figure~\ref{fig:Envs}c) is a complex interactive text environment that poses a significant challenge to agents’ scientific commonsense. This environment requires agents to navigate through eight distinct functional rooms (e.g., workshop, kitchen) and utilize the tools to complete tasks such as “measure the melting point of the orange juice”.\\

\textbf{BoxNet1}\quad BoxNet1 (Figure~\ref{fig:Envs}d) consists of robot arms, colored boxes (squares), and colored goal locations (circles). Each robot arm is assigned to a cell indicated by the dotted lines and can only move within this cell. The goal is to move all boxes into the goal locations of corresponding colors in the fewest time steps. Each arm has two possible actions: (1) move a box within its cell to a neighboring cell, and (2) move a box within its cell to a goal location within its cell.\\

\textbf{BoxNet2}\quad BoxNet2 (Figure~\ref{fig:Envs}e) is similar to BoxNet1 but has an additional constraint. In BoxNet2, boxes can only be moved between cells by being placed at the corners of cells (indicated by the small red circles), and each cell corner can only hold one box at a time. Each arm has two possible actions: (1) move a box from a corner to a different corner of its cell, and (2) move a box from a corner to a goal location within its cell.\\

\textbf{BoxLift}\quad BoxLift (Figure~\ref{fig:Envs}f) consists of robots of different types and boxes of different sizes and weights. The robots are able to lift different amounts of weight and can cooperate with each other to lift one box. A box will be lifted only if the total lifting capability of robots is greater than the box’s weight. The goal is to lift all boxes in fewest time steps. Further, the LLM agent can only observe the size of each box, not its actual weight. The weight of a box is roughly proportional to its size (with some randomness), so the LLM agent should benefit from incorporating prior state/action feedback when planning.\\

\textbf{Warehouse}\quad Warehouse (Figure~\ref{fig:Envs}g) consists of robots that need to move all boxes to a target delivery region in the fewest time steps. The free space for the robots to move is discretized into cells, and a robot can only move to an adjacent cell in a single time step. Each cell can only contain one robot at each timestep. A robot is able to pick up a box if it is in the cell adjacent to that box. Each robot has five possible actions: (1) \& (2) move left or right if the adjacent cell exists, (3) pick up an adjacent box, (4) place the box to the target delivery region, (5) move from target delivery region to any adjacent cell of free space.\\

\textbf{Gridworld1}\quad Gridworld1 (Figure~\ref{fig:Envs}h) consists of obstacles (black) and goals (red). The robot needs to visit all goals, and any attempt to move into obstacles or move out of the grid will result in failure. The robot has five possible actions: (1) move up, (2) move down, (3) move left, (4) move right, (5) visit goal.\\

\textbf{Gridworld2}\quad Gridworld2 is similar to Gridworld1, but the goals must be visited in a particular order. The robot action are the same as in Gridworld1, but 'visit goal' can be performed only when the corresponding goal is in the correct order.\\

\textbf{Blocksworld}\quad In Blocksworld (Figure~\ref{fig:Envs}i), the goal is to stack a set of blocks (brown) according to a specific order. A robot can pick up, unstack, or stack a block only when the block is clear. A block is clear if the block has no other blocks on top of it and if the block is not picked up. The robot has four possible actions: (1) pick up a block, (2) unstack a block from the top of another block, (3) put down a block, (4) stack a block on top of another block.\\

\textbf{Logistics}\quad Logistics (Figure~\ref{fig:Envs}j) consists of objects, locations, and cities. The objects can be packages, trucks, or airplanes. The locations can be generic locations or airports, and each location is associated with a single city. Trucks can travel to different locations within a city but not to a different city; airplanes can travel to any airports, including those in other cities. The goal is to transport packages to their goal locations via the trucks (such as for intra-city travel) and the airplanes (such as for inter-city travel). The available actions are: (1) load a package into a truck, (2) load a package into an airplane, (3) unload a package from a truck, (4) unload a package from an airplane, (5) drive a truck from one location to another location within a city, (6) fly an airplane from one airport to another airport.

\newpage
\section{Generalization to different models for optimized prompts}
\label{appendix sec: Generalization to different models for optimized prompts}
\begin{table*}[t]
\caption{Scores for initial and optimized prompts using different types of LLMs as TaskLLM. The optimized prompts are the best discovered prompts by PROMST for GPT-3.5-0613. The optimized prompts are further tested with GPT-3.5-0301 and GPT-4 to study whether they can keep better performances than the initial prompts.}
\label{result-table-generalization1}
\vskip 0.15in
\begin{center}
\begin{small}
\begin{sc}
\begin{tabular}{lccccccr}
\toprule
\multicolumn{1}{c}{} & \multicolumn{6}{c}{\textbf{GPT-3.5-0613-as-TaskLLM, GPT-4-as-PromptLLM}}\\
\midrule
\multicolumn{1}{c}{} & \multicolumn{2}{c|}{\textbf{GPT-3.5-0613}} & \multicolumn{2}{c|}{\textbf{GPT-3.5-0301}} & \multicolumn{2}{c}{\textbf{GPT-4}} \\
\multicolumn{1}{c}{Task} & \multicolumn{1}{c}{Human} & \multicolumn{1}{c|}{PROMST} & \multicolumn{1}{c}{Human} & \multicolumn{1}{c|}{PROMST} & \multicolumn{1}{c}{Human} & \multicolumn{1}{c}{PROMST} \\
\midrule
\multicolumn{1}{c}{Webarena} & \multicolumn{1}{c}{0.22} & \multicolumn{1}{c|}{\textcolor{magenta}{0.35}} & \multicolumn{1}{c}{0.29} & \multicolumn{1}{c|}{\textcolor{magenta}{0.34}} & \multicolumn{1}{c}{\textcolor{magenta}{0.57}} & \multicolumn{1}{c}{0.54} \\
\multicolumn{1}{c}{Alfworld} & \multicolumn{1}{c}{0.075} & \multicolumn{1}{c|}{\textcolor{magenta}{0.30}} & \multicolumn{1}{c}{0.17} & \multicolumn{1}{c|}{\textcolor{magenta}{0.21}} & \multicolumn{1}{c}{0.45} & \multicolumn{1}{c}{\textcolor{magenta}{0.49}} \\
\multicolumn{1}{c}{Scienceworld} & \multicolumn{1}{c}{0.18} & \multicolumn{1}{c|}{\textcolor{magenta}{0.21}} & \multicolumn{1}{c}{\textcolor{magenta}{0.16}} & \multicolumn{1}{c|}{0.13} & \multicolumn{1}{c}{\textcolor{magenta}{0.70}} & \multicolumn{1}{c}{0.68} \\
\multicolumn{1}{c}{BoxNet1} & \multicolumn{1}{c}{0.076} & \multicolumn{1}{c|}{\textcolor{magenta}{0.25}} & \multicolumn{1}{c}{0.28} & \multicolumn{1}{c|}{\textcolor{magenta}{0.38}} & \multicolumn{1}{c}{0.65} & \multicolumn{1}{c}{\textcolor{magenta}{0.67}} \\
\multicolumn{1}{c}{BoxNet2} & \multicolumn{1}{c}{0.044} & \multicolumn{1}{c|}{\textcolor{magenta}{0.22}} & \multicolumn{1}{c}{0.088} & \multicolumn{1}{c|}{\textcolor{magenta}{0.28}} & \multicolumn{1}{c}{\textcolor{magenta}{0.34}} & \multicolumn{1}{c}{0.31}\\
\multicolumn{1}{c}{BoxLift} & \multicolumn{1}{c}{0.31} & \multicolumn{1}{c|}{\textcolor{magenta}{0.90}} & \multicolumn{1}{c}{0.69} & \multicolumn{1}{c|}{\textcolor{magenta}{0.91}} & \multicolumn{1}{c}{1.0} & \multicolumn{1}{c}{1.0}\\
\multicolumn{1}{c}{WareHouse} & \multicolumn{1}{c}{0.0} & \multicolumn{1}{c|}{\textcolor{magenta}{0.028}} & \multicolumn{1}{c}{0.0} & \multicolumn{1}{c|}{\textcolor{magenta}{0.040}}  & \multicolumn{1}{c}{0.16} & \multicolumn{1}{c}{0.19}\\
\multicolumn{1}{c}{Gridworld1} & \multicolumn{1}{c}{0.23} & \multicolumn{1}{c|}{\textcolor{magenta}{0.38}} & \multicolumn{1}{c}{0.25} & \multicolumn{1}{c|}{\textcolor{magenta}{0.32}}  & \multicolumn{1}{c}{0.73} & \multicolumn{1}{c}{\textcolor{magenta}{0.85}}\\
\multicolumn{1}{c}{Gridworld2} & \multicolumn{1}{c}{0.036} & \multicolumn{1}{c|}{0.12} & \multicolumn{1}{c}{0.021} & \multicolumn{1}{c|}{0.13}  & \multicolumn{1}{c}{0.26} & \multicolumn{1}{c}{\textcolor{magenta}{0.29}}\\
\multicolumn{1}{c}{Blocksworld} & \multicolumn{1}{c}{0.19} & \multicolumn{1}{c|}{\textcolor{magenta}{0.60}} & \multicolumn{1}{c}{\textcolor{magenta}{0.33}} & \multicolumn{1}{c|}{0.24}  & \multicolumn{1}{c}{0.71} & \multicolumn{1}{c}{0.62}\\
\multicolumn{1}{c}{Logistics} & \multicolumn{1}{c}{0.083} & \multicolumn{1}{c|}{\textcolor{magenta}{0.18}} & \multicolumn{1}{c}{\textcolor{magenta}{0.12}} & \multicolumn{1}{c|}{0.083}  & \multicolumn{1}{c}{0.50} & \multicolumn{1}{c}{\textcolor{magenta}{0.63}}\\
\multicolumn{1}{c}{\textbf{Average}} & \multicolumn{1}{c}{0.13} & \multicolumn{1}{c|}{\textcolor{magenta}{0.32}} & \multicolumn{1}{c}{0.22} & \multicolumn{1}{c|}{\textcolor{magenta}{0.28}} & \multicolumn{1}{c}{0.55} & \multicolumn{1}{c}{\textcolor{magenta}{0.57}}\\
\bottomrule
\end{tabular}
\end{sc}
\end{small}
\end{center}
\vskip -0.1in
\end{table*}

\begin{table*}[t]
\caption{Scores for initial and optimized prompts using different types of LLMs as TaskLLM. The optimized prompts are the best discovered prompts by PROMST for GPT-4. The optimized prompts are further tested with GPT-3.5-0301 and GPT-3.5-0613 to study whether they can keep better performances than the initial prompts.}
\label{result-table-generalization2}
\vskip 0.15in
\begin{center}
\begin{small}
\begin{sc}
\begin{tabular}{lccccccr}
\toprule
\multicolumn{1}{c}{} & \multicolumn{6}{c}{\textbf{GPT-4-as-TaskLLM, GPT-4-as-PromptLLM}}\\
\midrule
\multicolumn{1}{c}{} & \multicolumn{2}{c|}{\textbf{GPT-3.5-0613}} & \multicolumn{2}{c|}{\textbf{GPT-3.5-0301}} & \multicolumn{2}{c}{\textbf{GPT-4}} \\
\multicolumn{1}{c}{Task} & \multicolumn{1}{c}{Human} & \multicolumn{1}{c|}{PROMST} & \multicolumn{1}{c}{Human} & \multicolumn{1}{c|}{PROMST} & \multicolumn{1}{c}{Human} & \multicolumn{1}{c}{PROMST} \\
\midrule
\multicolumn{1}{c}{Webarena} & \multicolumn{1}{c}{\textcolor{magenta}{0.22}} & \multicolumn{1}{c|}{0.18} & \multicolumn{1}{c}{0.29} & \multicolumn{1}{c|}{\textcolor{magenta}{0.32}} & \multicolumn{1}{c}{0.57} & \multicolumn{1}{c}{\textcolor{magenta}{0.62}} \\
\multicolumn{1}{c}{Alfworld} & \multicolumn{1}{c}{0.075} & \multicolumn{1}{c|}{\textcolor{magenta}{0.092}} & \multicolumn{1}{c}{0.17} & \multicolumn{1}{c|}{\textcolor{magenta}{0.19}} & \multicolumn{1}{c}{0.45} & \multicolumn{1}{c}{\textcolor{magenta}{0.57}} \\
\multicolumn{1}{c}{Scienceworld} & \multicolumn{1}{c}{\textcolor{magenta}{0.18}} & \multicolumn{1}{c|}{0.15} & \multicolumn{1}{c}{0.16} & \multicolumn{1}{c|}{\textcolor{magenta}{0.20}} & \multicolumn{1}{c}{0.70} & \multicolumn{1}{c}{\textcolor{magenta}{0.81}} \\
\multicolumn{1}{c}{BoxNet1} & \multicolumn{1}{c}{0.076} & \multicolumn{1}{c|}{\textcolor{magenta}{0.12}} & \multicolumn{1}{c}{0.28} & \multicolumn{1}{c|}{\textcolor{magenta}{0.32}} & \multicolumn{1}{c}{0.65} & \multicolumn{1}{c}{\textcolor{magenta}{0.79}} \\
\multicolumn{1}{c}{BoxNet2} & \multicolumn{1}{c}{0.044} & \multicolumn{1}{c|}{\textcolor{magenta}{0.14}} & \multicolumn{1}{c}{0.088} & \multicolumn{1}{c|}{\textcolor{magenta}{0.17}} & \multicolumn{1}{c}{0.34} & \multicolumn{1}{c}{\textcolor{magenta}{0.42}}\\
\multicolumn{1}{c}{WareHouse} & \multicolumn{1}{c}{0.0} & \multicolumn{1}{c|}{\textcolor{magenta}{0.019}} & \multicolumn{1}{c}{0.0} & \multicolumn{1}{c|}{0.025}  & \multicolumn{1}{c}{0.16} & \multicolumn{1}{c}{\textcolor{magenta}{0.51}}\\
\multicolumn{1}{c}{Gridworld1} & \multicolumn{1}{c}{0.23} & \multicolumn{1}{c|}{\textcolor{magenta}{0.26}} & \multicolumn{1}{c}{0.25} & \multicolumn{1}{c|}{\textcolor{magenta}{0.29}}  & \multicolumn{1}{c}{0.73} & \multicolumn{1}{c}{\textcolor{magenta}{0.86}}\\
\multicolumn{1}{c}{Gridworld2} & \multicolumn{1}{c}{0.036} & \multicolumn{1}{c|}{\textcolor{magenta}{0.057}} & \multicolumn{1}{c}{0.021} & \multicolumn{1}{c|}{\textcolor{magenta}{0.042}}  & \multicolumn{1}{c}{0.26} & \multicolumn{1}{c}{\textcolor{magenta}{0.60}}\\
\multicolumn{1}{c}{Blocksworld} & \multicolumn{1}{c}{0.19} & \multicolumn{1}{c|}{\textcolor{magenta}{0.21}} & \multicolumn{1}{c}{\textcolor{magenta}{0.33}} & \multicolumn{1}{c|}{0.24}  & \multicolumn{1}{c}{0.71} & \multicolumn{1}{c}{\textcolor{magenta}{0.95}}\\
\multicolumn{1}{c}{Logistics} & \multicolumn{1}{c}{0.083} & \multicolumn{1}{c|}{0.083} & \multicolumn{1}{c}{0.12} & \multicolumn{1}{c|}{\textcolor{magenta}{0.18}}  & \multicolumn{1}{c}{0.50} & \multicolumn{1}{c}{\textcolor{magenta}{0.74}}\\
\multicolumn{1}{c}{\textbf{Average}} & \multicolumn{1}{c}{0.11} & \multicolumn{1}{c|}{\textcolor{magenta}{0.13}} & \multicolumn{1}{c}{0.17} & \multicolumn{1}{c|}{\textcolor{magenta}{0.20}} & \multicolumn{1}{c}{0.56} & \multicolumn{1}{c}{\textcolor{magenta}{0.69}}\\
\bottomrule
\end{tabular}
\end{sc}
\end{small}
\end{center}
\vskip -0.1in
\end{table*}

\newpage
\section{Task progress score vs. task completion score}
\label{appendix sec: Task progress score vs. task completion score}
\begin{table*}[t]
\caption{Corresponding values of task progress rates (score format used in our study) and task completion rates (another possible score format). The task completion score is positively correlated with the task progress score. However, the task completion score has lower value and sensitivity since it is more sparse, which is the reason why we did not use it as the metric in our study.}
\label{ablation-table-completion-progress-score}
\vskip 0.15in
\begin{center}
\begin{small}
\begin{sc}
\begin{tabular}{lcccccccr}
\toprule
\multicolumn{7}{c}{\textbf{BoxLift, GPT-3.5-as-TaskLLM, GPT-4-as-PromptLLM}}\\
\midrule
\multicolumn{1}{c}{Level number} & \multicolumn{1}{c}{1} & \multicolumn{1}{c}{2} & \multicolumn{1}{c}{3} & \multicolumn{1}{c}{4} & \multicolumn{1}{c}{5} & \multicolumn{1}{c}{6}\\
\multicolumn{1}{c}{Task progress rates/Scores} & \multicolumn{1}{c}{0.34} & \multicolumn{1}{c}{0.69} & \multicolumn{1}{c}{0.79} & \multicolumn{1}{c}{0.86} & \multicolumn{1}{c}{0.90} & \multicolumn{1}{c}{0.90} \\
\multicolumn{1}{c}{Task completion rates/Scores} & \multicolumn{1}{c}{0.18} & \multicolumn{1}{c}{0.21} & \multicolumn{1}{c}{0.24} & \multicolumn{1}{c}{0.27} & \multicolumn{1}{c}{0.31} & \multicolumn{1}{c}{0.31}\\
\bottomrule
\multicolumn{7}{c}{\textbf{BoxNet1, GPT-4-as-TaskLLM, GPT-4-as-PromptLLM}}\\
\midrule
\multicolumn{1}{c}{Level number} & \multicolumn{1}{c}{1} & \multicolumn{1}{c}{2} & \multicolumn{1}{c}{3} & \multicolumn{1}{c}{4} & \multicolumn{1}{c}{5} & \multicolumn{1}{c}{6}\\
\multicolumn{1}{c}{Task progress rates/Scores} & \multicolumn{1}{c}{0.13} & \multicolumn{1}{c}{0.30} & \multicolumn{1}{c}{0.56} & \multicolumn{1}{c}{0.65} & \multicolumn{1}{c}{0.76} & \multicolumn{1}{c}{0.79} \\
\multicolumn{1}{c}{Task completion rates/Scores} & \multicolumn{1}{c}{0.03} & \multicolumn{1}{c}{0.06} & \multicolumn{1}{c}{0.14} & \multicolumn{1}{c}{0.16} & \multicolumn{1}{c}{0.19} & \multicolumn{1}{c}{0.21}\\
\bottomrule
\end{tabular}
\end{sc}
\end{small}
\end{center}
\vskip -0.1in
\end{table*}

\section{Extra ablation experiments of score models}
\label{appendix sec: Extra abalation experiments of score models}
\begin{figure*}[ht]
  \centering
   \includegraphics[width=0.95\linewidth]{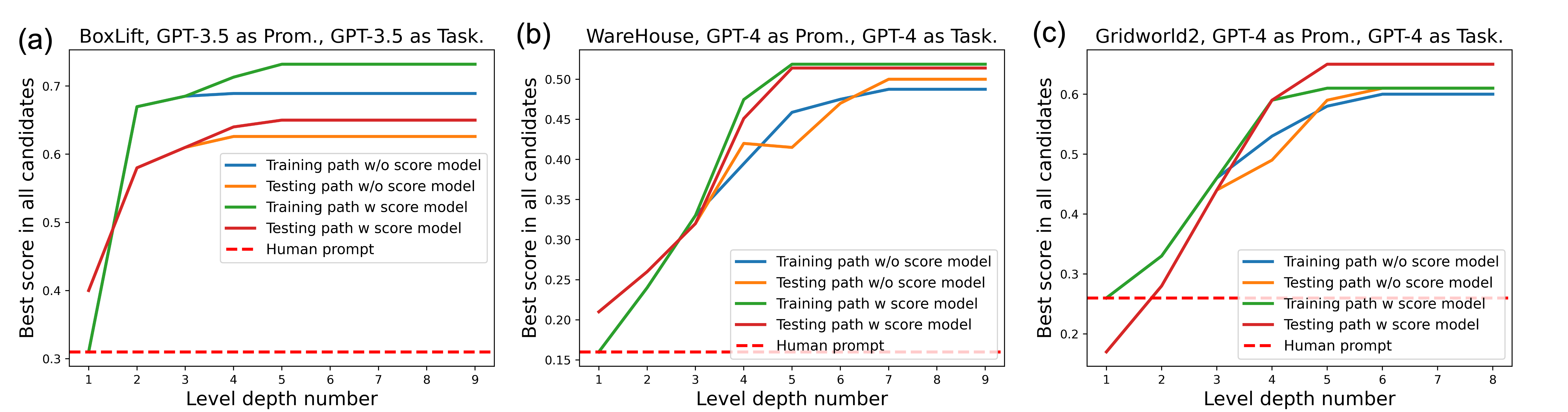}

   \caption{Ablation study of applying score models for prompt selection. The optimization trace with score model finds better prompts with less iteration steps, across all the three tasks BoxLift, WareHouse, and GridWorld2.}
   \label{fig:GPT-3.5score-model}
\end{figure*}

\newpage
\section{Efforts for designing human feedback rules}
\label{appendix sec: Efforts for designing human feedback rules}
Regarding the expected human efforts when designing the feedback rules, we note that several of the templates are common across all tasks (e.g. syntax errors). Thus, for a novel task, a human user is only expected to need to design a few (3-5) templates, and they are typically intuitive as they relate to the specific task. While this is a low-effort requirement, our primary experimental results show that this effort can significantly improve task performance.

In order to better understand the generalizability of the human-designed feedback templates, we perform an additional ablation study as shown in Table~\ref{table:ablation-table-human-feedback}. In our work, the original feedback templates did not require iterations via trial-and-error over possible versions. In the ablation study, we compare using the original templates with four other variations of the templates:

\textbf{(1) Paraphrased} We use GPT-4 to generate semantically-consistent paraphrased versions of the original feedback templates to replace the original templates. This both simulates variation across human users and tests the sensitivity of the wording.

\textbf{(2) Random} We use GPT-4 to generate 10 different versions of a template for each type of error. During optimization, we randomly sample from these 10 possible templates per error type, introducing more fine-grained variation than in (1).

\textbf{(3) WO response component} In the original feedback templates, we included the output of the TaskLLM that was incorrect (see Figure~\ref{fig:Human feedback}) for better reasoning of PromptLLM. In this ablation, we test the impact of removing this component.

\textbf{(4) WO stuck in the loop} We exclude the error type of being stuck in a loop to test the impact of a human user choosing to include different types of error feedback. The results for two tasks from our larger experiments are provided below.

The results in Table~\ref{table:ablation-table-human-feedback} show that variability over the wording of the templates has little impact on the performance of PROMST. Interestingly, randomly choosing paraphrased templates ((2) from the above description) generally improves performance; we suspect this may be due to increased diversity over generated prompt candidates and is worth further investigation. This ablation also shows that removing the TaskLLM response ((3) from above) and removing the 'stuck in a loop' error type both reduce the performance. Based on the above discussion, we can conclude that including the response in the feedback template is a useful task- and error-agnostic guiding principle.

\newpage
\section{Component changes in each environment}
\label{appendix sec: Component changes in each environment}
In this section we display the evolution of task executing characteristics (task success number, executing step number, syntactic error number, query limit error number, stuck in loop number, collision error number) vs. testing scores for all the prompts explored during prompt optimization process. The three shown tasks (BoxLift, BoxNet1, BoxNet2) share the same trends such as the rising executing step number and decreasing syntactic error number with the increasing testing score, but also have different trends such as stuck in loop number.

\begin{figure*}[ht]
  \centering
   \includegraphics[width=0.85\linewidth]{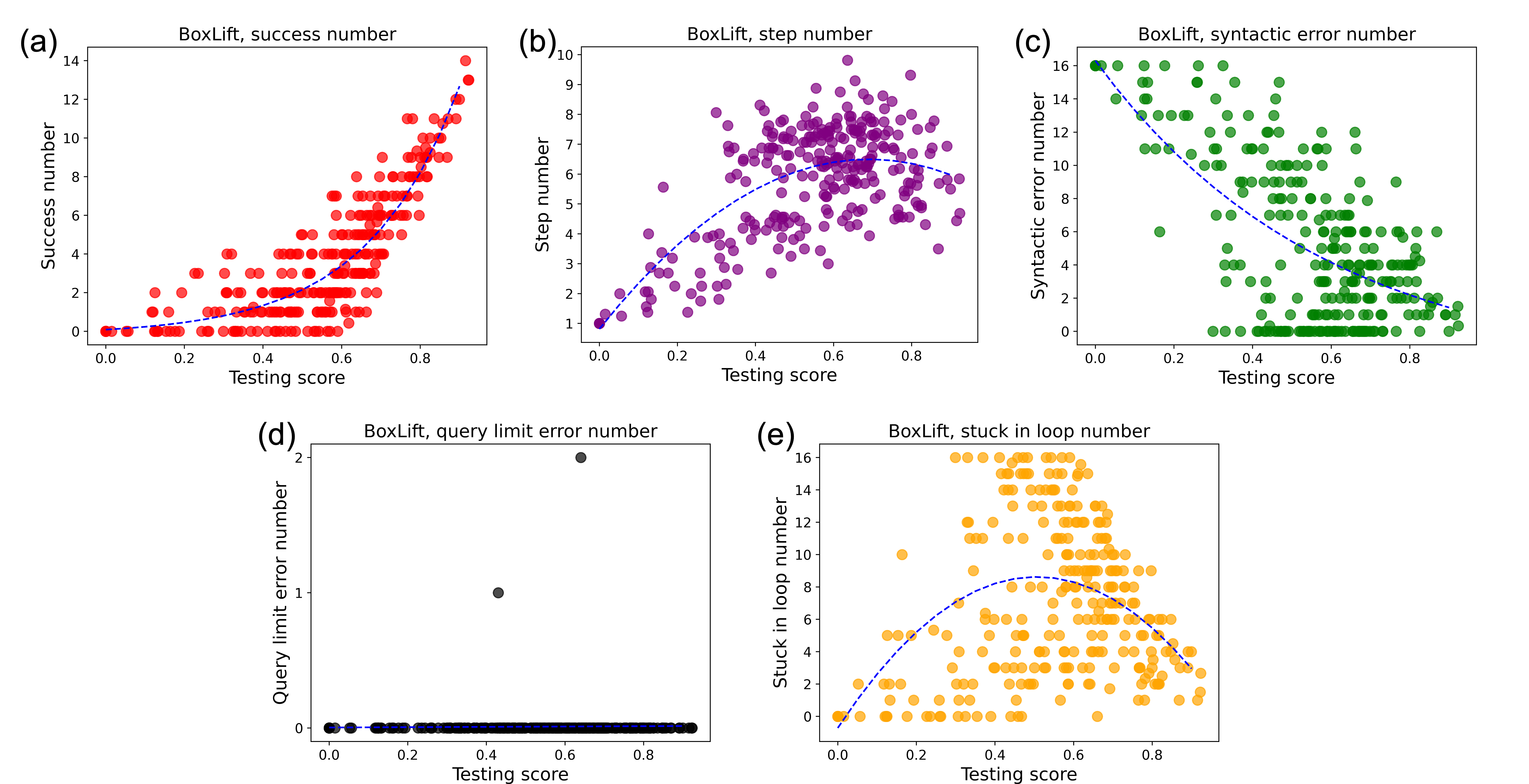}

   \caption{Component change of BoxLift.}
   \label{fig:Appendix-boxlift-component}
\end{figure*}

\begin{figure*}[ht]
  \centering
   \includegraphics[width=0.85\linewidth]{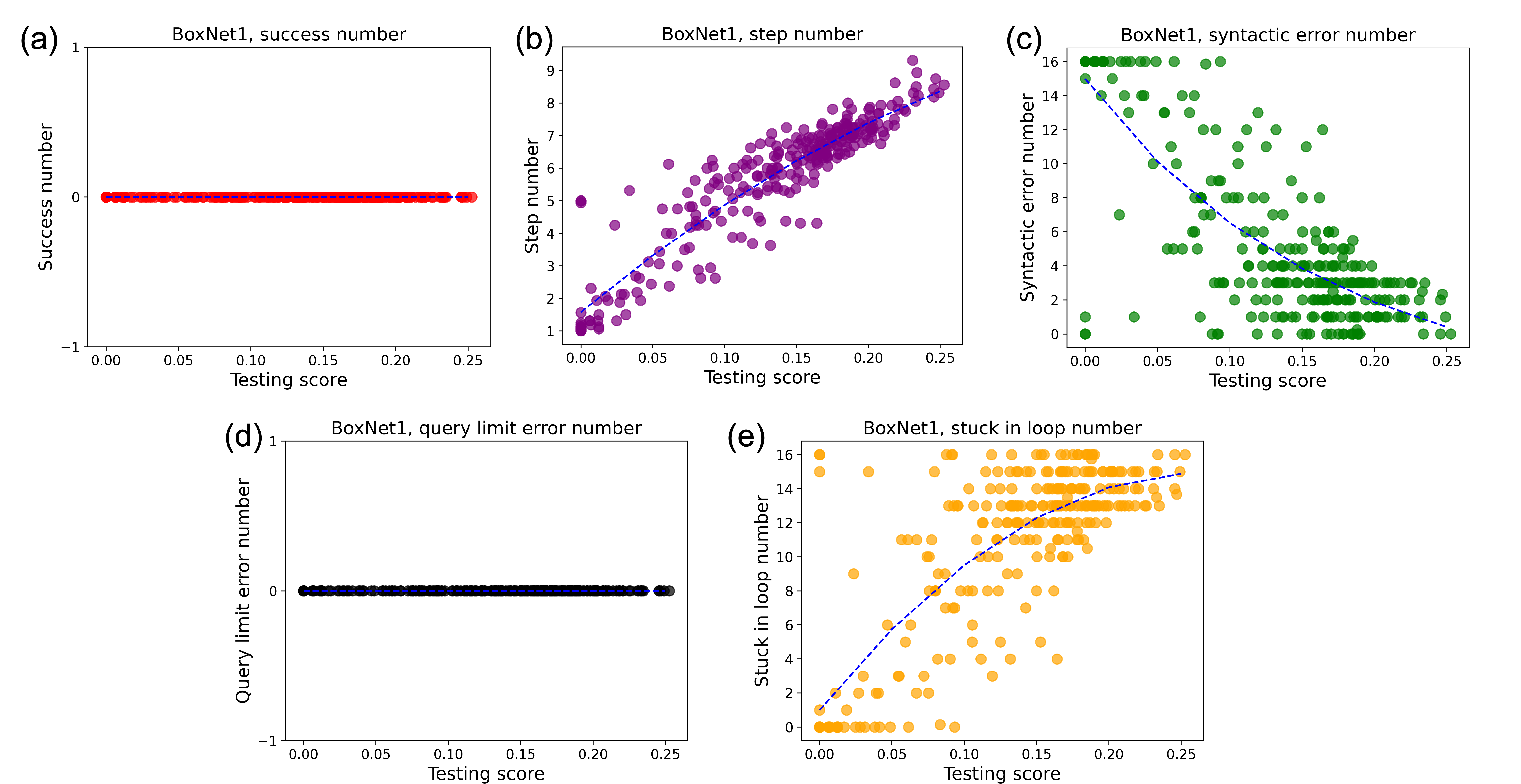}

   \caption{Component change of BoxNet1.}
   \label{fig:Appendix-boxnet1-component}
\end{figure*}

\begin{figure*}[ht]
  \centering
   \includegraphics[width=0.85\linewidth]{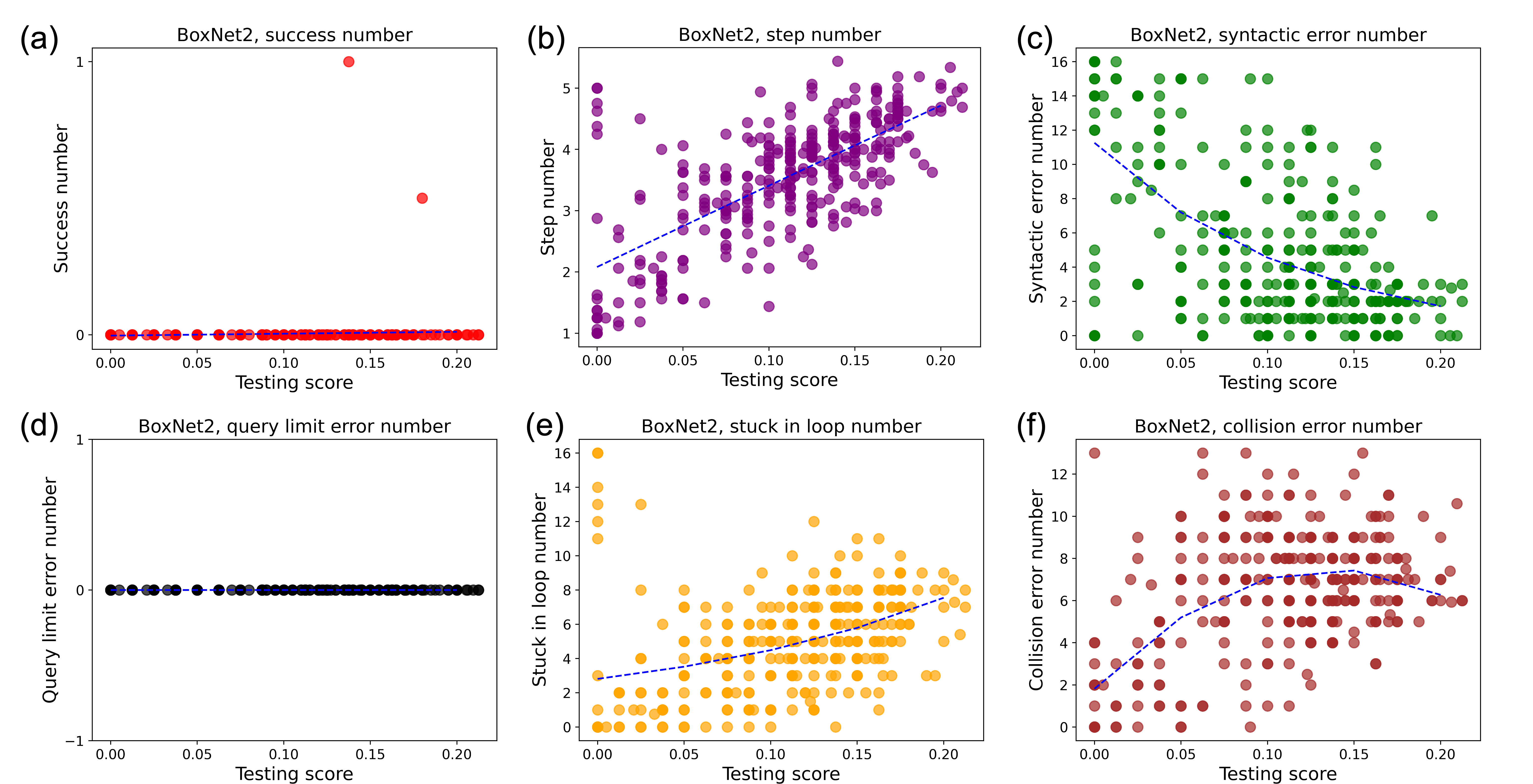}

   \caption{Component change of BoxNet2.}
   \label{fig:Appendix-boxnet2-component}
\end{figure*}

\newpage
\section{The influence of score functions}
\label{appendix sec: The influence of score functions}
The choice of score function impacts prompt optimization. The initial score functions in Equation~\ref{eq:score-func} are simple and intuitive, only caring about the number of goals/sub-steps accomplished. However, humans may have different preferences for the same task. For instance, a user may also care about efficiency (the number of action steps taken) or safety (collision avoidance). We observe a general trend that the step number increases as the prompt score increases in all the three shown tasks (see Appendix~\ref{appendix sec: Component changes in each environment}). However, in BoxNet2 (Figure~\ref{fig:Appendix-boxnet2-component}) the collision error number gradually increases with the increasing prompt scores. These two general trends are not aligned with the user preference.

\begin{figure*}[ht]
  \centering
   \includegraphics[width=0.85\linewidth]{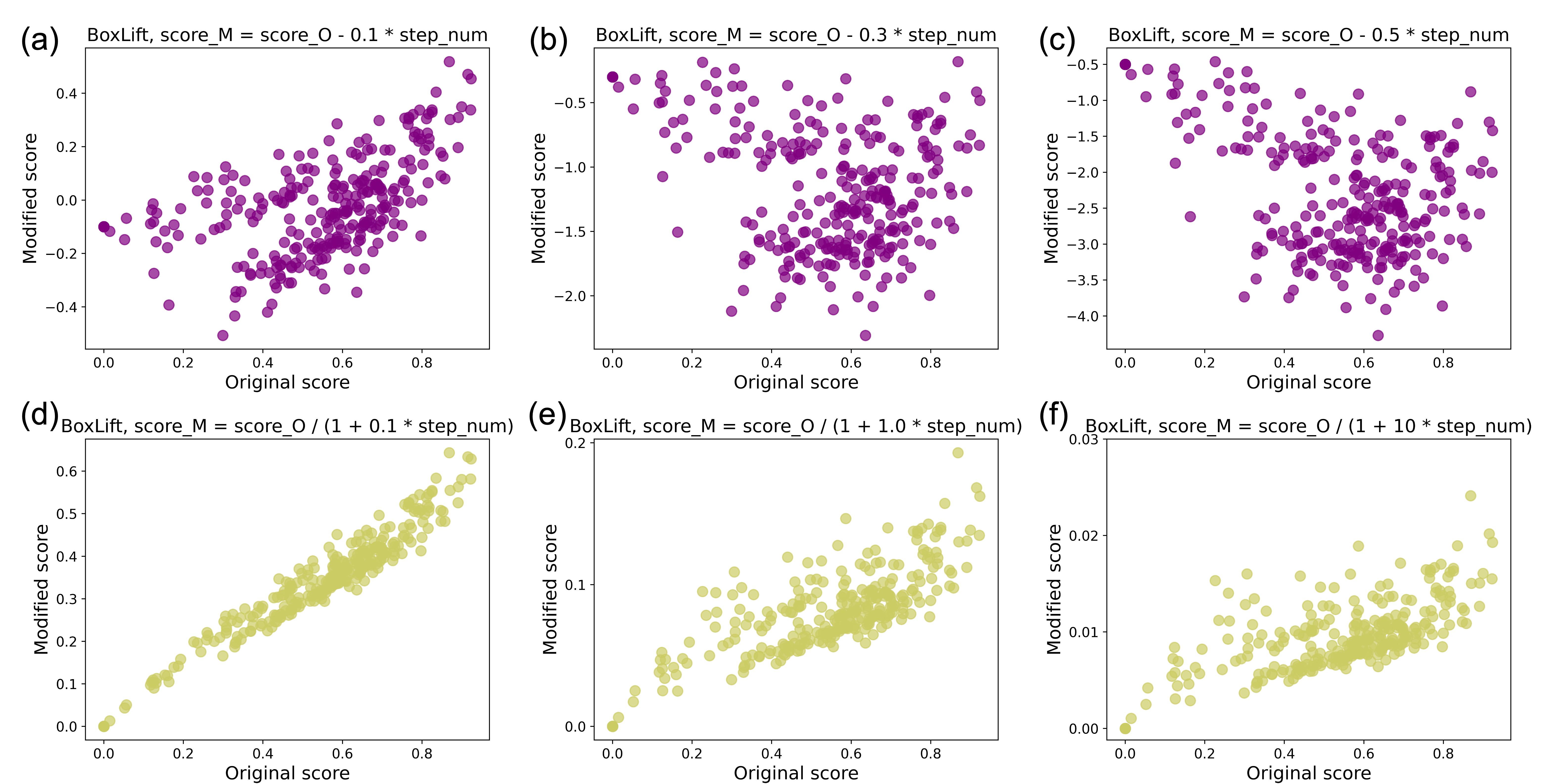}

   \caption{Modified score of BoxLift.}
   \label{fig:Appendix-boxlift-modified-score}
\end{figure*}

\begin{figure*}[ht]
  \centering
   \includegraphics[width=0.85\linewidth]{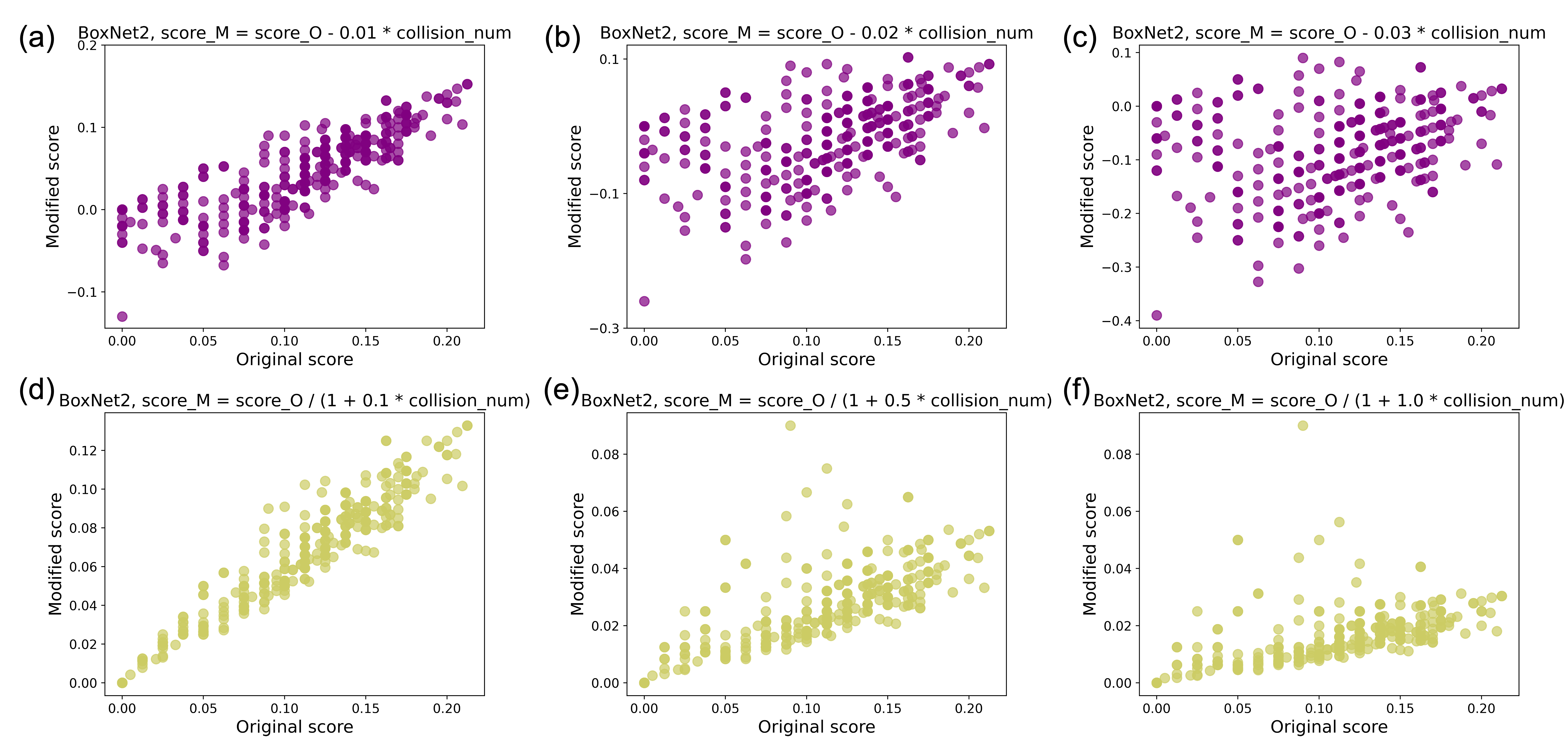}

   \caption{Modified score of BoxNet2.}
   \label{fig:Appendix-boxnet2-modified-score}
\end{figure*}

Then how to design the score function to balance user preferences remains an issue. In Appendix~\ref{appendix sec: The influence of score functions}, we tried the two forms of modified scores:
\begin{equation}
S_{M} = S_{O} - \text{ratio}*\text{factor_value}
\label{eq:modified-score-func1}
\end{equation}
\begin{equation}
S_{M} = S_{O}/ ( 1 + \text{ratio}*\text{factor_value}),
\label{eq:modified-score-func2}
\end{equation}
where $S_{M}$ and $S_{O}$ are the modified score and the original score (defined in Equation~\ref{eq:score-func}), respectively. The factor\_value is a factor that the user cares about, e.g., step number or collision error number. We find that the general $S_{M}$ vs. $S_{O}$ trend can be tuned quite disparately by adjusting the hyperparameter ratio (see Figure~\ref{fig:Appendix-boxlift-modified-score} and Figure~\ref{fig:Appendix-boxnet2-modified-score}).

\begin{figure}[ht]
  \centering
   \includegraphics[width=0.8\linewidth]{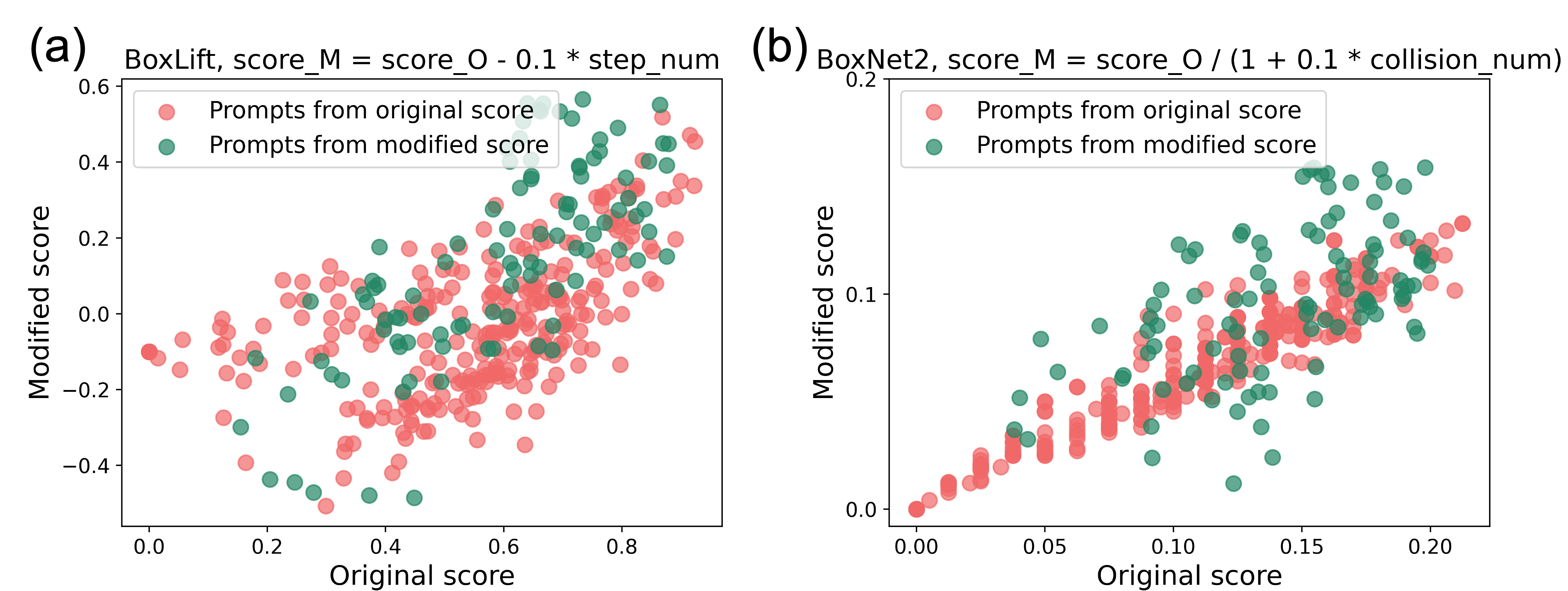}

   \caption{Human preference alignment via tuning score functions. The green dots are the new prompts further optimized over new score rules.}
   \label{fig:Preference}
\end{figure}

We choose two modified score functions that trend similarly to the original score function. Then we optimize the prompts with PROMST but using modified score function. To save computing resources, we initialize the prompt optimization with the best prompts found with the original score function. Figure~\ref{fig:Preference} shows the optimization results. Compared to the original prompts acquired with the original score function (red), the newly discovered prompts (green) generally have higher modified scores, though the values of original scores slightly decrease. This suggests that we can align with human preferences by changing the form of the score functions, which can be captured and revealed by the selection framework in PROMST.

\newpage
\section{Explanability for better prompts}
\label{appendix sec: Explanability for better prompts}
We are interested in whether there are features of the prompts that correlate with high scores.

\begin{figure*}[ht]
  \centering
   \includegraphics[width=0.95\linewidth]{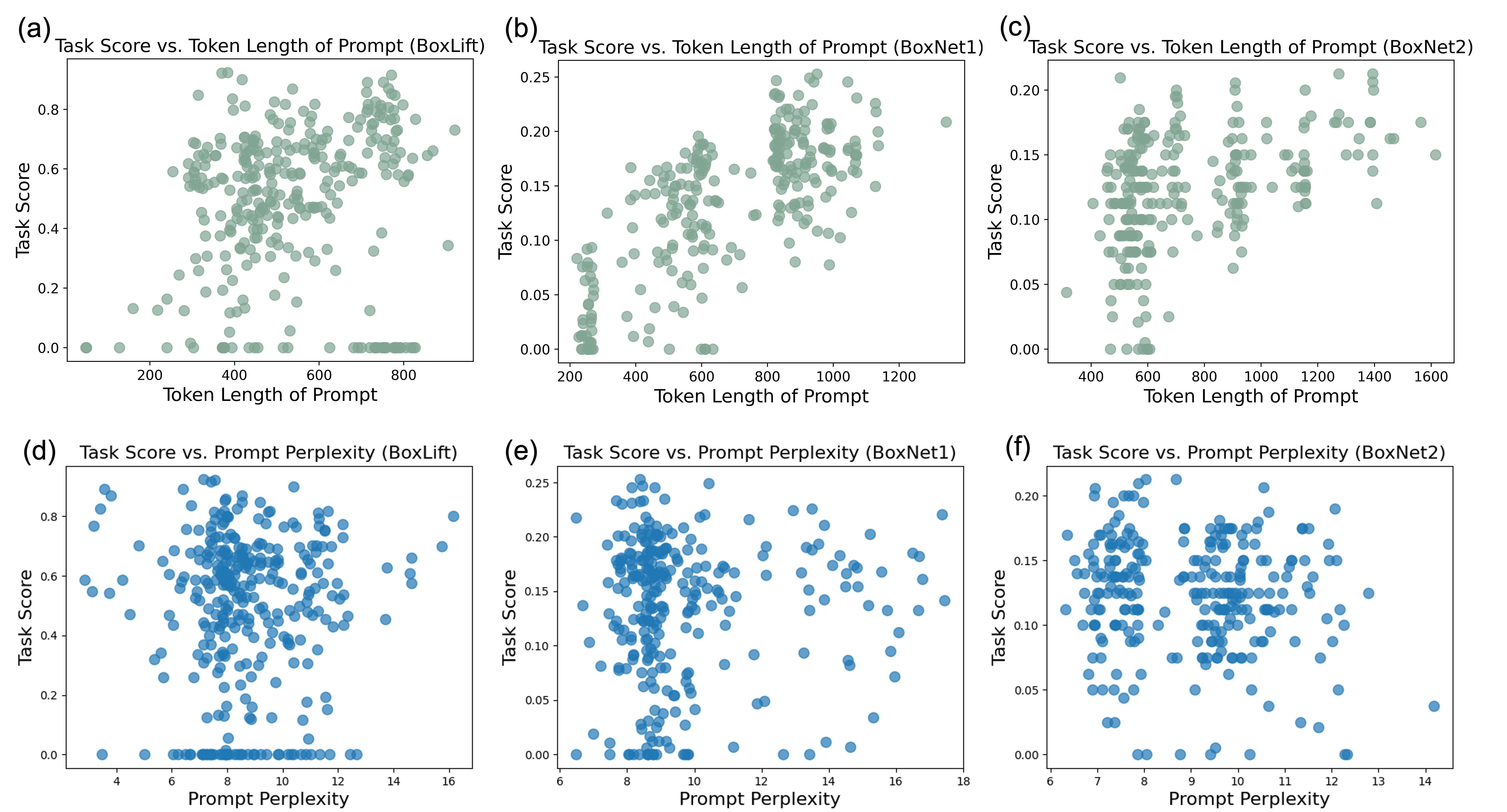}

   \caption{Score vs. prompt token length and score vs. prompt perplexity for all the explored prompts in each task BoxLift, BoxNet1, and BoxNet2.}
   \label{fig:Appendix-visua-token-len}
\end{figure*}

\textbf{Prompt score vs. token length and perplexity} As plotted in Figure~\ref{fig:Appendix-visua-token-len}, we found that there is a rough trend across different tasks that longer prompts corresponded with higher scores. We also investigated prompt perplexity (using GPT-2 to get prompt token log probabilities) but found no clear correlation.  All the initial and discovered best prompts are listed in Appendix~\ref{appendix sec: all prompts}.

\begin{figure*}[ht]
  \centering
   \includegraphics[width=0.9\linewidth]{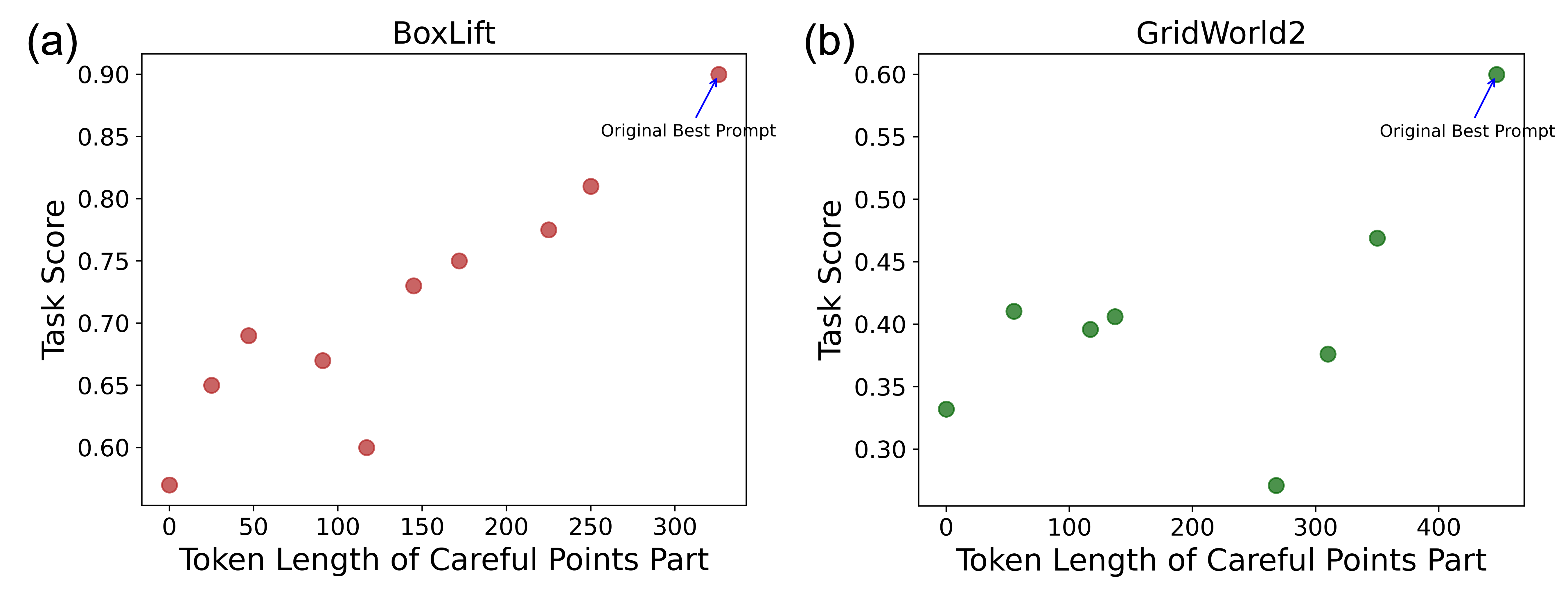}

   \caption{Evolution of Task Score vs. Token Length of Careful Points Part in BoxLift and GridWorld2. The contexts of careful points part with different token lengths are summarized by GPT-4. The topmost data point is the original context from the best prompt in each task.}
   \label{fig:prompt point-to-point component study}
\end{figure*}

\textbf{Listing careful points one by one clearly} We also find some clues about better component emergence. The best prompts tend to list all the careful points of the task one by one clearly, which is consistent to human intuitions. To study whether this characteristic can effectively improve the performance, we carry out the ablation study by compressing these careful points into shortened context summarized by GPT-4 and then testing their performance. We obtain contexts of varying token lengths by querying GPT-4 to summarize with different levels of detail. In the following two texts, we display the original best prompt of BoxLift with token length of 326 and one corresponding compressed prompt with token length of 145 as an example. As shown in Figure~\ref{fig:prompt point-to-point component study}, the task score of both BoxLift and GridWorld2 decline with the decreasing length of careful points part. This reveals that listing the careful points of the task one by one clearly is an effective method to enhance task executing performance, which is automatically emerged from prompt optimization process. This characteristic also inspires the designing of prompts in complex tasks with multiple constraints.

\begin{boxL}
\textbf{BoxLift Original Best prompt for GPT-3.5-turbo-16k-0613}\\
\textbf{Token Length of Careful Points Part = 326, Score = 0.90}\\

- Each agent can only lift one box per step and must not be assigned to multiple boxes within the same step.\\
- Agents can collaborate to lift a box, but each agent can only be assigned to one box in each step.\\
- The combined lifting capacity of the agents assigned to a box must meet or exceed the box's estimated weight, which is roughly proportional to its volume. Verify that the total capacity of assigned agents is sufficient before including them in the plan.\\
- Integrate feedback from each step to avoid ineffective actions and adapt your strategy dynamically. Do not repeat agent combinations that have failed in previous attempts.\\
- Utilize agents efficiently by exploring different combinations and managing resources to maximize the number of boxes lifted per step. Ensure that agents are not duplicated within the same action plan.\\
- Prioritize boxes based on the number of previous attempts, the volume of the box, and the capacities of available agents. Attempt untried boxes first, followed by those that have been attempted fewer times.\\
- Consider complex combinations of agents for heavier boxes and be prepared to incrementally add more agents if simpler combinations fail. Provide examples of how to form these combinations.\\
- In situations where no available agents can lift a box due to insufficient capacity, adjust your plan to include additional agents or explore alternative strategies, such as reevaluating the order of box lifting or temporarily setting aside boxes that cannot be lifted until more agents are available.\\
- Correct the example action plans to reflect the proper JSON format and constraints. Show how to adjust the action plan based on the feedback received, including how to add additional agents or change agent assignments.
\end{boxL}

\begin{boxL}
\textbf{BoxLift Compressed prompt for GPT-3.5-turbo-16k-0613}\\
\textbf{Token Length of Careful Points Part = 145, Score = 0.73}\\

Agents can lift one box per step and must not handle multiple boxes in the same step. Collaboration is allowed, but each agent is limited to one box per step. The combined lifting capacity of agents must meet or exceed a box's estimated weight. Verify agent capacity before planning. Avoid repeating failed combinations and adapt strategies dynamically. Optimize agent efficiency by exploring different combinations and managing resources to maximize lifted boxes per step. Prioritize untried boxes and those attempted fewer times. Use complex agent combinations for heavier boxes and adjust plans if no agents can lift a box, considering alternative strategies or reordering tasks. Correct action plans to reflect constraints and JSON format, showing adjustments based on feedback, including adding agents or changing assignments.
\end{boxL}

\newpage
\section{Comparison and combination with Reflexion}
\label{appendix: Comparison and Combination with Dynamic Approach}
Instead of utilizing the execution feedback by optimizing offline prompts such as PROMST, dynamic approaches like Reflexion~\cite{reflexion} directly optimizing the action plan during multiple online trials. These methods assume the agent has chances of trying multiple trials and the new trial depends on the reflection from previous failed trials. The contexts summarized from the reflection are saved in the memory module, which is unique for each testing case. Though the original approach can not be directly applied into the task of prompt optimization, here we modify the original online Reflexion into the offline Reflexion for the comparison with PROMST on prompt optimization. That is, we add the memory module into the initial prompt to explore whether it can help improve the prompt performance.

Since the memory module is varied in each testing case, we tried two methods to add onto the initial prompt: 1) \textbf{Offline Reflexion 1}: Directly concatenating several numbers/cases of memory module with the initial prompt; 2) \textbf{Offline Reflexion 2}: Querying LLM to summarize the memory from the multiple cases and then concatenate with the initial prompt. The prompts used for acquiring memory modules are from the Reflexion paper, while the summarization prompt is designed by ourselves. The feedback is purely from the environment without the human factors. 

Table~\ref{table: Comparison and Combination with Dynamic Approach} shows the testing results. We find that the Offline Reflexion 1 is not very effective in prompt optimization. The reason is that the distilled memory is always too case-specific so that the context can not well generalize to the diversified testing data. Meanwhile, the memory in each case is lengthy so that the final prompt can only contain 3-5 cases, which can not act as a well-rounded guidance during testing. The summarization over multiple case memories is a good way to mitigate this issue, as shown in Offline Reflexion 2. However, Offline Reflexion 2 is still not as effective as PROMST, which is reasonable since Offline Reflexion 2 does not have following modules compared to PROMST: optimization on classifying the error types, human feedback, genetic algorithms, and score prediction model.

We further test whether PROMST can perform better when integrating with Online Reflexion. That means we directly test the best prompts discovered in PROMST under the multi-trial setting and use Reflexion as the online feedback module. The results in Table~\ref{table: Comparison and Combination with Dynamic Approach} reveal that the online feedback truly further enhances the agent performance, which is consistent with the intuition since previous failed trials help a better decision making in the current trial. The above results show that PROMST performs better than Reflexion in offline prompt optimization and can well combine with Reflexion in online task execution, resulting into a better solution.

\begin{table*}[t]
\caption{Comparison and combination with dynamic approach Reflexion.}
\label{table: Comparison and Combination with Dynamic Approach}
\vskip 0.15in
\begin{center}
\begin{small}
\begin{sc}
\begin{tabular}{lcccccr}
\toprule
\multicolumn{5}{c}{\textbf{GPT-3.5-as-TaskLLM, GPT-4-as-PromptLLM}}\\
\multicolumn{1}{c}{Task} & \multicolumn{1}{c}{PROMST} & \multicolumn{1}{c}{Offline Reflexion 1} & \multicolumn{1}{c}{Offline Reflexion 2} & \multicolumn{1}{c}{PROMST + Online Reflexion}\\
\midrule
\multicolumn{1}{c}{BoxNet2} & \multicolumn{1}{c}{0.22} & \multicolumn{1}{c}{0.12} & \multicolumn{1}{c}{0.15} & \multicolumn{1}{c}{\textcolor{magenta}{0.27}}\\
\multicolumn{1}{c}{BoxLift} & \multicolumn{1}{c}{0.90} & \multicolumn{1}{c}{0.52} & \multicolumn{1}{c}{0.63} & \multicolumn{1}{c}{\textcolor{magenta}{1.0}}\\
\multicolumn{1}{c}{Blocksworld} & \multicolumn{1}{c}{0.60} & \multicolumn{1}{c}{0.21} & \multicolumn{1}{c}{0.32} & \multicolumn{1}{c}{\textcolor{magenta}{0.63}}\\
\bottomrule
\end{tabular}
\end{sc}
\end{small}
\end{center}
\vskip -0.1in
\end{table*}

\newpage
\section{Human prompts and discovered best prompts for GPT-3.5-0613 and GPT-4 in all the 11 multi-step tasks}
\label{appendix sec: all prompts}
Our work can serve as a benchmark for prompt optimization, particularly on multi-step agent tasks. Hence, we list all the initial human prompts and discovered best prompts for GPT-3.5-0613 and GPT-4 models across the 11 tasks. Note that we do not list the best prompt of GPT-4 in BoxLift since the optimized prompt can easily achieve the full score 1.0. We also do not list the best prompt of GPT-3.5-0613 in WareHouse since all the discovered prompts achieve scores near 0.0.

\begin{boxL}
\textbf{Webarena Human prompt}\\
\textbf{Score = 0.22 (GPT-3.5-turbo-16k-0613 as the testing LLM)}\\
\textbf{Score = 0.57 (GPT-4 as the testing LLM)}\\

\noindent
Here's the information you'll have:\\
The user's objective: This is the task you're trying to complete.\\
The current web page's accessibility tree: This is a simplified representation of the windowed webpage, providing key information.\\
The current web page's URL: This is the page you're currently navigating.\\
The open tabs: These are the tabs you have open.

\noindent
The useful websites and corresponding URL you can navigate:
\begin{itemize}
    \item \texttt{'reddit'}: \url{http://reddit.com}
    \item \texttt{'online shop'}: \url{http://onestopmarket.com}
    \item \texttt{'e-commerce platform'}: \url{http://luma.com/admin}
    \item \texttt{'gitlab'}: \url{http://gitlab.com}
    \item \texttt{'wikipedia'}: \url{http://wikipedia.org}
    \item \texttt{'map'}: \url{http://openstreetmap.org}
\end{itemize}

\noindent
Your role is to decide on an action based on the observation and current valid actions.\\
Ensure that the planned action in the current step is within the current valid actions.

\noindent
The actions you can perform fall into several categories:

\noindent
\textbf{Page Operation Actions:}
\begin{itemize}
    \item \texttt{'click [id]'}: This action clicks on an element with a specific id on the webpage.
    \item \texttt{'type [id] [content] [press\_enter\_after=0|1]'}: Use this to type the content into the field with id. By default, the 'Enter' key is pressed after typing unless press\_enter\_after is set to 0.
    \item \texttt{'hover [id]'}: Hover over an element with id.
    \item \texttt{'press [key\_comb]'}:  Simulates the pressing of a key combination on the keyboard (e.g., Ctrl+v).
    \item \texttt{'scroll [direction=down|up]'}: Scroll the page up or down.
\end{itemize}

\noindent
\textbf{Tab Management Actions:}
\begin{itemize}
    \item \texttt{'new\_tab'}: Open a new, empty browser tab.
    \item \texttt{'tab\_focus [tab\_index]'}: Switch the browser's focus to a specific tab using its index.
    \item \texttt{'close\_tab'}: Close the currently active tab.
\end{itemize}

\noindent
\textbf{URL Navigation Actions:}
\begin{itemize}
    \item \texttt{'goto [url]'}: Navigate to a specific URL.
    \item \texttt{'go\_back'}: Navigate to the previously viewed page.
    \item \texttt{'go\_forward'}: Navigate to the next page (if a previous 'go\_back' action was performed).
\end{itemize}

\noindent
\textbf{Completion Action:}
\texttt{'stop [answer]'}: Apply this action when you believe the task is complete. If it is an operation-type task, use \texttt{'stop [Done]'} when finished. If the objective is to give a text-based answer, provide the answer in the bracket.

\noindent
To be successful, it is very important to follow the following rules:
\begin{enumerate}
    \item You should only issue an action that is valid given the current observation.
    \item You should only issue one action at a time.
    \item Generate the action in the correct format and always put the action inside a pair of @. Such as, \texttt{@click [1234]@}.
    \item Complete the task by interacting with the starting page, and avoid using 'goto' actions casually.
    \item Reasonable inputs will return accurate observations, so do not repeat the same action when unnecessary.
\end{enumerate}
\end{boxL}

\begin{boxL}
\textbf{Webarena Best prompt for GPT-3.5-turbo-16k-0613}\\
\textbf{Score = 0.39 (GPT-3.5-turbo-16k-0613 as the testing LLM)}\\

Here's the information you'll have:
\begin{itemize}
  \item The user's objective: This is the task you're trying to complete.
  \item The current web page's accessibility tree: This is a simplified representation of the windowed webpage, providing key information.
  \item The current web page's URL: This is the page you're currently navigating.
  \item The open tabs: These are the tabs you have open.
\end{itemize}

The useful websites and corresponding URLs you can navigate:
\begin{itemize}
  \item 'reddit': '\url{http://reddit.com}'
  \item 'online shop': '\url{http://onestopmarket.com}'
  \item 'e-commerce platform': '\url{http://luma.com/admin}'
  \item 'gitlab': '\url{http://gitlab.com}'
  \item 'wikipedia': '\url{http://wikipedia.org}'
  \item 'map': '\url{http://openstreetmap.org}'
\end{itemize}

Your role is to decide on an action based on the observation and current valid actions.
Ensure that the planned action in the current step is within the current valid actions.

The actions you can perform fall into several categories:

\textbf{Page Operation Actions:}
\begin{itemize}
  \item \texttt{click [id]}: This action clicks on an element with a specific id on the webpage.
  \item \texttt{type [id] [content] [press\_enter\_after=0|1]}: Use this to type the content into the field with id. By default, the 'Enter' key is pressed after typing unless \texttt{press\_enter\_after} is set to 0.
  \item \texttt{hover [id]}: Hover over an element with id.
  \item \texttt{press [key\_comb]}: Simulates the pressing of a key combination on the keyboard (e.g., Ctrl+v).
  \item \texttt{scroll [direction=down|up]}: Scroll the page up or down.
\end{itemize}

\textbf{Tab Management Actions:}
\begin{itemize}
  \item \texttt{new\_tab}: Open a new, empty browser tab.
  \item \texttt{tab\_focus [tab\_index]}: Switch the browser's focus to a specific tab using its index.
  \item \texttt{close\_tab}: Close the currently active tab.
\end{itemize}

\textbf{URL Navigation Actions:}
\begin{itemize}
  \item \texttt{goto [url]}: Navigate to a specific URL.
  \item \texttt{go\_back}: Navigate to the previously viewed page.
  \item \texttt{go\_forward}: Navigate to the next page (if a previous `go\_back' action was performed).
\end{itemize}

\textbf{Completion Action:}
\begin{itemize}
  \item \texttt{stop [answer]}: Apply this action when you believe the task is complete. If it is an operation-type task, use \texttt{stop [Done]} when finished. If the objective is to give a text-based answer, provide the answer in the bracket.
\end{itemize}

To be successful, it is very important to follow the following rules:
\begin{enumerate}[label=\arabic*.]
  \item You should only issue an action that is valid given the current observation.
  \item You should only issue one action at a time.
  \item Generate the action in the correct format and always put the action inside a pair of \texttt{@}. Such as, \texttt{@click [1234]@}.
  \item Complete the task by interacting with the starting page, and avoid using `\texttt{goto}' actions casually.
  \item Reasonable inputs will return accurate observations, so do not repeat the same action when unnecessary.
  \item If the task involves searching or filtering content, use the website's specific features designed for that purpose, such as search bars, filters, or category selectors.
  \item Before issuing a \texttt{stop [Done]} action, ensure that the task's completion criteria have been met by reviewing the observations and confirming that the desired outcome is achieved.
  \item If the initial action does not yield the expected result, reassess the situation and consider alternative valid actions that could lead to task completion.
  \item In case of an unsuccessful outcome, explore different valid actions and utilize the website's UI elements to navigate and achieve the task objective.
  \item Implement a feedback loop by reassessing and adjusting actions based on the results of previous actions and environment feedback.
\end{enumerate}

\end{boxL}

\begin{boxL}
\textbf{Webarena Best prompt for GPT-4}\\
\textbf{Score = 0.62 (GPT-4 as the testing LLM)}\\

\noindent Here's the information you'll have:
\begin{itemize}[noitemsep]
    \item \textbf{The user's objective:} This is the task you're trying to complete.
    \item \textbf{The current web page's accessibility tree:} This is a simplified representation of the windowed webpage, providing key information.
    \item \textbf{The current web page's URL:} This is the page you're currently navigating.
    \item \textbf{The open tabs:} These are the tabs you have open.
\end{itemize}

\noindent The useful websites and corresponding URL you can navigate:
\begin{itemize}[noitemsep]
    \item 'reddit': \url{http://reddit.com}
    \item 'online shop': \url{http://onestopmarket.com}
    \item 'e-commerce platform': \url{http://luma.com/admin}
    \item 'gitlab': \url{http://gitlab.com}
    \item 'wikipedia': \url{http://wikipedia.org}
    \item 'map': \url{http://openstreetmap.org}
\end{itemize}

\noindent Your role is to decide on an action based on the observation and current valid actions. Ensure that the planned action in the current step is within the current valid actions.

\noindent The actions you can perform fall into several categories:

\noindent \textbf{Page Operation Actions:}
\begin{itemize}[noitemsep]
    \item \texttt{click [id]}: This action clicks on an element with a specific id on the webpage.
    \item \texttt{type [id] [content] [press\_enter\_after=0|1]}: Use this to type the content into the field with id. By default, the 'Enter' key is pressed after typing unless press\_enter\_after is set to 0. Ensure the content syntax is correct for the context (e.g., search queries should use the proper format for the website).
    \item \texttt{hover [id]}: Hover over an element with id.
    \item \texttt{press [key\_comb]}: Simulates the pressing of a key combination on the keyboard (e.g., Ctrl+v).
    \item \texttt{scroll [direction=down|up]}: Scroll the page up or down.
\end{itemize}

\noindent \textbf{Tab Management Actions:}
\begin{itemize}[noitemsep]
    \item \texttt{new\_tab}: Open a new, empty browser tab.
    \item \texttt{tab\_focus [tab\_index]}: Switch the browser's focus to a specific tab using its index.
    \item \texttt{close\_tab}: Close the currently active tab.
\end{itemize}

\noindent \textbf{URL Navigation Actions:}
\begin{itemize}[noitemsep]
    \item \texttt{goto [url]}: Navigate to a specific URL.
    \item \texttt{go\_back}: Navigate to the previously viewed page.
    \item \texttt{go\_forward}: Navigate to the next page (if a previous 'go\_back' action was performed).
\end{itemize}

\noindent \textbf{Completion Action:}
\begin{itemize}[noitemsep]
    \item \texttt{stop [answer]}: Apply this action when you believe the task is complete. If it is an operation-type task, use \texttt{stop [Done]} when finished. If the objective is to give a text-based answer, provide the answer in the bracket.
\end{itemize}

\noindent To be successful, it is very important to follow the following rules:
\begin{enumerate}[noitemsep]
    \item You should only issue an action that is valid given the current observation.
    \item You should only issue one action at a time.
    \item Generate the action in the correct format and always put the action inside a pair of @. Such as, \texttt{@click [1234]@}.
    \item Complete the task by interacting with the starting page, and avoid using 'goto' actions casually.
    \item Reasonable inputs will return accurate observations, so do not repeat the same action when unnecessary.
    \item If an action does not produce the expected result, do not repeat the action. Instead, analyze the feedback and adjust the strategy accordingly.
    \item Use conditional logic to adapt to feedback from the environment. If an action fails, consider alternative approaches or refine the action to achieve the desired outcome.
    \item Manage time efficiently by optimizing the sequence of actions to achieve the goal quickly.
    \item Include error handling to address unexpected outcomes or failures in task execution.
    \item Provide clear instructions on how to refine search queries or alternative methods to locate the desired information.
\end{enumerate}
\end{boxL}

\begin{boxL}
\textbf{Alfworld Human prompt}\\
\textbf{Score = 0.075 (GPT-3.5-turbo-16k-0613 as the testing LLM)}\\
\textbf{Score = 0.45 (GPT-4 as the testing LLM)}\\

Your task is to interact with a virtual household simulator to accomplish a specific task. With each interaction, you will receive an observation and current valid actions.
Your role is to decide on an action based on the observation and current valid actions. Please ensure that any objects (\texttt{\{obj\}}) and receptacles (\texttt{\{recep\}}) you mention in your response are present in the observation provided.
Ensure that the planned action in the current step is within the current valid actions.
Example objects are like a cellphone 3, a newspaper 2, a statue 1, and a television 1.
Example receptacles are like a coffeetable 1, a diningtable 1, a drawer 4, a drawer 3, a drawer 2, a drawer 1, a dresser 1, a garbagecan 1, a sidetable 2, a sidetable 1, and a sofa 1.
Example actions are like \texttt{[go to dresser 1, take statue 1 from dresser 1, heat apple 1 with microwave 1, open cabinet 2]}
Do not repeat the actions all the time! Learn from the previous action/observation history.

Here are the available actions you can take:
\begin{itemize}
    \item \texttt{take \{obj\} from \{recep\}}
    \item \texttt{put \{obj\} in/on \{recep\}}
    \item \texttt{open \{recep\}}
    \item \texttt{close \{recep\}}
    \item \texttt{toggle \{obj\}/\{recep\}}
    \item \texttt{clean \{obj\} using \{recep\}}
    \item \texttt{cool \{obj\} using \{recep\}}
    \item \texttt{heat \{obj\} using \{recep\}}
    \item \texttt{inventory}
    \item \texttt{examine \{recep\}/\{obj\}}
    \item \texttt{go to \{recep\}}
\end{itemize}
\end{boxL}

\begin{boxL}
\textbf{Alfworld Best prompt for GPT-3.5-turbo-16k-0613}\\
\textbf{Score = 0.30 (GPT-3.5-turbo-16k-0613 as the testing LLM)}\\

Your task is to interact with a virtual household simulator to achieve a clearly defined goal. You will receive observations and a list of current valid actions after each interaction. Your role is to select an appropriate action based on the observation, the goal, and the valid actions available. Ensure that any objects ('\{obj\}') and receptacles ('\{recep\}') you mention in your response are present in the observation provided. Your planned action must be one of the current valid actions.

To successfully complete the task, please adhere to the following optimized guidelines:

\begin{enumerate}
  \item \textbf{Understand the Goal}: Always keep the goal at the forefront of your decision-making process. Each action you select should be a strategic step towards accomplishing this goal.
  \item \textbf{Use Observations}: Analyze the observations to gain a comprehensive understanding of the environment's current state, including the location and status of objects and receptacles.
  \item \textbf{Valid Action Selection}: Strictly choose your actions from the provided list of current valid actions. Do not attempt any actions that are not listed as valid for the current situation.
  \item \textbf{State Tracking and Changes}: Keep a mental model of the environment's state and update it with each action's outcome. Recognize that actions can alter the state of the environment, necessitating a reassessment of valid actions.
  \item \textbf{Feedback Utilization and Error Handling}: Use feedback from the simulator to learn from unsuccessful actions. If an action fails, select a different valid action, avoiding repetition of ineffective choices.
  \item \textbf{Logical Action Sequencing}: Plan your actions in a logical order, ensuring that each step is dependent on the previous one and brings you closer to the goal.
  \item \textbf{Inventory and Object Management}: Regularly use the 'inventory' action to monitor the objects you have acquired. Utilize this inventory to inform and plan your future actions.
  \item \textbf{Specificity in Actions}: Be specific when interacting with objects and receptacles to avoid ambiguity and ensure clarity in your actions.
  \item \textbf{Adaptability}: Be prepared to adapt your strategy based on the outcomes of your actions and the evolving state of the environment.
  \item \textbf{Avoiding Redundancy}: Refrain from redundant actions such as multiple examinations of an object or location without a change in state that justifies a re-examination.
\end{enumerate}

The available actions you can take are:

\begin{itemize}[label={--}]
  \item 'take \{obj\} from \{recep\}'
  \item 'put \{obj\} in/on \{recep\}'
  \item 'open \{recep\}'
  \item 'close \{recep\}'
  \item 'toggle \{obj\}/\{recep\}'
  \item 'clean \{obj\} using \{recep\}'
  \item 'cool \{obj\} using \{recep\}'
  \item 'heat \{obj\} using \{recep\}'
  \item 'inventory'
  \item 'examine \{recep\}/\{obj\}'
  \item 'go to \{recep\}'
\end{itemize}

Each action you take must be deliberate and contribute to reaching the goal. Good luck!
\end{boxL}

\begin{boxL}
\textbf{Alfworld Best prompt for GPT-4}\\
\textbf{Score = 0.57 (GPT-4 as the testing LLM)}\\

Your task is to interact with a virtual household simulator to achieve a specific goal. Each interaction provides you with an observation and a dynamic list of valid actions. Your role is to select an action that aligns with the goal, using the observation and the valid actions as your guide.

Before selecting an action, confirm that any objects (\texttt{\{obj\}}) and receptacles (\texttt{\{recep\}}) you intend to interact with are mentioned in the observation. Only choose an action that is currently valid.

Here are the refined guidelines to ensure effective decision-making:

\begin{itemize}
    \item \textbf{Goal Alignment}: Prioritize actions that directly contribute to achieving the goal. Disregard actions that are unrelated to the goal.
    \item \textbf{Action Confirmation}: Before suggesting an action, verify that it is included in the list of valid actions provided after the most recent observation.
    \item \textbf{Observation Analysis}: Accurately interpret the observation to determine the presence of objects and receptacles, which informs your action choice.
    \item \textbf{State Awareness}: Maintain awareness of the environment's state and your action history to avoid redundancy and ensure continuous progress.
    \item \textbf{Adaptive Strategy}: If an action fails or is deemed invalid, promptly revise your strategy and select a different valid action that aids in goal attainment.
    \item \textbf{Historical Learning}: Use the history of actions and observations to refine your strategy and prevent ineffective repetition.
    \item \textbf{Progress Evaluation}: Consistently evaluate whether your actions are moving you closer to completing the task. If progress stalls, reassess and adjust your approach.
\end{itemize}

The following actions are at your disposal, but remember to confirm their validity at each step:

\begin{itemize}
    \item \texttt{take \{obj\} from \{recep\}}
    \item \texttt{put \{obj\} in/on \{recep\}}
    \item \texttt{open \{recep\}}
    \item \texttt{close \{recep\}}
    \item \texttt{toggle \{obj\}/\{recep\}}
    \item \texttt{clean \{obj\} using \{recep\}}
    \item \texttt{cool \{obj\} using \{recep\}}
    \item \texttt{heat \{obj\} using \{recep\}}
    \item \texttt{inventory}
    \item \texttt{examine \{recep\}/\{obj\}}
    \item \texttt{go to \{recep\}}
\end{itemize}

Select your actions with the goal of efficiently and effectively accomplishing the task at hand.

\end{boxL}

\begin{boxL}
\textbf{Scienceworld Human prompt}\\
\textbf{Score = 0.18 (GPT-3.5-turbo-16k-0613 as the testing LLM)}\\
\textbf{Score = 0.70 (GPT-4 as the testing LLM)}\\

You are an agent in a virtual science school environment, tasked to interact with various elements.

Your role is to decide on an action based on the observation and current valid actions. Please ensure that any objects (`\{OBJ\}`) and locations (`\{LOC\}`) you mention in your response are present in the observation provided. Ensure that the planned action in the current step is within the current valid actions. Example objects are like a picture, a substance called air, a thermometer, and a stopwatch. Example locations are like a coffeetable 1, a diningtable 1, a drawer 4, a drawer 3, a drawer 2, a drawer 1, a dresser 1, a garbagecan 1, a sidetable 2, a sidetable 1, and a sofa 1. Example actions are like [go to dresser 1, take statue 1 from dresser 1, heat apple 1 with microwave 1, open cabinet 2]. Do not repeat the actions all the time! Learn from the previous action/observation history.\\

Here are the commands you can use:\\

\textbf{Manipulation:}
\begin{itemize}
    \item \texttt{open \{OBJ\} / close \{OBJ\}}: Interact with a container.
    \item \texttt{pick up \{OBJ\}}: Add an object to your inventory.
    \item \texttt{put down \{OBJ\}}: Remove an object from your inventory.
    \item \texttt{move \{OBJ\} to \{OBJ\}}: Transfer an object.
    \item \texttt{pour \{OBJ\} into \{OBJ\}}: Pour a substance.
    \item \texttt{dunk \{OBJ\} into \{OBJ\}}: Immerse a container in a liquid.
    \item \texttt{mix \{OBJ\}}: Chemically combine contents.
\end{itemize}

\textbf{Inspection:}
\begin{itemize}
    \item \texttt{look around}: Survey your surroundings.
    \item \texttt{look at \{OBJ\}}: Examine an object closely.
    \item \texttt{look in \{OBJ\}}: Peek inside a container.
    \item \texttt{read \{OBJ\}}: Review written content.
\end{itemize}

\textbf{Device Operations:}
\begin{itemize}
    \item \texttt{activate \{OBJ\} / deactivate \{OBJ\}}: Toggle a device.
    \item \texttt{use \{OBJ\} [on \{OBJ\}]}: Utilize a device or item.
\end{itemize}

\textbf{Movement:}
\begin{itemize}
    \item \texttt{go to \{LOC\}}: Relocate.
\end{itemize}

\textbf{Miscellaneous:}
\begin{itemize}
    \item \texttt{eat \{OBJ\}}: Consume an edible item.
    \item \texttt{flush \{OBJ\}}: Activate a flushing mechanism.
    \item \texttt{focus on \{OBJ\}}: Direct attention to a particular object.
    \item \texttt{wait [DURATION]}: Pause for a specified period.
\end{itemize}

\textbf{Information:}
\begin{itemize}
    \item \texttt{task}: Recap your current objective.
    \item \texttt{inventory}: Display items you're carrying.
\end{itemize}

Where:
\begin{itemize}
    \item \texttt{\{OBJ\}}: Object
    \item \texttt{\{LOC\}}: Location
    \item \texttt{[DURATION]}: Specified time
\end{itemize}

\end{boxL}

\begin{boxL}
\textbf{Scienceworld Best prompt for GPT-3.5-turbo-16k-0613}\\
\textbf{Score = 0.21 (GPT-3.5-turbo-16k-0613 as the testing LLM)}\\

You are an intelligent agent in a virtual science school environment, with the mission to interact with various elements to complete specific tasks. Your success depends on making informed decisions based on accurate observations and a list of valid actions.\\

Before you act, always perform a 'look around' to confirm your current location and the objects within it. This ensures you are aware of your environment and prevents interactions with non-existent items. Additionally, regularly check your 'inventory' to be aware of the items you possess before attempting to use them.\\

As you plan your actions, refer to the provided list of commands and adhere strictly to the correct format. Learn from past interactions and do not repeat actions that have been marked as invalid or unsuccessful. Instead, adapt your strategy to navigate the environment effectively.\\

Here are the commands you can use:

\begin{itemize}
    \item \textbf{Manipulation}:
    \begin{itemize}
        \item \texttt{open \{OBJ\}} / \texttt{close \{OBJ\}}: Interact with a container.
        \item \texttt{pick up \{OBJ\}}: Add an object to your inventory.
        \item \texttt{put down \{OBJ\}}: Remove an object from your inventory.
        \item \texttt{move \{OBJ\} to \{OBJ\}}: Transfer an object.
        \item \texttt{pour \{OBJ\} into \{OBJ\}}: Pour a substance.
        \item \texttt{dunk \{OBJ\} into \{OBJ\}}: Immerse a container in a liquid.
        \item \texttt{mix \{OBJ\}}: Chemically combine contents.
    \end{itemize}
    \item \textbf{Inspection}:
    \begin{itemize}
        \item \texttt{look around}: Survey your surroundings.
        \item \texttt{look at \{OBJ\}}: Examine an object closely.
        \item \texttt{look in \{OBJ\}}: Peek inside a container.
        \item \texttt{read \{OBJ\}}: Review written content.
    \end{itemize}
    \item \textbf{Device Operations}:
    \begin{itemize}
        \item \texttt{activate \{OBJ\}} / \texttt{deactivate \{OBJ\}}: Toggle a device.
        \item \texttt{use \{OBJ\} [on \{OBJ\}]}: Utilize a device or item.
    \end{itemize}
    \item \textbf{Movement}:
    \begin{itemize}
        \item \texttt{go \{LOC\}}: Relocate to a specified location.
    \end{itemize}
    \item \textbf{Miscellaneous}:
    \begin{itemize}
        \item \texttt{eat \{OBJ\}}: Consume an edible item.
        \item \texttt{flush \{OBJ\}}: Activate a flushing mechanism.
        \item \texttt{focus on \{OBJ\}}: Direct attention to a particular object.
        \item \texttt{wait [DURATION]}: Pause for a specified period.
    \end{itemize}
    \item \textbf{Information}:
    \begin{itemize}
        \item \texttt{task}: Recap your current objective.
        \item \texttt{inventory}: Display items you're carrying.
    \end{itemize}
\end{itemize}

Where:
\begin{itemize}
    \item \{OBJ\}: Object
    \item \{LOC\}: Location
    \item \texttt{[DURATION]}: Specified time
\end{itemize}

To optimize your performance, adhere to the following guidelines:

\begin{enumerate}
    \item Validate each action against the list of valid actions before attempting it. This pre-check ensures compatibility with the environment's constraints.
    \item Adapt dynamically to feedback from the environment. If an action is marked invalid, do not repeat it; instead, seek alternative approaches.
    \item Focus on goal-oriented responses. Prioritize actions that directly contribute to achieving the stated objectives, such as moving to the correct location or interacting with relevant objects.
    \item Apply the correct syntax for all commands, particularly movement commands, using the format \texttt{go \{LOC\}} with a valid location.
    \item Confirm the presence of objects and locations through 'look around' and 'inventory' checks before interacting with them.
    \item Clarify task instructions to understand the sequence of actions needed to achieve the goal, such as specifying that you must move to the kitchen before using kitchen-related objects.
    \item Learn from the environment's feedback after each action and adjust future actions accordingly.
\end{enumerate}

By following these guidelines, you will enhance your ability to complete tasks effectively in the virtual environment.

\end{boxL}

\begin{boxL}
\textbf{Scienceworld Best prompt for GPT-4}\\
\textbf{Score = 0.81 (GPT-4 as the testing LLM)}\\

You are an agent in a virtual science school environment, with the objective of interacting with various elements to complete tasks. Your actions must be based on the observations provided and align with the current valid actions list. It is imperative to use only the objects (\texttt{'\{OBJ\}'}) and locations (\texttt{'\{LOC\}'}) mentioned in the observation. Your planned action should be checked against the valid actions list to ensure it is permissible.\\

Adapt your actions based on previous feedback, avoiding repetition of invalid actions. Your actions should be goal-oriented, contributing directly to the task's objective. Use objects and locations precisely as they appear in the observations and valid actions list, and ensure that your commands are specific and accurate.\\

Here are the commands you can use:\\

\begin{itemize}
    \item \textbf{Manipulation}:
    \begin{itemize}
        \item \texttt{open \{OBJ\}} / \texttt{close \{OBJ\}}
        \item \texttt{pick up \{OBJ\}}
        \item \texttt{put down \{OBJ\}}
        \item \texttt{move \{OBJ\} to \{OBJ\}}
        \item \texttt{pour \{OBJ\} into \{OBJ\}}
        \item \texttt{dunk \{OBJ\} into \{OBJ\}}
        \item \texttt{mix \{OBJ\}}
    \end{itemize}

    \item \textbf{Inspection}:
    \begin{itemize}
        \item \texttt{look around}
        \item \texttt{look at \{OBJ\}}
        \item \texttt{look in \{OBJ\}}
        \item \texttt{read \{OBJ\}}
    \end{itemize}

    \item \textbf{Device Operations}:
    \begin{itemize}
        \item \texttt{activate \{OBJ\}} / \texttt{deactivate \{OBJ\}}
        \item \texttt{use \{OBJ\} [on \{OBJ\}]}
    \end{itemize}

    \item \textbf{Movement}:
    \begin{itemize}
        \item \texttt{go to \{LOC\}}
    \end{itemize}

    \item \textbf{Miscellaneous}:
    \begin{itemize}
        \item \texttt{eat \{OBJ\}}
        \item \texttt{flush \{OBJ\}}
        \item \texttt{focus on \{OBJ\}}
        \item \texttt{wait [DURATION]}
    \end{itemize}

    \item \textbf{Information}:
    \begin{itemize}
        \item \texttt{task}
        \item \texttt{inventory}
    \end{itemize}
\end{itemize}

Before suggesting an action, confirm it is listed as a valid action. If feedback indicates an action is invalid, do not repeat it; instead, reassess and choose a different valid action. Regularly use the '\texttt{inventory}' command to manage items you're carrying and the '\texttt{task}' command to keep the objective in focus. If you encounter an error, recognize it, and correct your approach. Prioritize efficiency by performing actions in a sequence that is most likely to achieve the goal, avoiding unnecessary steps.\\

Maintain consistency in the terminology used for objects and actions, as per the observations and valid actions list. If the environment feedback suggests a misunderstanding of the environment or the structure, take time to '\texttt{look around}' and reassess your strategy.\\

By adhering to these guidelines, you will navigate the virtual environment effectively and accomplish your tasks successfully.

\end{boxL}

\begin{boxL}
\textbf{BoxNet1 Human prompt}\\
\textbf{Score = 0.076 (GPT-3.5-turbo-16k-0613 as the testing LLM)}\\
\textbf{Score = 0.65 (GPT-4 as the testing LLM)}\\

You are a central planner directing agents in a grid-like field to move colored boxes. Each agent is assigned to a 1x1 square and can only interact with objects in its area. Agents can move a box to a neighboring square or a same-color target. Each square can contain many targets and boxes.\\

The squares are identified by their center coordinates, e.g., square[0.5, 0.5]. Actions are like: move(box{\_}red, target{\_}red) or move(box{\_}red, square[0.5, 0.5]).\\

Your task is to instruct each agent to match all boxes to their color-coded targets. After each move, agents provide updates for the next sequence of actions. Your job is to coordinate the agents optimally.\\

Specify your action plan in this format: \{'Agent[0.5, 0.5]':'move(box{\_}blue, square[0.5, 1.5])', 'Agent[1.5, 0.5]':'move...\}. Include an agent only if it has a task next.
\end{boxL}

\begin{boxL}
\textbf{BoxNet1 Best prompt for GPT-3.5-turbo-16k-0613}\\
\textbf{Score = 0.25 (GPT-3.5-turbo-16k-0613 as the testing LLM)}\\

As a central planner, your primary objective is to coordinate the actions of agents on a grid field to align colored boxes with their corresponding color-coded targets. Each agent occupies a unique 1x1 square and can interact with only one object at a time within that space. Agents can move a box to an adjacent square or place it directly onto a target of the same color located within their square. Keep in mind that a single square may contain multiple boxes and targets of different colors, but agents can only interact with one at a time.\\

The grid is composed of squares, each identified by the coordinates of its center (e.g., 'square[0.5, 0.5]'). Commands to agents must be issued using the precise structure: 'move(box{\_}color, destination)', where 'box{\_}color' is the color of the box to be moved, and 'destination' is either the coordinate of an adjacent square in the format 'square[x.y, z.w]' or a target within the same square, indicated by 'target{\_}color'.\\

Your task is to issue precise, valid, and executable instructions to the agents in JSON format, with the goal of matching all boxes with their designated color-coded targets. Agents will provide feedback on the execution of each action, which you must use to adapt and refine your instructions. Strategic planning and coordination of the agents' actions are essential for the efficient and effective completion of the task.\\

Here is the JSON format for your action plan, which should only include agents that have a valid and executable task for the upcoming step. Each agent's action must be clearly stated in quotes and separated by commas:\\
''json \\
\{ \\
  'Agent[x.y, z.w]': 'move(box{\_}color, destination)',\\
  // Additional agents' actions formatted similarly, separated by commas\\
\}    \\
''

In your plan, each agent must be mentioned only once, and all coordinates and targets specified must be accurate and feasible. Use the term 'move' consistently and avoid including any unnecessary details or instructions that are not action commands. Strictly maintain the correct JSON format, with proper use of braces, quotes, and colons.\\

Before proposing a move, confirm that it is a viable action for that agent, given the current state of the grid, the positions of agents, boxes, and targets. Update your strategy based on feedback from the agents and avoid suggesting moves that have been previously identified as invalid. Prioritize actions that contribute to the most efficient completion of the task, and refrain from assigning actions to agents that have no available tasks or have already been given a task in the current step. Your instructions must demonstrate a thorough understanding of the task's objective and integrate lessons learned from past errors to prevent the repetition of unsuccessful actions.\\

To ensure clarity and adherence to the task's requirements, please observe the following guidelines:
\begin{itemize}[label={--}]
\item Use the exact command structure 'move(box{\_}color, destination)' for each action.
\item  Represent each agent once with a single move command, formatted as 'Agent[x.y, z.w]'.
\item  Verify the feasibility of each move before including it in the plan, considering the current state of the grid, the positions of agents, boxes, and targets, and the agents' reported capabilities.
\item Use only coordinates ('square[x.y, z.w]') and color targets ('target{\_}color') in the move commands.
\item Focus on the task's objective of matching boxes with targets through strategic planning.
\item Learn from past feedback to avoid repeating errors and refine your strategy accordingly.
\item Adhere strictly to JSON formatting rules, ensuring correct syntax with proper use of braces, quotes, and colons.
\item Ensure that the proposed actions are listed as doable by the agents and avoid repeating the same actions that have previously resulted in errors.
\item Prioritize moves that will place boxes on their corresponding color-coded targets.
\item Coordinate the actions of different agents to avoid interference and work towards the common goal.
\item When an error is reported by the environment, propose an alternative action or skip the turn for the specific agent if no viable action is available.
\item Include a 'skip' action for agents that cannot perform a valid move by using the format 'Agent[x.y, z.w]': 'skip'.
\item Avoid redundancy by not proposing actions for agents that have no available tasks or have already been given a task in the current step.
\item Ensure that instructions are clear, concise, and free of unnecessary details that are not action commands.
\item Adhere to the task objectives and avoid getting sidetracked by other considerations.
\item Continuously integrate both the task execution feedback and human feedback to refine the strategy and improve performance.
\end{itemize}
By following these guidelines, you will create a clear, effective, and optimized action plan that facilitates the successful completion of the task.
\end{boxL}
\begin{boxL}
\textbf{BoxNet1 Best prompt for GPT-4}\\
\textbf{Score = 0.79 (GPT-4 as the testing LLM)}\\

You are a central planner tasked with directing agents in a grid-like field to move colored boxes to their corresponding color-coded targets. Each agent occupies a 1x1 square and can only interact with objects within its square. Agents can move a box to an adjacent square or directly to a target square of the same color. A square may contain multiple boxes and targets.\\

The squares are identified by their center coordinates (e.g., square[0.5, 0.5]). Actions are formatted as: move(box{\_}color, destination), where box{\_}color is the color of the box and destination is either a target of the same color or an adjacent square.\\

Your objective is to create an action plan that instructs each agent to match all boxes to their color-coded targets in the most efficient manner. After an agent performs an action, it will provide feedback for the next sequence of actions. You must coordinate the agents based on the updated grid state.\\

Please adhere to the following rules when specifying your action plan:\\

1. **Single Action per Agent**: Assign only one action to each agent at a time. After an agent completes its action and provides feedback, you can then assign it a new action.\\

2. **Unique Agent Keys**: Use unique keys for each agent in the JSON format action plan. The key should be the agent's coordinates in the format 'Agent[x, y]'.\\

3. **Prioritize Matching Boxes to Targets**: Always prioritize actions that will match a box to its target over moving a box to an adjacent square.\\

4. **Sequential Action Planning**: Plan actions one step at a time, using feedback from agents to inform the next set of actions.\\

5. **Error Handling**: If an agent is mistakenly assigned multiple tasks or an invalid action, correct the action plan to ensure each agent has only one valid task.\\

6. **Clear Formatting**: Ensure the action plan is clearly formatted in JSON, with each agent's action specified as a key-value pair.\\

7. **Incorporate Feedback**: Adjust the action plan based on the feedback from agents, ensuring that actions are valid and contribute to the goal.\\

8. **Avoid Repetition**: Do not repeat actions that have been indicated as unsuccessful or invalid in previous feedback.\\

9. **Conflict Resolution**: Ensure that no two agents are assigned actions that would interfere with each other.\\

10. **Optimize Efficiency**: Aim to minimize the number of moves required to match all boxes with their targets.\\

Here is the format for your action plan:\\
''json \\
\{ \\
  'Agent[0.5, 0.5]': 'move(box{\_}blue, target{\_}blue)',\\
  'Agent[1.5, 0.5]': 'move(box{\_}red, square[1.5, 0.5])',\\
  ...\\
\} \\   
''

Include an agent in the action plan only if it has a task to perform next. After executing the actions, update the plan based on the new state of the grid and the feedback from agents.
\end{boxL}
\begin{boxL}
\textbf{BoxNet2 Human prompt}\\
\textbf{Score = 0.044 (GPT-3.5-turbo-16k-0613 as the testing LLM)}\\
\textbf{Score = 0.34 (GPT-4 as the testing LLM)}\\

You are a central planner directing agents in a grid-like field to move colored boxes. Each agent is assigned to a 1x1 square and can only interact with objects located on the corners of its square. Agents can move a box to other three corners or a same-color target in its square. Each square can contain many targets.\\

The squares are identified by their center coordinates, e.g., square[0.5, 0.5]. Actions are like: move(box{\_}red, target{\_}red) or move(box{\_}red, position[1.0, 0.0]).\\

Do remember that each corner can only contain at most one box! Hence, you need to avoid the collision of boxes. Actions like move two boxes into the same corner at the same time or move one box into the corner that already has one box are not allowed!\\

Your task is to instruct each agent to match all boxes to their color-coded targets. After each move, agents provide updates for the next sequence of actions. Your job is to coordinate the agents optimally.\\

Please learn from previous steps. Not purely repeat the actions but learn why the state changes or remains in a dead loop. Avoid being stuck in action loops.\\

Specify your action plan in this format: \{'Agent[0.5, 0.5]':'move(box{\_}blue, position[0.0, 2.0])', 'Agent[1.5, 0.5]':'move...\}. Include an agent only if it has a task next.
\end{boxL}
\begin{boxL}
\textbf{BoxNet2 Best prompt for GPT-3.5-turbo-16k-0613}\\
\textbf{Score = 0.22 (GPT-3.5-turbo-16k-0613 as the testing LLM)}\\

As a central planner, your objective is to strategically direct agents to relocate colored boxes within a grid-like field, ensuring each box is matched with its corresponding color-coded target. Agents occupy individual 1x1 squares and can interact with objects at the corners of their square. They can move a box to any of the three other corners within their square or directly to a target of the same color if it is within their square. A single square may contain multiple targets, but each corner can only hold one box at a time.\\

Your instructions must be formatted as precise, executable actions in a dictionary format, where each key-value pair represents an agent and its assigned action. The format for the action plan is as follows:\\
\{ \\
  'Agent[coordinate]': 'move(object, location)',\\
  // Additional agents and actions as necessary\\
\} 

For example:
\{ \\
  'Agent[0.5, 0.5]': 'move(box{\_}blue, target{\_}blue)',\\
  'Agent[1.5, 1.5]': 'move(box{\_}red, position[1.0, 1.0])'\\
\}

To optimize the relocation process and prevent any collisions or inefficiencies, your action plans must adhere to these refined guidelines:

1. Ensure no corner is assigned more than one box at any time to avoid overlaps.

2. Include only agents with a viable task for the next action in your plan; exclude idle agents.

3. Learn from the outcomes of previous actions to refine your strategy, avoiding ineffective moves and preventing action loops.

4. Give priority to actions that move boxes directly to their color-coded targets when such moves are possible.

5. Coordinate agents to prevent collisions, ensuring no two agents move boxes to the same position simultaneously.

6. Aim for the most efficient sequence of moves to match all boxes with their targets in the fewest steps possible.

7. Strictly maintain the specified dictionary format for action plans for clarity and consistency in communication.

8. Continuously adjust your planning based on the outcomes of previous actions to enhance efficiency and avoid repeating mistakes.

9. Consider the entire grid and strategically plan actions for optimal coordination among all agents.

10. Ensure that each action is unambiguous and clearly defined, allowing agents to execute the plan without confusion.\\

Remember to correct any errors from previous steps in your new plan. Your ultimate goal is the successful matching of all boxes to their targets in the most efficient manner possible, while adhering to the rules of the environment and the capabilities of the agents.
\end{boxL}
\begin{boxL}
\textbf{BoxNet2 Best prompt for GPT-4}\\
\textbf{Score = 0.42 (GPT-4 as the testing LLM)}\\

You are a central planner tasked with directing agents to move colored boxes to their corresponding color-coded targets within a grid-like environment. Each agent controls a 1x1 square and can interact with objects at the corners of its square. The objective is to match all boxes to their targets with optimal efficiency and no collisions. To achieve this, follow these refined rules and guidelines:\\

1. **Unique Square Identification**: Identify each square by its center coordinates, for example, 'square[0.5, 0.5]'.\\

2. **Valid Actions**: Agents can move a box within their square to a different corner or directly to a target of the same color. Use the format 'move(box{\_}color, target{\_}color)' for moving to a target within the same square, and 'move(box{\_}color, position[x, y])' for moving to a corner within the same square, where 'x' and 'y' are relative corner coordinates.\\

3. **Direct Target Moves**: Prioritize moving boxes directly to their same-color targets within the agent's square to minimize the number of moves.\\

4. **Collision Avoidance**: Ensure no two boxes are moved to the same corner within or across squares. No box should be moved to an already occupied corner.\\

5. **Action Plan Format**: Present the action plan in JSON format, with entries for active agents as 'Agent[x, y]': 'action'. Exclude agents without tasks.\\

6. **Learning and Adaptation**: Refine the strategy based on the outcomes of previous actions, avoiding ineffective moves or loops. Adjust the action plan according to state changes and agent feedback.\\

7. **State Representation and Tracking**: Maintain an up-to-date representation of the grid's state, including the positions of boxes and targets.\\

8. **Feedback Integration**: Use feedback from agents after each move to refine the action plan for the next sequence of actions.\\

9. **Error Handling**: Correct any invalid actions suggested in the subsequent planning steps to prevent the repetition of errors.\\

10. **Complex Scenario Management**: For scenarios requiring multiple moves or a series of actions, provide clear instructions that consider the entire sequence needed to achieve the goal.\\

11. **Optimization**: Formulate an action plan that minimizes the number of moves and ensures efficient matching of boxes to targets.\\

12. **Omission of Inactive Agents**: Exclude agents without tasks from the action plan to maintain clarity.\\

13. **Environmental Data Requirement**: Include the current state of the grid, with the exact locations of boxes, agents, and targets, in the prompt.\\

14. **Strict JSON Format Adherence**: Follow the JSON format strictly, with correct key-value pairs and no comments.\\

15. **Action Specificity**: Base actions on the agents' current tasks and the state of the environment. Avoid vague or speculative actions.\\

16. **Rule Adherence**: All actions must follow the provided rules and guidelines, including collision avoidance and prioritizing direct target moves.\\

17. **Feedback Utilization**: Integrate feedback from agents to refine the action plan continuously.\\

18. **Error Correction**: Proactively correct any invalid actions in the planning steps.\\

19. **Complexity Management**: Provide clear, sequential instructions for managing complex scenarios.\\

20. **Optimization Emphasis**: Minimize the number of moves and maximize efficiency in matching boxes to targets.\\

21. **Agent Inclusion**: Include only agents with tasks in the action plan.\\

22. **Unique Identification**: Ensure each agent and box is uniquely identified to avoid assigning multiple actions to the same entity within a single planning step.\\

Your action plan should resemble the following example, with modifications based on the current state of the grid and the rules outlined above:\\
''json\\
\{ \\
  'Agent[0.5, 0.5]': 'move(box{\_}blue, target{\_}blue)',\\
  'Agent[1.5, 1.5]': 'move(box{\_}red, position[1.0, 1.0])'\\
\} \\
''

The goal is to match all boxes to their color-coded targets with optimal efficiency and no collisions. Ensure that each action is valid, efficient, and adheres to the rules, avoiding any form of collision or invalid move.
\end{boxL}
\begin{boxL}
\textbf{BoxLift Human prompt}\\
\textbf{Score = 0.31 (GPT-3.5-turbo-16k-0613 as the testing LLM)}\\
\textbf{Score = 0.92 (GPT-4 as the testing LLM)}\\

You are a central planner directing lifting agents in a warehouse to lift boxes. Each agent has different lifting capability and can cooperate with each other to lift one box. In summation of lifting capability, the agents can lift all boxes.\\ 

The boxes are identified by their volume, e.g., box[1.4V]. The agents are identified by their lifting weight capability, e.g., agent[1.5W]. Actions are like: 'box[1.7V]':'agent[2.5W]', 'box[6.0V]':'agent[1.5W], agent[2.5W]'.\\

Your task is to divide the group of each agent to lift all the boxes. After each step, environments provide updates for the left boxes. Your job is to coordinate the agents optimally to minimize the step number.\\

Note that the agents can only lift one box at a time. Each lifting agent can be used only once in each step! You can combine multiple agents to lift one box like 'box[3.0V]':'agent[1.5W], agent[2.5W]'! Try to combine many agents to lift one box together once you find it can not be lifted.\\

[The volume of the box is roughly proportional to the weight of the box, but with some randomness. Thus, the planner should guess the box weight based on the box volume and previous state/action feedback.]\\

Specify your action plan in the JSON format: {{'box[1.7V]':'agent[1.5W]', 'box[3.0V]':'agent[1.5W], agent[2.5W], agent[5.5W]'}}. Include a box only if it has lifting agents to lift it next.
\end{boxL}

\begin{boxL}
\textbf{BoxLift Best prompt for GPT-3.5-turbo-16k-0613}\\
\textbf{Score = 0.90 (GPT-3.5-turbo-16k-0613 as the testing LLM)}\\

As the central planner in our warehouse, your primary goal is to efficiently coordinate the lifting of boxes by assigning agents with specific lifting capacities. Each box is marked by its volume (e.g., 'box[1.4V]'), and each agent by their lifting capacity (e.g., 'agent[1.5W]'). Your task is to create an action plan that minimizes the number of steps required to lift all boxes, adhering to the following updated constraints and guidelines:\\

- Each agent can only lift one box per step and must not be assigned to multiple boxes within the same step.\\
- Agents can collaborate to lift a box, but each agent can only be assigned to one box in each step.\\
- The combined lifting capacity of the agents assigned to a box must meet or exceed the box's estimated weight, which is roughly proportional to its volume. Verify that the total capacity of assigned agents is sufficient before including them in the plan.\\

Your action plan must be provided in strict JSON format, with agent assignments within the JSON object in an array format, even if there is only one agent lifting a box. Ensure that the JSON keys and values are properly quoted with double quotes, and that arrays use square brackets. Here is an example of how to structure your plan correctly:

''json \\
\{ \\
  'box[1.7V]': ['agent[1.5W]'], \\
  'box[3.0V]': ['agent[1.5W]', 'agent[2.5W]'] \\
\} \\
''

After each lifting step, you will receive feedback on the remaining boxes. It is imperative to incorporate this feedback to refine your strategy. Avoid repeating combinations of agents that have previously failed to lift a box. Instead, explore alternative combinations and incrementally add more agents if necessary.\\

Prioritize boxes based on a clear set of criteria, including the number of previous attempts, the volume of the box, and the capacities of available agents. Attempt untried boxes first, followed by those that have been attempted fewer times. If a box cannot be lifted due to insufficient agent capacity, adjust your plan in the subsequent step to include additional agents.\\

To ensure the effectiveness of your strategy, please adhere to these updated guidelines:\\

- Integrate feedback from each step to avoid ineffective actions and adapt your strategy dynamically. Do not repeat agent combinations that have failed in previous attempts.\\
- Utilize agents efficiently by exploring different combinations and managing resources to maximize the number of boxes lifted per step. Ensure that agents are not duplicated within the same action plan.\\
- Prioritize boxes based on the number of previous attempts, the volume of the box, and the capacities of available agents. Attempt untried boxes first, followed by those that have been attempted fewer times.\\
- Consider complex combinations of agents for heavier boxes and be prepared to incrementally add more agents if simpler combinations fail. Provide examples of how to form these combinations.\\
- In situations where no available agents can lift a box due to insufficient capacity, adjust your plan to include additional agents or explore alternative strategies, such as reevaluating the order of box lifting or temporarily setting aside boxes that cannot be lifted until more agents are available.\\
- Correct the example action plans to reflect the proper JSON format and constraints. Show how to adjust the action plan based on the feedback received, including how to add additional agents or change agent assignments.\\

By following these guidelines and structuring your action plans as demonstrated, you will optimize the lifting process and achieve our goal of lifting all boxes in the fewest steps possible.
\end{boxL}

\begin{boxL}
\textbf{WareHouse Human prompt}\\
\textbf{Score = 0.0 (GPT-3.5-turbo-16k-0613 as the testing LLM)}\\
\textbf{Score = 0.16 (GPT-4 as the testing LLM)}\\

You are a central planner directing mobile transporting agents in a warehouse to pick boxes and place them into the target place.\\

Agent can only walk on horizontal tracks and enter specific regions for picking up boxes. Each agent can only hold one box each time. Each agent can do the actions: \\

1) When the robot is on the track, it can pick up one box whose location is 0.5 away from the robot (either location difference in x or y.). For example, 'pick box{\_}1.5{\_}1.0'Note that the agent can only pick the box near its location, their row locations should have difference of 0.5, and column difference should be 0.0, e.g., agent0 is in track{\_}1 and column{\_}3 and can do 'pick box{\_}1.5{\_}3.0' or 'pick box{\_}0.5{\_}3.0'.\\

2) When the robot is on the track, it can move its position with distance 1 either to the left or to the right. For example, 'move left', 'move right”\\

3) When the robot is on the target, it can move its position to the track to get onto the track and carry the boxes. For example, 'move to track{\_}1”\\

4) When the robot is on the track, it can move its position to the target to pour the box into the target. For example, 'move to target'Note that robots without box on it can also move to target to avoid being obstacle of other robots. All robots moving to the target will pour their boxes. Hence, the final goal is to pour all the boxes into the target. Multiple robots can locate in target in the same time, but cannot be in the same track position in the same time.\\
The warehouse playground has left side column 0 and right side, if the agent column is at these two sides, they can only move right or move left but not both directions.\\
If the agent in the target, it can move to the left side of all the tracks\\
If the agent is in the left side of the track, it can move to the target and drop the box.\\

Your task is to assign each agent the task in the next step. After each step, environments provide updates for each agent and the state of left boxes. Your job is to coordinate the agents optimally to minimize the step number.\\

[Do remember that each position(track and column locations) can only accommodate one agent each step! Hence, you need to avoid the collision with other agents. Actions like move two agents into the same position at the same time or move one agent into the position that already has one agent are not allowed!]\\

Specify your action plan in this format: \{'agent0':'move left', 'agent1':'move to track{\_}1', 'agent2':'pick box{\_}1.5{\_}1.0', 'agent3':'move to target', 'agent4':'move right', 'agent5':'pick box{\_}1.5{\_}3.0'\}. Include an agent only if it has actions in the next step.
\end{boxL}

\begin{boxL}
\textbf{WareHouse Best prompt for GPT-4}\\
\textbf{Score = 0.512 (GPT-4 as the testing LLM)}\\

You are a central planner tasked with the strategic coordination of autonomous mobile agents within a warehouse environment. Your primary goal is to orchestrate the movement of these agents to efficiently transport boxes from their initial locations to a designated target area. Each agent can carry only one box at a time. To successfully accomplish this task, agents must adhere to a set of rules and constraints that govern their actions.\\

The agents can perform the following actions, under specific conditions:\\

1) Pick Up Box: An agent can pick up a box if it is directly adjacent to it on the track, specifically 0.5 units away either in the x or y direction. For instance, an agent positioned at track{\_}1, column{\_}3, can execute 'pick box{\_}1.5{\_}3.0' or 'pick box{\_}0.5{\_}3.0' if the box is present and the agent is not already carrying a box.\\

2) Move Horizontally: An agent on the track can move horizontally by one unit either to the left or to the right, unless it is at the extremities of the tracks (column 0 or the last column), where it can only move away from the extremity. Use the commands 'move left' or 'move right' to direct this action.\\

3) Move to Track: An agent in the target area can move to the leftmost side of any track. The command 'move to track{\_}X' positions the agent at the leftmost point of track{\_}X.\\

4) Move to Target: An agent carrying a box can move to the target area to deposit the box using 'move to target' when the agent is at the leftmost side of the track.\\

The following constraints must be observed:\\

- An agent not carrying a box may move to the target area to prevent obstructing the path of other agents.\\
- Multiple agents can occupy the target area simultaneously, but they must not be positioned on the same track and column at the same time.\\
- Agents at the extremities of the tracks are restricted to moving in one direction only (to the right from column 0 and to the left from the last column).\\
- Collision avoidance is mandatory: no two agents are allowed to occupy the same track and column position at the same time.\\

Your responsibility is to devise a plan for the next move of each agent with the aim of minimizing the total number of steps required. After each move, you will receive updated information about the positions of each agent and the locations of the remaining boxes. Use this information to refine your strategy and prevent collisions.\\

Action plans must be formatted as follows: \{'agent0':'move left', 'agent1':'move to track{\_}1', 'agent2':'pick box{\_}1.5{\_}1.0', 'agent3':'move to target', 'agent4':'move right', 'agent5':'pick box{\_}1.5{\_}3.0'\}. Include an agent in your action plan only if it needs to take action in the next step.\\

The overarching objective is to transport all boxes to the target area with maximum efficiency, in compliance with the established rules and constraints. Your planning must be reflective of the current warehouse conditions, including the agents' positions, whether they are carrying a box, and the box locations, to ensure seamless operations. Use feedback from the environment to adjust future actions, avoiding repetition of actions that were previously indicated as not doable, and ensure that the action plan is precise and includes only necessary agent movements.
\end{boxL}

\begin{boxL}
\textbf{Gridworld1 Human prompt}\\
\textbf{Score = 0.23 (GPT-3.5-turbo-16k-0613 as the testing LLM)}\\
\textbf{Score = 0.73 (GPT-4 as the testing LLM)}\\

You (the robot) are in a grid-like field to pick up all the goals in order and avoid all the obstacles. Each goal and obstacle is assigned to a 1x1 square.\\

The robot can move in four directions: up, down, left, and right. The robot can move to a square only if it is not occupied by an obstacle.\\

If the robot is in the same square with a goal, you can pick up the goal and the square becomes empty.\\

[(1) Note that the coordinate system is different from the Cartesian coordinate system. The origin is at the top left corner. The coordinate representation is [row{\_}number, column{\_}number].\\
For example, if you are in the square [3,2], Move up leads to [2,2], Move down leads to [4,2], Move left leads to [3,1], and Move right leads to [3,3].\\
(2) In your response, you can only use \{\} to specify your action. For example, \{Move up\}. Do not add any other words or symbols in your response. Also use \{\} only once in your whole response
so that we know what is next action without ambiguity.]\\

Please learn from previous steps. Not purely repeat the actions but learn why the state changes or remains in a dead loop. Avoid being stuck in action loops.\\

Do remember do not move to the square occupied by an obstacle! Do remember do not move out of the field! Plan your action in each step based on your relative distance to goals.\\

All the possible actions are: Move up, Move down, Move left, Move right, Pick goal\\

Specify your action in this format at the end of your answer: \{Move up\}, \{Move down\}, \{Move left\}, \{Move right\}, \{Pick goal\}.
\end{boxL}

\begin{boxL}
\textbf{Gridworld1 Best prompt for GPT-3.5-turbo-16k-0613}\\
\textbf{Score = 0.38 (GPT-3.5-turbo-16k-0613 as the testing LLM)}\\

You (the robot) are tasked with navigating a grid-like field to sequentially collect all goals while avoiding obstacles. Each goal and obstacle occupies a distinct 1x1 square on the grid. Your current position is known, and you must use this information to make strategic decisions that adhere to the following optimized, clarified, and refined rules:\\

1. **Immediate Goal Collection**: If a goal is located on your current square, immediately collect it with the action \{Pick goal\} before considering any movement.\\

2. **Enhanced Obstacle and Boundary Avoidance**: Before planning a move, confirm that the intended path is free of obstacles and within the grid limits. The grid's origin is at the top left corner, with coordinates [row{\_}number, column{\_}number]. Do not attempt to move into a square with an obstacle or beyond the grid boundaries.\\

3. **Strategic Goal Pursuit**: Identify the location of the nearest goal using the most efficient path calculation and plan a path towards it, circumventing any obstacles as necessary. Your moves should be calculated to reduce the distance to the nearest goal unless an obstacle dictates a detour.\\

4. **Dynamic Strategy Adaptation**: Reflect on the outcomes of previous actions to enhance your decision-making process. Avoid actions that have previously led to collisions or have not progressed you towards a goal. Adjust your strategy to be more effective.\\

5. **Prioritization of Actions**: The collection of goals is your primary mission. Move only if it is strategic for goal acquisition or essential for obstacle circumvention.\\

6. **Continuous State Assessment and Adjustment**: Consistently verify and update your current state after each action. This includes your position, the positions of goals, and the locations of obstacles to ensure your next action is based on the most current information.\\

7. **Feedback-Driven Action Refinement**: Integrate feedback from the environment and your previous actions to refine your approach. If an action was ineffective or incorrect, adopt a different strategy that complies with the established rules.\\

8. **Explicit and Valid Action Execution**: If an invalid action is attempted, acknowledge the mistake and select a valid and strategic action instead.\\

9. **Precise Obstacle Mapping**: Maintain a clear and updated understanding of obstacle positions relative to your current location to avoid any prohibited moves.\\

10. **Boundary Awareness and Compliance**: Always be aware of the grid boundaries to prevent any attempts to move outside the grid.\\

11. **Error Identification and Strategic Correction**: Recognize any errors in action promptly and correct your course of action to align with the goal-oriented strategy.\\

12. **Effective Feedback Application**: Utilize feedback from the environment to continuously improve your actions, particularly after an unsuccessful or ineffective move.\\

13. **Nearest Goal Prioritization**: Always determine the nearest goal's location from your current position before planning your next move. This ensures that your actions are optimized for goal collection efficiency.\\

14. **State Verification Before Action**: Before planning your next move, verify your current state, including the presence of goals and obstacles, to ensure that your next action is appropriate and strategic.\\

15. **Avoidance of Ineffective Repetition**: Use feedback from the environment to avoid repeating actions that have been proven ineffective or incorrect. Learn from past outcomes to make better decisions.\\

16. **Clear Movement Decision Criteria**: When multiple movement options are available, choose the direction that brings you closest to the nearest goal without violating obstacle and boundary rules. If equidistant, prioritize moves in the following order: up, left, down, right.\\

17. **Loop Prevention and Progress Assessment**: If you find yourself oscillating between two or more squares without making progress, reassess the situation and choose a different path to break the loop. After each move, assess whether you are closer to the nearest goal to ensure progress is being made.\\

18. **Action Execution Confirmation**: After performing an action, confirm its outcome to ensure it was executed as intended and adjust your strategy accordingly.\\

19. **Proactive Error Prevention and Strategic Decision Making**: Before executing any action, proactively consider potential errors and choose the action that has the highest likelihood of success based on the current state and established rules. Make strategic decisions that prioritize goal collection and efficient navigation.\\

20. **Feedback Mechanism Accuracy**: Ensure that the feedback mechanism is correctly interpreting the robot's actions, particularly when collecting goals. If the feedback indicates an error in goal collection when the action was correct, the mechanism should be adjusted to recognize the successful collection.\\

21. **Boundary and Obstacle Confirmation**: Before each move, perform a boundary and obstacle check to confirm that the intended path is valid. This check must be accurate to prevent invalid moves that violate the rules.\\

22. **Goal Collection Confirmation**: When on a square with a goal, confirm the collection of the goal before any movement is considered. This action must be prioritized over all others to align with the mission's primary objective.\\

23. **Error Recognition and Recovery**: The robot must be capable of recognizing when an error has occurred, such as attempting to move into an obstacle or outside the grid, and take immediate corrective action.\\

24. **Comprehensive State Verification**: Continuously verify the robot's current state, including its position, the positions of goals, and the locations of obstacles, before planning and executing the next move.\\

25. **Valid Action Assurance**: Prior to action execution, ensure that the chosen action is valid and possible within the current state of the environment.\\

26. **Intelligent Directional Decision**: When the robot is equidistant from a goal or has multiple paths to choose from, it should consider the history of its moves and environmental feedback to select a path that is most likely to be successful, avoiding previously unsuccessful paths.\\

27. **Goal Proximity Alert**: The robot should have an internal alert system that triggers when it is adjacent to a goal, prompting it to prioritize the goal's collection before any other action.\\

28. **Consistent Path Following**: When the robot has initiated a successful path towards a goal, it should continue on that path unless an obstacle or boundary requires a change in direction.\\

Execute only one action per response in the specified format to maintain clarity and avoid ambiguity: \{Move up\}, \{Move down\}, \{Move left\}, \{Move right\}, \{Pick goal\}. Your next action should be clearly indicated using this format.
\end{boxL}

\begin{boxL}
\textbf{Gridworld1 Best prompt for GPT-4}\\
\textbf{Score = 0.86 (GPT-4 as the testing LLM)}\\

You (the robot) are tasked with navigating a grid-like field to collect all goals in sequence while avoiding obstacles. Each goal and obstacle is located on a separate 1x1 square within the grid.\\

Your capabilities include moving in the four cardinal directions: up, down, left, and right. You are only permitted to move onto a square if it is not occupied by an obstacle.\\

When you reach a square that contains a goal, you must pick up the goal, which will then clear the square.\\

Adhere to these optimized guidelines for navigation and task execution:\\

1. The grid's origin is at the top left corner, with positions denoted by [row{\_}number, column{\_}number]. For example, from [3,2], {Move up} takes you to [2,2], {Move down} to [4,2], {Move left} to [3,1], and {Move right} to [3,3].\\

2. Clearly communicate your intended action using braces {}, and limit your response to one action for clarity, such as: {Move up}.\\

3. Use the history of your actions and the feedback received to avoid repeating ineffective moves and to prevent looping behavior. Learn from past outcomes to improve your decision-making process.\\

4. Before each move, check for obstacles in all four adjacent squares. Never attempt to move into a square with an obstacle.\\

5. Stay within the grid's boundaries to avoid moving off the field.\\

6. Prioritize goals based on proximity, and plan the most efficient route to the nearest goal, taking into account the positions of all goals and obstacles. Use a heuristic such as the Manhattan distance to determine the closest goal.\\

7. Once you have chosen a direction that brings you closer to a goal, continue moving in that direction until you reach the goal, encounter an obstacle, or would move outside the grid's boundaries.\\

8. When you reach a goal's location, immediately pick up the goal with the action {Pick goal}.\\

9. Continuously update your knowledge of the grid's current state, including the locations of goals, obstacles, and your own position, to avoid repeating ineffective actions or entering into loops.\\

10. After each move, dynamically adjust your path based on new information and feedback to ensure the most efficient completion of the task.\\

11. If a chosen path is blocked by an obstacle or leads to a dead end, backtrack and select an alternative route that brings you closer to the nearest goal without revisiting recently occupied squares unless it is part of an efficient path to a goal.\\

12. If you find yourself repeating the same action without progress, reassess your strategy and consider all remaining goals and obstacles to find a new efficient path.\\

13. Implement a strategy to recognize when you are not making progress towards a goal, such as visiting the same square multiple times without collecting a goal, and then reassess your path.\\

Your ultimate goal is to collect all goals in the most efficient manner possible, circumventing obstacles and staying within the grid's limits. Implement these optimized guidelines to dynamically refine your path and ensure successful task completion.\\

The permissible actions are: \{Move up\}, \{Move down\}, \{Move left\}, \{Move right\}, \{Pick goal\}.
\end{boxL}

\begin{boxL}
\textbf{Gridworld2 Human prompt}\\
\textbf{Score = 0.036 (GPT-3.5-turbo-16k-0613 as the testing LLM)}\\
\textbf{Score = 0.26 (GPT-4 as the testing LLM)}\\

You (the robot) are in a grid-like field to pick up all the goals in order and avoid all the obstacles. Each goal and obstacle is assigned to a 1x1 square.\\

The robot can move in four directions: up, down, left, and right. The robot can move to a square only if it is not occupied by an obstacle.\\

If the robot is in the same square with a goal, you can pick up the goal and the square becomes empty. However, you should pick the goals in order, from 0 to larger.\\

If the goal in the current square is not the next goal, you can not pick it up. You should move to other squares to find the next goal.\\

[(1) Note that the coordinate system is different from the Cartesian coordinate system. The origin is at the top left corner. The coordinate representation is [row{\_}number, column{\_}number].\\
For example, if you are in the square [3,2], Move up leads to [2,2], Move down leads to [4,2], Move left leads to [3,1], and Move right leads to [3,3].\\
(2) The robot should pick up all the goals in order, index from 0 to larger. For example, if there are 3 goals, the robot should pick up the goal{\_}0 first, then the goal 1, and finally the goal 2.\\
(3) In your response, you can only use \{\} to specify your action. For example, {Move up}. Do not add any other words or symbols in your response. Also use \{\} only once in your whole response\\
so that we know what is next action without ambiguity.]\\

Please learn from previous steps. Not purely repeat the actions but learn why the state changes or remains in a dead loop. Avoid being stuck in action loops.\\

Do remember do not move to the square occupied by an obstacle! Do remember do not move out of the field! Plan your action in each step based on your relative distance to goals.\\

All the possible actions are: Move up, Move down, Move left, Move right, Pick goal\\

Specify your action in this format at the end of your answer: \{Move up\}, \{Move down\}, \{Move left\}, \{Move right\}, \{Pick goal\}.
\end{boxL}

\begin{boxL}
\textbf{Gridworld2 Best prompt for GPT-3.5-turbo-16k-0613}\\
\textbf{Score = 0.17 (GPT-3.5-turbo-16k-0613 as the testing LLM)}\\

You (the robot) are tasked with navigating a grid-like field to collect a series of numbered goals in the correct numerical sequence, from goal{\_}0 to the highest-numbered goal, while avoiding obstacles. Each goal and obstacle occupies a distinct 1x1 square on the grid.\\

Objective:\\
- Collect all goals in numerical order without violating any movement or collection rules.\\

Movement Rules:\\
- You may move one square at a time in one of four directions: up, down, left, or right.\\
- You must not move into squares with obstacles or beyond the grid boundaries.\\

Goal Collection Rules:\\
- You must pick up a goal only if it is the next in sequence and you are on the same square as that goal.\\
- Once a goal is picked up, the square it occupied becomes traversable.\\
- If you encounter a goal that is not the next in sequence, you cannot pick it up and must navigate to find the correct goal.\\

Coordinate System:\\
- The grid's origin is at the top left corner, with coordinates given as [row{\_}number, column{\_}number].\\
- Moving up decreases the row number, moving down increases the row number, moving left decreases the column number, and moving right increases the column number.\\

Action Specification:\\
- Specify your action using only one of the following commands within curly braces: \{Move up\}, \{Move down\}, \{Move left\}, \{Move right\}, \{Pick goal\}.\\
- Do not include any additional words, symbols, or multiple actions within the braces.\\

Adaptive Learning and Error Correction:\\
- Learn from the outcome of each action to avoid ineffective or rule-violating moves.\\
- Continuously update your strategy based on your current position, the positions of remaining goals, and the locations of obstacles.\\
- Avoid repeating a sequence of moves that does not change your state or bring you closer to the next goal.\\
- If an action does not progress towards the goal or violates the rules, reassess and choose a different action.\\

Action Planning and Efficiency:\\
- Before each move, verify your current position and assess the most efficient path to the next goal, avoiding obstacles and grid edges.\\
- If you are on the same square as the next goal, the only valid action is \{Pick goal\}.\\
- If the next goal is not directly accessible, plan an alternative route that brings you closer to the goal without violating movement rules.\\
- Prioritize picking up the goal over moving if you are on the goal square.\\

State Verification:\\
- Before suggesting an action, confirm your current position and the location of the next goal to ensure the action is valid and efficient.\\

Your ultimate goal is to collect all goals in the correct sequence as efficiently as possible, adhering strictly to the movement and collection rules.
\end{boxL}

\begin{boxL}
\textbf{Gridworld2 Best prompt for GPT-4}\\
\textbf{Score = 0.60 (GPT-4 as the testing LLM)}\\

You (the robot) are tasked with navigating a grid-like field to sequentially collect goals, labeled from goal{\_}0 to the highest-numbered goal, while avoiding obstacles. Each goal and obstacle occupies a distinct 1x1 square on the grid.\\

Your movements are limited to four directions: up, down, left, and right. You may only move onto a square if it is not occupied by an obstacle.\\

**Critical Rule for Goal Collection**: You must collect goals in strict numerical order, starting with goal{\_}0. Before suggesting \{Pick goal\}, you must perform a state verification checkpoint. This involves confirming that the goal is the next in the numerical sequence and that you are on the correct square.\\

Adhere to these optimized rules for successful navigation and goal collection:\\

1. **Sequential Goal Collection**: Before suggesting \{Pick goal\}, explicitly state the number of the goal you are attempting to collect and confirm it is the next in the sequence. Do not attempt to collect a goal if it is not the correct one in the order.\\

2. **State and Position Awareness**: Continuously update your current position on the grid and the location of the next goal. Plan your moves to efficiently reach the next goal, avoiding obstacles and grid boundaries.\\

3. **Action Preconditions**: Only suggest {Pick goal} when you have verified that you are on the correct goal square and that the goal is the next in the sequence. Provide a clear justification for your action by stating your current position and the goal's position.\\

4. **Learning from Errors**: If an action is ineffective, analyze the outcome, learn from the mistake, and adjust your strategy to avoid repeating the error. State the reason for the error and the adjustment you will make.\\

5. **Obstacle and Boundary Consideration**: Plan moves that avoid obstacles and stay within the grid's boundaries to ensure a clear path to the next goal.\\

6. **Strategic Path Planning**: Choose the most direct and efficient path to the next goal, avoiding obstacles and boundaries. Re-evaluate your path after each move.\\

7. **Single Action Response**: Provide only one action in the specified format per response: \{Action\}.\\

8. **Adaptive Strategy**: As goals are collected and the grid's layout changes, adapt your strategy to ensure continuous progress towards the next goal in sequence.\\

9. **Avoiding Action Loops**: Recognize and break free from loops of non-productive actions by altering your approach. Implement a mechanism to detect repeated non-productive actions and change strategy if necessary.\\

10. **Feedback Utilization**: Use feedback from the environment and previous errors to inform your subsequent actions and improve your navigation strategy.\\

11. **Explicit Change of Strategy**: If a strategy is not leading to success, explicitly state and implement a new approach to find a path to the goal.\\

12. **Clear Movement Rules**: Adhere to the rules of movement and goal collection without ambiguity, ensuring that each action is deliberate and aligns with the goal sequence.\\

Before suggesting an action, confirm your current position, the location of the next goal, and the absence of obstacles in your path. Justify your action choice by referencing the goal sequence and your current position relative to the next goal. If an error occurs, analyze why it happened and adjust your strategy accordingly.\\

The coordinate system for the grid has its origin at the top left corner, with coordinates represented as [row{\_}number, column{\_}number]. For example, from [3,2], \{Move up\} results in [2,2], \{Move down\} in [4,2], \{Move left\} in [3,1], and \{Move right\} in [3,3].\\

Your possible actions are: \{Move up\}, \{Move down\}, \{Move left\}, \{Move right\}, \{Pick goal\}. Respond with only one of these actions, formatted as shown, at the end of each turn. Before taking an action, ensure it aligns with the goal sequence and the rules provided.
\end{boxL}

\begin{boxL}
\textbf{Blocksworld Human prompt}\\
\textbf{Score = 0.19 (GPT-3.5-turbo-16k-0613 as the testing LLM)}\\
\textbf{Score = 0.71 (GPT-4 as the testing LLM)}\\

I am playing with a set of blocks where I need to arrange the blocks into stacks. Here are the actions I can do

Pick up a block\\
Unstack a block from on top of another block\\
Put down a block\\
Stack a block on top of another block\\

I have the following restrictions on my actions:\\
I can only pick up or unstack one block at a time.\\
I can only pick up or unstack a block if my hand is empty.\\
I can only pick up a block if the block is on the table and the block is clear. A block is clear if the block has no other blocks on top of it and if the block is not picked up.\\
I can only unstack a block from on top of another block if the block I am unstacking was really on top of the other block.\\
I can only unstack a block from on top of another block if the block I am unstacking is clear.\\
Once I pick up or unstack a block, I am holding the block.\\
I can only put down a block that I am holding.\\
I can only stack a block on top of another block if I am holding the block being stacked.\\
I can only stack a block on top of another block if the block onto which I am stacking the block is clear.\\
Once I put down or stack a block, my hand becomes empty.\\
Once you stack a block on top of a second block, the second block is no longer clear.\\

Please learn from previous steps. Not purely repeat the actions but learn why the state changes or remains in a dead loop. Avoid being stuck in action loops.
Specify your action in this format at the end of your answer: pick up the \{\}, put down the \{\}, stack the \{\} on top of the \{\},unstack the \{\} from on top of the {}.
\end{boxL}

\begin{boxL}
\textbf{Blocksworld Best prompt for GPT-3.5-turbo-16k-0613}\\
\textbf{Score = 0.6 (GPT-3.5-turbo-16k-0613 as the testing LLM)}\\

I am tasked with arranging a set of blocks into specific configurations through a block-stacking activity. My available actions are:

- Pick up a block that is clear and on the table.\\
- Unstack a clear block from the top of another block.\\
- Put down a block onto the table, ensuring my hand is empty afterward.\\
- Stack a block onto another clear block, ensuring my hand is empty afterward.\\

To ensure successful completion of these actions, I must follow these rules:

1. I can only manipulate one block at a time.\\
2. My hand must be empty before I can pick up or unstack a block.\\
3. A block is considered clear and eligible to be picked up if it has no blocks on top of it, is on the table, and is not being held.\\
4. I can unstack a block only if it is the topmost block on another and there are no blocks above it.\\
5. When I pick up or unstack a block, I will be holding it.\\
6. I can only put down or stack a block that I am currently holding.\\
7. A block can be stacked onto another only if the bottom block is clear.\\
8. My hand must be empty before and after I place or stack a block.\\
9. Stacking a block on top of another makes the bottom block non-clear.\\

To optimize task execution and avoid errors, I will adhere to the following strategies:

- Conduct a comprehensive state verification before each action to ensure all preconditions are met: my hand is empty before picking up or unstacking; the block is clear, on the table, and not being held for picking up; and I am holding a block before putting down or stacking.\\
- Maintain an accurate and constantly updated mental model of the block arrangement, noting the clear status of each block, the current stack configurations, and whether my hand is empty or holding a block.\\
- Develop a strategic action plan that is directly aligned with achieving the desired final block configuration, taking into account the current state and the steps required to reach the goal.\\
- Integrate feedback after each action to assess the success of the action and to update my strategy, ensuring that I do not repeat ineffective actions and that I learn from any mistakes to avoid non-progressive loops.\\
- Communicate my intended actions clearly and precisely, using the format: 'pick up \{color\} block', 'put down \{color\} block', 'stack \{color\} block on top of \{color\} block', 'unstack \{color\} block from on top of \{color\} block'.\\
- Implement an enhanced loop detection mechanism to identify and interrupt any repetitive, non-progressive action sequences, choosing a different action if necessary.\\
- Set and pursue intermediate goals that are necessary steps towards the final configuration, ensuring that each action is deliberate and contributes to the end goal in an incremental fashion.\\
- Establish a timeout or step limit to prevent exceeding the query time limit without completing the task, and reassess my strategy if progress stalls to ensure that I am always moving towards task completion.\\
- Explicitly state the preconditions that have been verified before proposing an action, and clearly communicate any adjustments made to the strategy based on feedback received.\\
- Introduce a robust error handling strategy that allows for backtracking or reassessment of the plan when an action fails, ensuring alternative actions adhere to the rules and contribute to the final goal.\\

By following these refined guidelines and continuously updating my approach based on the state of the blocks and the feedback received, I aim to efficiently and effectively complete the block-stacking task.
\end{boxL}

\begin{boxL}
\textbf{Blocksworld Best prompt for GPT-4}\\
\textbf{Score = 0.95 (GPT-4 as the testing LLM)}\\

To effectively arrange a set of blocks into the desired stacks, adhere to the following structured approach, which has been refined based on previous feedback and identified errors:\\

1. **Evaluate the Goal State**: Examine the goal state configuration in detail and compare it with the current state to discern the exact actions required to achieve the goal. Maintain a clear and constant visualization of the final desired arrangement of blocks throughout the task.\\

2. **Action Sequence Planning**: Construct a strategic plan that delineates a sequence of actions that will methodically transition the current state towards the goal state. Prioritize actions that make definitive progress towards the goal and eliminate redundant or non-contributory steps.\\

3. **Preconditions Verification**: Before initiating any action, rigorously check that all preconditions are satisfied. Confirm that your hand is empty before attempting to pick up or unstack a block, and ensure that the block to be manipulated is unobstructed and either on the table or atop another block.\\

4. **Execute Actions**: Implement the necessary actions, strictly following the prescribed format and constraints:

- To pick up a block: 'pick up the \{color\} block.'\\
- To unstack a block: 'unstack the \{color\} block from on top of the \{color\} block.'\\
- To put down a block: 'put down the \{color\} block.'\\
- To stack a block: 'stack the \{color\} block on top of the \{color\} block.'\\

5. **Loop and Error Prevention**: Vigilantly observe your actions to identify any repetitive or non-productive patterns. Upon detecting a loop, promptly reassess and revise the action plan. Document past errors to prevent their recurrence.\\

6. **State Change Analysis**: After executing an action, conduct a state change analysis to verify that the system is incrementally closer to the goal state. If the action does not yield the expected progress, reevaluate and modify the plan.\\

7. **Continuous Learning**: Log the results of previous actions, noting both successes and failures, to refine future strategies and enhance task efficiency.\\

8. **Clear Goal Specification**: Keep the goal state at the forefront of your strategy, ensuring that every action is intentionally aimed at achieving that state.\\

9. **Feedback Integration**: After each action, incorporate feedback to improve your understanding of the current state and to guide future actions.\\

10. **Loop Detection and Correction**: Establish a robust mechanism to detect when you are in a loop and to prompt a strategic reassessment of the action plan.\\

11. **Goal State Reassessment**: Frequently reevaluate both the goal state and the current state to confirm that your actions are consistently aligned with the goal.\\

12. **Action Format Standardization**: Adhere to the specified action format with precision, refraining from adding prefixes or narrative explanations unless the context demands it.\\

13. **State Change Verification**: Post-action, ensure that the state has altered as intended and that the system is nearer to the goal state.\\

14. **Error Handling**: Enhance error handling protocols to avert the repetition of unsuccessful actions.\\

15. **Optimize Query Time**: Employ methods to expedite the planning and execution of actions, aiming for task completion with optimal efficiency.\\

This refined approach is designed to systematically guide you towards arranging the blocks into the goal state configuration while minimizing errors and enhancing task performance.
\end{boxL}

\begin{boxL}
\textbf{Logistics Human prompt}\\
\textbf{Score = 0.083 (GPT-3.5-turbo-16k-0613 as the testing LLM)}\\
\textbf{Score = 0.50 (GPT-4 as the testing LLM)}\\

You have to plan logistics to transport packages within cities via trucks and between cities via airplanes. Locations within a city are directly connected (trucks can move between any two such locations), and so are the cities. In each city there is exactly one truck and each city has one location that serves as an airport.\\

Here are the actions that can be performed:

Load a package into a truck at a location.\\
Load a package into an airplane at a location.\\
Unload a package from a truck at a location.\\
Unload a package from an airplane at a location.\\
Drive a truck from one location to another location within a city.\\
Fly an airplane from one location in a city to another location in another city.\\

The following are the restrictions on the actions:\\
A package can be loaded into a truck only if the package and the truck are in the same location.\\
Once a package is loaded into a truck, the package is not at the location and is in the truck.\\
A package can be loaded into an airplane only if the package and the airplane are in the same location.\\
Once a package is loaded into an airplane, the package is not at the location and is in the airplane.\\
A package can be unloaded from a truck only if the package is in the truck.\\
Once a package is unloaded from a truck, the package is not in the truck and is at the location of the truck.\\
A package can be unloaded from an airplane only if the package in the airplane.
Once a package is unloaded from an airplane, the package is not in the airplane and is at the location of the airplane.\\
A truck can be driven from one location to another if the truck is at the from-location and both from-location and to-location are locations in the same city.
Once a truck is driven from one location to another, it is not at the from-location and is at the to-location.\\
An airplane can be flown from one city to another if the from-location and the to-location are airports and the airplane is at the from-location.\\
Once an airplane is flown from one city to another the airplane is not at the from-location and is at the to-location.\\

Please learn from previous steps. Not purely repeat the actions but learn why the state changes or remains in a dead loop. Avoid being stuck in action loops.
Specify your action in this format at the end of your answer: load \{\} into \{\} at \{\}, unload \{\} from \{\} at \{\}, drive \{\} from \{\} to \{\} in \{\}, fly \{\} from \{\} to \{\}.
\end{boxL}

\begin{boxL}
\textbf{Logistics Best prompt for GPT-3.5-turbo-16k-0613}\\
\textbf{Score = 0.18 (GPT-3.5-turbo-16k-0613 as the testing LLM)}\\

To optimize the logistics of transporting packages within cities using trucks and between cities using airplanes, follow these enhanced and precise guidelines:\\

1. **Loading and Unloading Preconditions:**\\
- Load a package into a truck only when the package and the truck are co-located.\\
- Load a package into an airplane only at an airport, ensuring both the package and the airplane are present.\\
- Unload a package from a truck only if it has been verified that the package is in that truck.\\
- Unload a package from an airplane only if it has been verified that the package is in that airplane.\\

2. **Movement Rules:**\\
- Trucks are restricted to travel within their respective city limits.\\
- Airplanes must fly between airports in different cities without exception.\\

3. **State Changes:**\\
- Reflect the package's new location as inside the vehicle upon loading and at the vehicle's location upon unloading.\\

4. **Action Format:**\\
- Actions must be articulated as follows:\\
- For loading/unloading: 'load \{package\} into \{vehicle\} at \{location\}' or 'unload \{package\} from \{vehicle\} at \{location\}'\\
- For driving: 'drive \{truck\} from \{from-location\} to \{to-location\} in \{city\}'\\
- For flying: 'fly \{airplane\} from \{from-airport\} to \{to-airport\}'\\

5. **Feedback and Learning:**\\
- Update the state of packages, trucks, and airplanes with each action taken.\\
- Log unsuccessful actions due to precondition failures and avoid their repetition.\\
- Refine plans based on feedback to ensure all actions are valid and goal-aligned.\\

6. **Goal-Oriented Strategy:**\\
- Actions must form a logical sequence that advances a package towards its destination in the most direct manner possible.\\

7. **Avoiding Loops:**\\
- Exclude any action that has been attempted unsuccessfully.\\
- Keep a comprehensive log of actions to identify and prevent cyclical patterns, revising the strategy as needed.\\

8. **Task Decomposition:**\\
- Segment the task into discrete sub-tasks, such as intra-city and inter-city package transfers.\\
- Tackle each sub-task systematically, one at a time.\\

9. **Time Management:**\\
- Streamline the planning process to ensure task completion within a set timeframe.\\
- Give precedence to actions that maximize time efficiency while complying with the above guidelines.\\

By adhering to these updated guidelines, you will devise a logistics plan that is both accurate and efficient, guaranteeing the successful delivery of packages to their designated locations.
\end{boxL}

\begin{boxL}
\textbf{Logistics Best prompt for GPT-4}\\
\textbf{Score = 0.74 (GPT-4 as the testing LLM)}\\

Your task is to manage the logistics of transporting packages within and between cities using trucks and airplanes. Each city has a network of locations for truck movement and an airport for airplane transfers. There is one truck per city for local deliveries and one airport per city for intercity transfers.\\

To enhance logistics operations and avoid errors, follow these optimized steps:\\

1. **State Verification**: Prior to any action, rigorously confirm the current locations of all packages, trucks, and airplanes. This step is crucial to ensure that all subsequent actions are based on the most recent and accurate state information.\\

2. **Action Execution**: Execute actions strictly adhering to these preconditions:\\
- Load a package into a truck at a location only if the package and the truck are confirmed to be at that location.\\
- Load a package into an airplane at an airport only if the package and the airplane are confirmed to be at that airport.\\
- Unload a package from a truck at a location only if the package is confirmed to be in that truck.\\
- Unload a package from an airplane at an airport only if the package is confirmed to be in that airplane.\\
- Drive a truck from one location to another within the same city only if the truck's presence at the starting location is confirmed.\\
- Fly an airplane from one city's airport to another city's airport only if the airplane's presence at the starting airport is confirmed.\\

3. **State Update**: Immediately after each action, update the environment state to reflect the new locations of packages, trucks, and airplanes. This updated state must be used for verifying preconditions for the next actions.\\

4. **Efficient Planning**: Deliver all packages to their destinations using the fewest actions possible. Prioritize the shortest routes and avoid any actions that do not directly contribute to reaching the delivery goals.\\

5. **Adaptive Learning**: Utilize feedback from the outcomes of previous actions to continuously refine planning strategies. Avoid repeating ineffective actions and adjust plans based on the latest state information and feedback.\\

6. **Error Management**: If an action fails, quickly reassess the situation based on the current state and propose a new, valid action that moves towards the delivery goals.\\

7. **Clear Action Formatting**: Clearly express actions using the specified structure to avoid misunderstandings:\\

- load \{package\} into \{truck/airplane\} at \{location/airport\}\\
- unload \{package\} from \{truck/airplane\} at \{location/airport\}\\
- drive \{truck\} from \{location\} to \{location\} in \{city\}\\
- fly \{airplane\} from \{airport\} to \{airport\}\\

8. **Goal-Focused Actions**: Ensure every action is purposeful and directly contributes to the final destination of the packages. Eliminate any actions that are not goal-oriented.\\

9. **Time-Efficient Queries**: Streamline the planning process to complete tasks within the query time limit, maintaining a balance between swift operations and careful action validation.\\

10. **Simplified Instructions**: Provide instructions that are clear, concise, and easy to follow, ensuring they are understood and executed correctly.\\

By diligently following these optimized guidelines, you will significantly improve the efficiency and accuracy of the logistics operation for package delivery.
\end{boxL}

\end{document}